\pdfoutput=1
\documentclass[final]{kaist-ucs}

\usepackage[numbers]{natbib}
\usepackage{bibentry}
\usepackage{hyperref}
\usepackage{url}
\usepackage{kotex}
\usepackage{pifont}
\usepackage[noend]{algorithmic}
\usepackage{algorithm}

\usepackage{graphicx}
\usepackage{subcaption}
\usepackage{booktabs}
\usepackage{multirow}
\usepackage{pgfplots}
\usepackage{enumitem}
\usepackage{wrapfig}
\usepackage{makecell}
\usepackage{comment}
\hypersetup{
    colorlinks = false
}

\usepackage[utf8]{inputenc} 
\usepackage[T1]{fontenc}    
\usepackage{nicefrac}       
\usepackage{microtype}      
\usepackage{xcolor}         

\usepackage{bbm}

\usepackage{amsthm} 
\theoremstyle{plain}
\newtheorem{theorem}{Theorem}[section]

\newtheorem{lemma}[theorem]{Lemma}

\theoremstyle{definition}
\newtheorem{definition}[theorem]{Definition}
\newtheorem{assumption}[theorem]{Assumption}

\theoremstyle{remark}

\usepackage{xspace}
\usepackage{array}
\usepackage{mathtools}
\usepackage{balance}
\usepackage{bm}
\usepackage{placeins}
\usepackage{csquotes}
\usepackage{bbold}

\newcommand\Tstrut{\rule{0pt}{2.4ex}}       

\DeclareMathOperator*{\argminB}{argmin}
\DeclareMathOperator*{\argmaxA}{arg\,max}
\DeclareMathOperator*{\argmaxB}{argmax}
\usepackage{rotating}
\usepackage{lscape}

\usepackage{graphics}

\newcommand{\algnamea}{{TAUFE}}
\newcommand{\algname}{{MQNet}}
\newcommand{\algnamec}{{Prune4ReL}}
\newcommand{\algnamed}{{FP-Instruction}}
\newcommand{\tblalgnamec}{{Pr4ReL}}

\title[korean]{노이즈 데이터에서 딥러닝을 위한 \\ 정보력 높은 특성과 샘플 선별}
\title[english]{Prioritizing Informative Features and Examples \\ for Deep Learning from Noisy Data}

\author[korean] {박}{동 민}
\author[korean2] {박}{동민}    
\author[chinese]{朴}{東 珉}
\author[english]{Park}{Dongmin}

\advisor[major]{이 재 길}{Jae-Gil Lee}{signed}
\advisor[major2]{이재길}{Jae-Gil Lee}{signed} 
\advisorinfo{Professor of School of Computing} 
\department{DS}{engineering}{b}

\referee[1]{이 재 길}
\referee[2]{이 문 용}
\referee[3]{신 진 우}
\referee[4]{김 희 영}
\referee[5]{이 강 욱}

\approvaldate{2023}{12}{11}

\refereedate{2023}{12}{11}

\gradyear{2024}

\begin{document}


   \thesisinfo
    \begin{summary}      
    심층신경망은 양질의 대용량 데이터를 기반으로 컴퓨터 비전, 자연어 처리 등의 다양한 분야에서 눈부신 성공을 거두었다. 반면, 실세계에서 수집된 데이터는 지저분한 노이즈를 수반할때가 많은데, 심층신경망의 높은 표현 성능은 이러한 노이즈를 불필요하게 암기하여 성능 하락의 주요한 원인이 되고 있다. 노이즈에 강건한 심층 신경망 학습방법들이 활발히 연구되어 왔지만, 대부분의 연구는 모델 학습과정을 개선하는데에 집중하고 있다. 반면, 노이즈 데이터는 모델 학습 과정 이외에도 데이터 선별과 정제, 레이블링을 포함한 심층신경망 모델 개발과정 전반에 걸쳐 악영향을 끼치고있다. 예를들어, 목표작업에 관계없는 분포외 데이터는 목표 작업에 관련된 레이블을 달 수 없으므로 레이블링을 하는 사람들의 시간적 비용을 낭비하기도 하며, 미처 정제되지 못한 잘못된 레이블을 가진 노이즈 데이터는 모델 학습 성능에 악영향을 주기도 한다. 이에따라 데이터내의 정보력 높은 특성과 샘플을 데이터 전처리 및 모델 학습 시스템 전반에 걸쳐 체계적으로 활용하는 방식에 대한 연구의 필요성이 대두대고 있다.
    
    본 학위 논문에서는 심층신경망 모델 개발과정 전반에 걸쳐 정보력 높은 특성과 샘플을 효과적으로 선별하는 체계적인 방식을 제안한다. 구체적으로는,  정보력 높은 특성과 샘플 선별을 통해 심층 학습 개발 과정의 특성 학습, 능동 학습, 데이터 선별 단계의 성능을 개선한다. 
    첫번째로, 추가적인 분포외 데이터를 사용하여 목표 모델이 분포외 데이터에서는 등장하지 않는 정보력 높은 특성들만 선별할 수 있는 특성 정규화 방식을 제안한다. 분포외 데이터의 노이즈 특성을 이용하여 타겟 분포의 노이즈 특성을 불활성시킬 수 있다. 
    두번째로, 레이블이 되지않은 노이즈 데이터에 대해 정보력이 높은 샘플 선별 방식을 제안하여 능동 학습의 레이블링시 비용 낭비를 효과적으로 줄인다. 정보력 높은 샘플을 뽑을때 많은 노이즈 샘플이 선택되는 순도-정보도 딜레마를 풀기위하여 두 요인의 최선의 균형을 찾는 메타 모델을 제안한다. 마지막으로, 레이블이 되어있는 노이즈 데이터에 대해 정보력이 높은 샘플 선별 방식을 제안하여 선별된 데이터에서 학습된 모델 성능을 최대한 유지하며 학습 효율을 개선한다. 
    레이블된 이미지 노이즈 데이터에 대해서는 이웃 샘플 신뢰도를 고려한 데이터 선별 방식으로 최신 재레이블링 모델의 성능을 유지하며, 레이블된 텍스트 노이즈 데이터에 대해서는 다양성을 고려한 집단 프롬프팅 방식으로 언어지시 데이터를 선별하여 거대언어모델의 성능을 유지및 개선한다.
    종합적으로, 제안된 방식은 심층 신경망 개발 과정을 노이즈 데이터에 강건하게 만드는 통합적인 시스템으로서 실세계에서 발생하는 노이즈 특성과 샘플들을 동시에 효과적으로 완화시킬 수 있다.
    \end{summary}
   
    \begin{Korkeyword}
    심층 학습, 노이즈 데이터, 분포외 데이터, 특성 정규화, 능동 학습, 데이터 가지치기, 핵심집합 선별, 거대언어모델 
    \end{Korkeyword}

    \begin{abstract}
    Deep neural networks\,(DNNs) have achieved remarkable success in various fields such as computer vision and natural language processing based on vast amounts of high-quality data. However, real-world data collections are invariably noisy and DNNs are reported to unintentionally memorize most of such noise, resulting in severe performance degradation. Although noise-robust learning approaches for DNNs have been actively developed, most works focus on improving the model training stage. However, such noise data disrupt DNNs not only during model training but throughout the entire model development process including sample selection, cleaning, and labeling. For example, the unlabeled noisy data obtained from out-of-distribution waste the labeling cost since a human labeler can not assign any label on them, while the non-filtered labeled noisy data can significantly degrade the model performance. This calls attention to developing a systematic method to avoid such noise and utilize highly informative features and examples throughout the model development process. 
    
    In this dissertation, we propose a systemic framework that \emph{prioritize informative features and examples} to enhance each stage of the development process.
    Specifically, we prioritize informative features and examples and improve the performance of feature learning, data labeling, and data selection.
    We first propose an approach to extract only informative features that are inherent to solving a target task by using auxiliary out-of-distribution data. We deactivate the noise features in the target distribution by using that in the out-of-distribution data.
    Next, we introduce an approach that prioritizes informative examples from unlabeled noisy data in order to reduce the labeling cost of active learning.
    In order to solve the purity-information dilemma, where an attempt to select informative examples induces the selection of many noisy examples, we propose a meta-model that finds the best balance between purity and informativeness.
    Lastly, we suggest an approach that prioritizes informative examples from labeled noisy data to preserve the performance of data selection.
    For labeled image noise data, we propose a data selection method that considers the confidence of neighboring samples to maintain the performance of the state-of-the-art Re-labeling models.
    For labeled text noise data, we present an instruction selection method that takes diversity into account for ranking the quality of instructions with prompting, thereby enhancing the performance of aligned large language models.
    Overall, our unified framework induces the deep learning development process robust to noisy data, thereby effectively mitigating noisy features and examples in real-world applications. 
    \end{abstract} 
     
    \begin{Engkeyword}
    Deep learning, noisy data, out-of-distribution data, feature regularization, active learning, data pruning, coreset selection, large language models
    \end{Engkeyword}

    \addtocounter{pagemarker}{1}                 
    \newpage

    \tableofcontents

    \listoftables

    \listoffigures



\chapter{Introduction}
\label{sec:overall_introduction}

\begin{figure*}[b]
\begin{center}
\includegraphics[width=\linewidth]{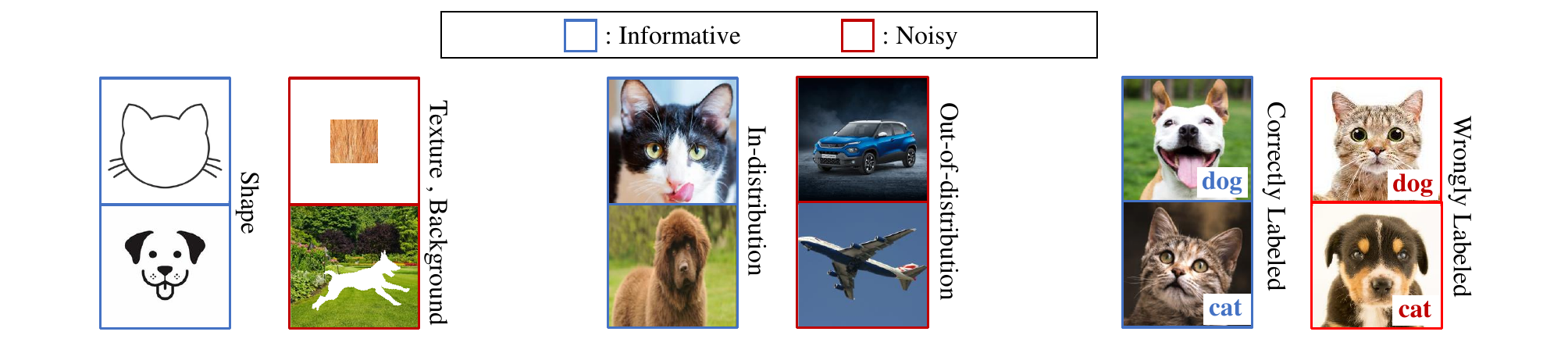}
\end{center}
\vspace{-0.0cm}
\hspace*{1.3cm} {\small (a) Feature types.} \hspace*{1.2cm} {\small (b) Unlabeled example types.} \hspace*{0.8cm} {\small (c) Labeled example types.}
\caption{
Informative/noisy features and examples.
}
\label{fig:informative_noisy_feature_example}
\end{figure*}

\section{Motivation and Background}

Deep neural networks\,(DNNs) have achieved remarkable success in various fields such as computer vision and natural language processing based on vast amounts of high-quality data\,\cite{dosovitskiy2020image, brown2020language, radford2021learning}. 
However, real-world data collections are invariably noisy and DNNs are reported to unintentionally memorize most of such noise, i.e., noisy features and examples, resulting in severe performance degradation\,\cite{zhang2016understanding, song2022learning}.
Therefore, \emph{prioritizing informative features and examples over the noisy ones} can be a fundamental way to increase the usability of deep learning in real-world applications with noisy data.

\emph{Noisy features}, which are informally defined as those not relevant to a target task, frequently appear in training data; for example, the background is a noisy feature for recognizing objects in images. 
In fact, many noisy features are statistically correlated with labels, even though they are unnecessary and sometimes even harmful for the target task\,\cite{bahng2020learning}; for example,  the ``desert'' background feature is correlated with ``camels'' because the camels frequently appear in a desert. 
Such noisy features\,(e.g., desert background) rather yield unreliable predictions because they are easily shifted in other unseen data\,(e.g., images of the camels on the road).
On the contrary, \emph{informative features}, which are defined as those semantically relevant to a target task\,(e.g., shape of objects) induce correct and reliable predictions.
This negative effect necessities prioritizing the informative features over the noisy ones in \emph{feature learning} (See Fig\,\ref{fig:informative_noisy_feature_example}(a) for visualization of informative and noisy features).

{Noisy examples} usually consist of two types: (i) {unlabeled} noisy and (ii) {labeled} noisy. 
\emph{Unlabeled noisy examples} are those collected from out-of-distribution\,(OOD); for example, the non-animal images for animal image classification (See Figure \ref{fig:informative_noisy_feature_example}(b)).
Most real-world unlabeled data collections using \emph{casual} data curation processes such as web crawling contain such unlabeled noisy examples; the precision of image retrieval of Google search engine is reported to be 82$\%$ on average, and it is worsened to 48$\%$ for unpopular entities\,\cite{uyar2017investigating, cheshmehsohrabi2021performance}.
When labeling the unlabeled data, these unlabeled noisy examples result in a significant waste of labeling costs because they are unnecessary for the target task; a human annotator is unable to assign a target label to them, wasting labeling time.
In this regard, prioritizing informative examples over noisy examples is crucial for \emph{active learning} (AL), where an active learner iteratively queries a small number of data examples to a human oracle with a limited labeling budget.

\emph{Labeled noisy examples} are the data examples with wrong annotation; for example, a dog image with a cat label (See Figure \ref{fig:informative_noisy_feature_example}(c)).
Many real-world labeled dataset contains this label noise due to the frequent wrong annotation by non-expert humans or automatic labeling tools.
Such labeled noisy examples are widely known to severely degrade the generalization capability of deep learning.
Therefore, selecting informative examples that can preserve the model performance is very important to many machine learning applications such as \emph{data pruning}.

\begin{figure*}[t!]
\begin{center}
\includegraphics[width=\linewidth]{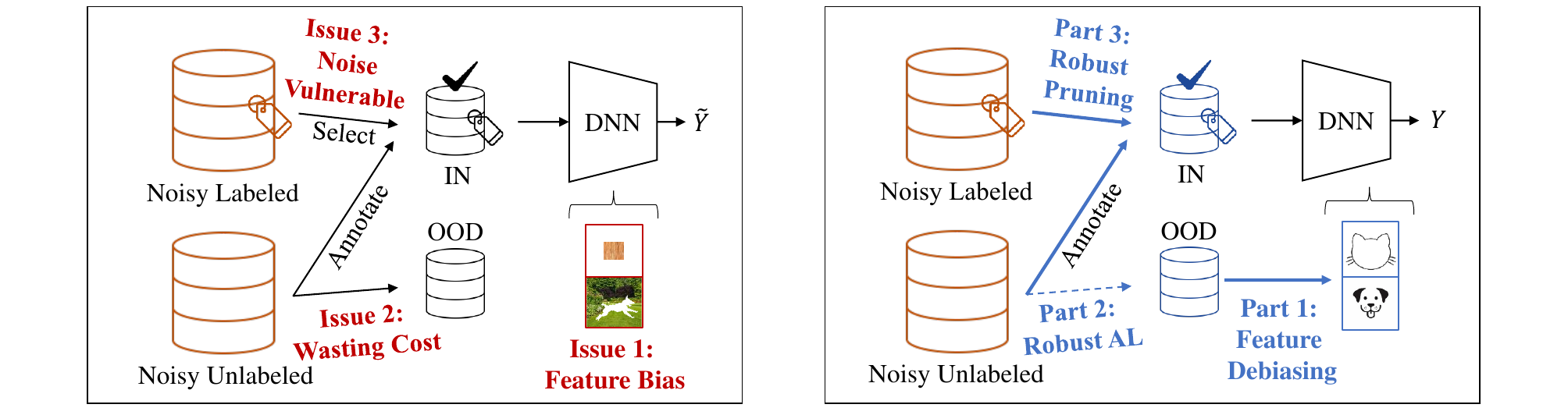}
\end{center}
\vspace{-0.2cm}
\hspace*{0.9cm} {\small (a) Issues on deep learning with data noise.} \hspace*{2.1cm} {\small (b) Proposed framework.}
\caption{
Negative effect of noisy features and examples throughout model development process, and our solution.
}
\label{fig:our_framework}
\end{figure*}

As shown in Figure \ref{fig:our_framework}(a), the noisy features and examples disrupt DNNs not only during model training but throughout the entire model development process, from data preprocessing to feature learning, and even simultaneously.
This motivation calls attention to developing a \emph{unified framework} to mitigate the negative effect of noisy features and examples for robust deep learning.

\section{Main Contributions}

We propose a systemic framework that \emph{prioritize informative features and examples} to enhance each stage of the model development including feature learning, data labeling, and data selection. Figure \ref{fig:our_framework}(b) summarizes the three main parts of this dissertation.

\textbf{Part 1: Noisy Feature Debiasing.}\hspace{0.1cm}
We first propose a feature debiasing approach to extract only informative features that are inherent to solving a target task by using auxiliary OOD data.
Due to high expressive power of a DNN, its prediction can be easily biased to noisy features, which are not essential for solving the target task and are even imperceptible to a human, thereby resulting in poor generalization. 
Leveraging plenty of undesirable features in OOD examples has emerged as a potential solution for de-biasing such features, and a recent study shows that softmax-level calibration of OOD examples can successfully remove the contribution of undesirable features to the last fully connected layer of a classifier. 
However, its applicability is confined to the classification task, and its impact on a DNN feature extractor is not properly investigated. In this part, we propose \algnamea{}, a novel regularizer that deactivates many undesirable features using OOD examples in the feature extraction layer and thus removes the dependency on the task-specific softmax layer. To show the task-agnostic nature of \algnamea{}, we rigorously validate its performance on three tasks, classification, regression, and a mix of them.

\textbf{Part 2: Robust Active Learning.}\hspace{0.1cm} 
Next, we introduce a robust active learning approach for unlabeled examples that prioritizes informative unlabeled examples over noisy examples in order to reduce the labeling cost. 
Unlabeled data examples awaiting annotations contain unlabeled noisy examples inevitably. A few active learning studies have attempted to deal with this unlabeled noise for sample selection by filtering out the noisy examples. 
However, because focusing on the purity of examples in a query set leads to overlooking the informativeness of the examples, the best balancing of purity and informativeness remains an important question. In this part, to solve this \emph{purity-informativeness dilemma} in active learning under noise, we propose a novel \emph{Meta-Query-Net\,(\algname{})} that adaptively finds the best balancing between the two factors. 
Specifically, by leveraging the multi-round property of active learning, we train \algname{} using a query set without an additional validation set. Furthermore, a clear dominance relationship between unlabeled examples is effectively captured by \algname{} through a novel \emph{skyline} regularization. Extensive experiments on multiple noisy active learning scenarios demonstrate that the proposed \algname{} significantly outperforms the state-of-the-art methods.

\textbf{Part 3: Robust Data Pruning.}\hspace{0.1cm}
Last, we suggest a robust data pruning approach called \algnamec{} that prioritizes informative labeled examples that can maximally maintain the target model performance trained on the reduced subset.
When utilizing state-of-the-art \emph{Re-labeling} methods that self-correct erroneous labels and reuse them for training, it is challenging to identify which subset induces the most accurate re-labeling of erroneous labels in the entire training set.
In this part, we formalize the problem of \emph{data pruning with Re-labeling}. We first show that the likelihood of a training example being correctly re-labeled is proportional to the prediction confidence of its neighborhood in the subset. Therefore, we plan to propose a novel data pruning algorithm that finds a subset such that the total neighborhood confidence of the entire training examples is maximized, thereby maximizing the re-labeling accuracy and generalization performance.
Experimental evaluations demonstrate the substantial superiority of \algnamec{} compared to existing pruning methods in the presence of label noise.
In addition, we extend this idea to noisy text data by prioritizing informative labeled instruction examples to enhance the performance of fine-tuned large language models.
To do so, we present a new instruction selection algorithm called \algnamed{} that ensures the cleanness, diversity, and quality of the selected subset by leveraging a cluster-wise prompting technique with a teacher LLM.

\begin{table*}[t!]
\def\arraystretch{1.15}
\centering
\caption{Dissertation outline.}
\vspace{-0.1cm}
\begin{tabular}{c |c |c } \toprule
& Input Noise ($X$ noise) & Label Noise ($Y$ noise) \\\midrule
{\makecell[c]{Labeled\\Data}}
& {\makecell[c]{(\textbf{Part 1}) \emph{published at NeurIPS'21}\\Feature learning with extra OOD datasets\\Informative features > Noisy features}}
& {\makecell[c]{(\textbf{Part 3}) \emph{published at NeurIPS'23}\\Data pruning under labeled noisy data\\Informative examples > Noisy examples}}  
\\ \cline{1-3}
{\makecell[c]{Unlabeled\\Data}}
& {\makecell[c]{(\textbf{Part 2}) \emph{published at NeurIPS'22} \\ Active learning under unlabeled noisy data \\ Informative examples > Noisy examples}}
& N/A 
\\ \bottomrule
\end{tabular}
\label{table:outline}
\vspace{-0.2cm}
\end{table*}


\section{Outline}

In this dissertation, we propose a systemic framework that \emph{prioritize informative features and examples} to enhance each stage of the development process from feature learning (Part 1), active learning (Part 2), to data pruning (Part 3).
In Table \ref{table:outline}, we summarize the three parts with respect to the types of noise and the types of data we use.
In Chapter 2, we provide a comprehensive literature review on the negative influence of noisy features and examples throughout the model development process.
In Chapter 3, we propose an approach for feature learning (Part 1) that prioritizes informative features over noisy features to induce the training of DNNs being more robust to noisy features. 
In Chapter 4, we introduce an approach for active learning (Part 2) that prioritizes informative unlabeled examples over noisy examples in order to reduce the labeling cost. 
In Chapter 5, we present an approach for data pruning (Part 3) that prioritizes informative examples from labeled noisy data to preserve the performance of the model trained on a reduced dataset, and in Chapter 6, we suggest a new approach for instruction selection (extension of Part 3) that prioritizes informative text examples from labeled noisy text data to preserve the performance of the large language models trained on a selected dataset.

\chapter{Background and Related Work}
\label{sec:overall_related_work}

\section{Prioritizing Informative Features}
\label{sec:related_work_paper1}

\vspace*{0.1cm}
\subsection{Effects of Informative Features and Noisy Features on DNNs}
\label{sec:effect_informative_features}

DNNs tend to overly capture all available signals from training data even when they are not essential for solving a given task\,\cite{wang2020high, ilyas2019adversarial}. 
The occurrence of undesirable features and their negative impact have been recently witnessed in various types of learning tasks.
In image classification, a classification model often uses background or texture features as an undesirable shortcut for making a prediction instead of using the intrinsic shape of a target class\,\cite{ilyas2019adversarial,geirhos2018imagenet}.
In object detection, a detector model easily overfits the background features for localizing target objects in a scene\,\cite{oksuz2020imbalance, singh2018analysis}.
In video action recognition, a recognition model often relies on static cues in a single frame rather than temporal actions over consecutive frames\,\cite{weinzaepfel2021mimetics, li2019repair}.
In natural language processing\,(NLP) tasks, a language model often makes its predictions based on frequent but meaningless words instead of using semantically meaningful words\,\cite{gururangan2018annotation}.

\vspace*{0.1cm}
\subsection{Connection with Adversarial Features}
\label{sec:connect_with_adversarial_features}

DNNs are easily deceived by adversarial perturbations of the inputs, so-called adversarial examples\,\cite{yuan2019adversarial}. Differently from standard learning, the undesirable features are maliciously added and then make the model incur more errors. In addition, it is widely known that such adversarial perturbations are transferable even from different domains\,\cite{naseer2019cross}; that is, an adversarial attack can drastically degrade the generalization capability of the classifier without knowing its internals\,\cite{naseer2019cross}. To remedy this problem, the use of OOD examples has gained great attention in that they enhance the robustness against the adversarial examples by preventing the model from overfitting to the undesirable features\,\cite{lee2021removing}. 

\vspace*{0.1cm}
\subsection{Removing Noisy Feature Contribution}
\label{sec:removing_noisy_feature}

Numerous studies have attempted to prevent overfitting to the undesirable features in standard supervised learning tasks.
A typical way is \emph{de-biasing}, which removes the undesirable feature contribution based on the pre-defined bias for the target task.
Geirhos et al.\,\cite{geirhos2018imagenet} took advantage of data augmentation techniques to generate de-biased examples from training data. 
Lee et al.\,\cite{lee2019srm} and Shetty et al.\,\cite{shetty2019not} synthesized de-biased examples by leveraging a generative model for image stylization or object removal. 
Wang et al.\,\cite{wang2018learning} quantified the local feature bias by using the neural gray-level co-occurrence matrix. 
Bahng et al.\,\cite{bahng2020learning} proposed a framework that leverages a bias-characterizing model to remove pixel-level local undesirable features. 
This family of methods successfully removes the \emph{pre-defined} bias from the undesirable features but is not generalizable to other types of bias. Even worse, it is hard to identify the types of undesirable features in advance since they are not comprehensible even to a human.

\vspace*{0.1cm}
\subsection{Extracting Informative Features from Out-of-distribution Data}
\label{sec:extracting_informative_features_from_OOD}

Motivated by the transferability of undesirable features in different domains, the usefulness of OOD examples for de-biasing started to be discussed. 
OAT\,\cite{lee2021removing} shows that the undesirable features can be successfully reduced by regularizing all the predictions of OOD examples to be the uniform distribution.
Although the representative softmax calibrator, OAT\,\cite{lee2021removing}, does not need a pre-defined bias type, it suffers from two limitations, lack of flexibility and feature entanglement. Many aspects, such as high generalizability and theoretical analysis, are yet to be explored.

\section{Prioritizing Informative Examples from Unlabeled Noisy Data}
\label{sec:related_work_paper2}

\vspace*{0.1cm}
\subsection{Active Learning}
\label{sec:active_learning}

{Active Learning} (AL) is a learning framework to reduce the human labeling cost by finding the most {informative} examples given unlabeled data\,\cite{ren2021survey, park2022active}.
Numerous active learning scores for measuring the informativeness of examples without given the ground-truth labels have been proposed\,\cite{ren2021survey}.
One popular direction is uncertainty-based sampling.
Typical approaches have exploited prediction probability, \textit{e.g.}, soft-max confidence\,\cite{lewis1994heterogeneous, wang2014active}, margin\,\cite{roth2006margin}, and entropy\,\cite{joshi2009multi}.
Some approaches obtain uncertainty by Monte Carlo Dropout on multiple forwards passes\,\cite{gal2017deep, kirsch2019batchbald, gal2016dropout}.
LL\,\cite{yoo2019learning} predicts the loss of examples by jointly learning a loss prediction module with a target model.
Meanwhile, diversity-based sampling has also been widely studied.
To incorporate diversity, most methods use a clustering\,\cite{nguyen2004active} or coreset selection algorithm\,\cite{sener2018active}.
Notably, CoreSet\,\cite{sener2018active} finds the set of examples having the highest distance coverage on the entire unlabeled data.
BADGE\,\cite{Ash2020badge} is a hybrid of uncertainty- and diversity-based sampling which uses $k$-means{++} clustering in the gradient embedding space.
Margin-Cluster\,\cite{citovsky2021batch} scales up the inefficient uncertainty- and diversity-based sampling with a heuristic rule based on a conventional hierarchical clustering.
This work also shows that the proposed efficient sample selection algorithm works well on active learning for the multi-class classification task.
However, this family of approaches is not appropriate for open-set active learning since they do not consider how to handle such useless OOD examples for query selection.

\subsection{Open-set Recognition}
\label{sec:openset_recognition}

{Open-set Recognition (OSR)} is a detection task to recognize the examples outside of the target domain\,\cite{geng2020recent}. Closely related to this purpose, OOD detection has been actively studied\,\cite{yang2021generalized}. 
Recent work can be categorized into classifier-dependent, density-based, and self-supervised approaches.
The classifier-dependent approach leverages a pre-trained classifier and introduces several scoring functions, such as Uncertainty\,\cite{MSP}, ODIN\,\cite{ODIN}, Mahalanobis distance\,(MD)\,\cite{MD}, and Energy\cite{ENERGY}.
Recently, ReAct\,\cite{sun2021react} shows that rectifying penultimate activations can enhance most of the aforementioned classifier-dependent OOD scores.
In detail, Uncertainty\,\cite{MSP} first shows that the prediction uncertainty can be a simple baseline for OOD detection.
ODIN\,\cite{ODIN} further enhances the uncertainty-based OOD detection by injecting an adversarial noise.
Mahalanobis distance\,(MD)\,\cite{MD} shows the Mahalanobis distance in the embedding space is more robust to OOD detection than the uncertainty calibration in the final prediction layer.
Energy\cite{ENERGY} theoretically proves the energy score is a more suitable measure than uncertainty for OOD detection.
The density-based approach learns an auxiliary generative model like a variational auto-encoder to compute likelihood-based OOD scores\,\cite{, ren2019likelihood, serra2019input, DGM}.
Most self-supervised approaches leverage contrastive learning\,\cite{tack2020csi, winkens2020contrastive, sehwag2021ssd}.
CSI shows that contrasting with distributionally-shifted augmentations can considerably enhance the OSR performance\,\cite{tack2020csi}.

\vspace*{0.1cm}
The OSR performance of classifier-dependent approaches degrades significantly if the classifier performs poorly\,\cite{vaze2021open}. Similarly, the performance of density-based and self-supervised approaches heavily resorts to the amount of clean IN data\,\cite{DGM, tack2020csi}.
Therefore, open-set active learning is a challenging problem to be resolved by simply applying the OSR approaches since it is difficult to obtain high-quality classifiers and sufficient IN data at early AL rounds.

\vspace*{0.1cm}
\subsection{Open-set Active learning}
\label{sec:open-active_learning}

Two recent approaches have attempted to handle the open-set noise for AL\,\cite{du2022conal, kothawade2021similar}.
Both approaches try to increase purity in query selection by effectively filtering out the OOD examples.
CCAL\,\cite{du2022conal} learns two contrastive coding models each for calculating informativeness and OODness of an example, and  combines the two scores using a heuristic balancing rule.
SIMILAR\,\cite{kothawade2021similar} selects a pure and core set of examples that maximize the distance on the entire unlabeled data while minimizing the distance to the identified OOD data.
However, we found that CCAL and SIMILAR are often worse than standard AL methods since they always put higher weights on purity although informativeness should be emphasized when the open-set noise ratio is small or in later AL rounds.
This calls for developing a new solution to carefully find the best balance between purity and informativeness.

\section{Prioritizing Informative Examples from Labeled Noisy Data}
\label{sec:related_work_paper3}

\vspace*{0.1cm}
\subsection{Robust Learning under Noisy Labels}
\label{sec:robust_learning_under_noisy_labels}

A long line of literature has been proposed to improve the robustness of DNNs against label noise---refer to \cite{song2022learning} for a detailed survey for deep learning with noisy labels.
Some studies have focused on modifying the architectures\,\cite{goldberger2017training, han2018masking, yao2018deep}.
In detail, the \emph{s-model} \cite{goldberger2017training} is similar to the \emph{dropout noise model} but dropout is not applied. The \emph{c-model} \cite{goldberger2017training} is an extension of the s-model that models the instance-dependent noise, which is more realistic than the symmetric and asymmetric noises.
\emph{Masking} \cite{han2018masking} is a human-assisted approach to convey the human cognition of invalid label transitions. Recently, the \emph{contrastive-additive noise network} \cite{yao2018deep} was proposed to adjust incorrectly estimated label transition probabilities by introducing a new concept of quality embedding, which models the trustworthiness of noisy labels.
Some studies have focused on modifying the loss functions\,\cite{ghosh2017robust, hendrycks2018using, ma2020normalized}.
In detail, the \emph{robust MAE} \cite{ghosh2017robust} showed that the mean absolute error\,(MAE) loss achieves better generalization than the CCE loss because only the MAE loss satisfies the aforementioned condition.
The \emph{active passive loss}\,(APL) \cite{ma2020normalized} is a combination of two types of robust loss functions, an active loss that maximizes the probability of belonging to the given class and a passive loss that minimizes the probability of belonging to other classes.
Other works have opted for sample selection approaches\,\cite{jiang2018mentornet, han2018co, wei2020combating, wu2020topological} that select as many clean examples as possible while discarding noisy examples based on some cleanness criterion, such as small-loss\,\cite{han2018co}. 
In \emph{MentorNet} \cite{jiang2018mentornet}, a pre-trained mentor network guides the training of a student network in a collaborative learning manner. Based on the small-loss trick, the mentor network provides the student network with examples whose labels are likely to be correct. \emph{Co-teaching} \cite{han2018co} also maintain two DNNs, but each DNN selects a certain number of small-loss examples and feeds them to its peer DNN for further training. \emph{Co-teaching+} further employs the disagreement strategy of \emph{Decouple} compared with \emph{Co-teaching}. {In contrast, \emph{JoCoR} \cite{wei2020combating} reduces the diversity of two networks via co-regularization, making predictions of the two networks closer.}
Note that, these works do not consider the compactness or efficiency of the selected sample set. 

\textbf{Re-labeling.}\hspace{0.1cm} 
Meanwhile, to further exploit even noisy examples for training, \emph{Re-labeling}\,\cite{song2019selfie, zhou2021robust} approaches try to correct noisy labels and use them for training with a re-labeling module, e.g., a heuristic rule\,\cite{song2019selfie}.
Notably, according to recent benchmark studies on real-world noisy datasets\,\cite{wei2021learning}, a family of Re-labeling methods with \emph{self-consistency regularization}\,\cite{xie2020unsupervised} has shown state-of-the-art performance.
In general, Re-labeling with a self-consistency regularizer is based on the following optimization form:
\begin{equation}
\mathcal{L}_{{Re\text{-}labeling}}(\tilde{\mathcal{D}}; \theta,\!\mathcal{A})=\sum_{(x,\tilde{y}) \in \tilde{\mathcal{D}}} \mathbbm{1}_{[{C}_{{\theta}}(x)\ge\tau]} \mathcal{L}_{ce}(x,\tilde{y};{\theta})+\lambda ~\!\sum_{x \in \tilde{\mathcal{D}}}\mathcal{L}_{reg}(x;{\theta},\!\mathcal{A}),
\label{eq:relabeling}
\end{equation}
where $\mathcal{\tilde{D}}=\{(x_i, \tilde{y}_i)\}_{i=1}^{m}$ is a given noisy training set obtained from a noisy joint distribution $\mathcal{X} \times \mathcal{\tilde{Y}}$, ${\theta}$ is a classifier, $\mathcal{A}$ is a strong data augmentation, $\mathcal{C}_{{\theta}}(\cdot)$ is a prediction confidence score, and $\tau$ is a threshold to identify confident (clean) examples for the supervised loss $\mathcal{L}_{ce}(x,\tilde{y};{\theta})$, i.e., cross-entropy. The noisy labels are implicitly corrected by the self-consistency loss $\mathcal{L}_{reg}(x;{\theta})$ exploiting the power of strong augmentations\,\cite{cubuk2020randaugment}.
DivideMix\,\cite{li2020dividemix}, ELR+\,\cite{liu2020early}, CORES\,\cite{cheng2020learning}, SOP+\,\cite{liu2022robust} are popular approaches belonging to this family.
DivideMix uses a co-training framework to further improve re-labeling accuracy and SOP+ introduces additional learnable variables combined with self-consistency loss.
For simplicity, we call this Re-labeling family with self-consistency regularization as ``Re-labeling'' throughout the paper.
Despite their effectiveness, Re-labeling tends to require more computation time due to additional data augmentations, multiple backbones, and longer training epochs, which raises the need to study a new data pruning approach for enhancing its efficiency.

\vspace*{0.1cm}
\subsection{Data Pruning}
\label{sec:data_pruning}

In order to achieve high generalization performance with a selected subset, general data pruning approaches often prioritize the selection of hard or uncertain examples. 
Specifically, \textit{uncertainty-based} methods\,\cite{wang2014active,roth2006margin,joshi2009multi}  favor the selection of lower-confidence examples over highly confident ones, as the former is assumed to be more informative than the latter.
Similarly, \textit{geometry-based} methods\,\cite{chen2012super, sener2018active} focus on removing redundant examples that are close to each other in the feature space, while \textit{loss-based} methods\,\cite{toneva2018empirical, paul2021deep, killamsetty2021grad, mirzasoleiman2020coresets} involve selecting the samples with high loss or gradient measured during training. 
In detail, Coleman et al.\,\cite{coleman2019selection} shows that the uncertainty-based scores, e.g., Confidence\,\cite{wang2014active}, Margin\,\cite{roth2006margin}, and Entropy\,\cite{joshi2009multi}, can be effective metrics for subset selection in that selecting lower confident examples is more helpful for model generalization than selecting higher confident ones.
Some works used the geometric distance in the feature space to avoid selecting examples with redundant information. Herding\,\cite{chen2012super} incrementally extends the selected coreset by greedily adding a data example that can minimize the distance between the center of the coreset and that of the original training set. kCenterGreedy\,\cite{sener2018active} selects $k$ examples that maximize the distance coverage on the entire unlabeled data.
Recently, many approaches try to directly exploit the components of deep learning with given ground-truth labels.
Forgetting\,\cite{toneva2018empirical} selects examples that are easy to be forgotten by the classifier, and it finds such samples by counting how frequently predicted label changes during several warm-up training epochs.
GraNd\,\cite{paul2021deep} uses the average norm of the gradient vector to measure the contribution of each example for minimizing the training loss.
GradMatch\,\cite{killamsetty2021grad} and CRAIG\,\cite{mirzasoleiman2020coresets} try to find an optimal coreset that its gradient can be matched with the gradient of the full training set.
Glister\,\cite{killamsetty2021glister} introduces a bi-level optimization framework that the outer loop for selecting the coreset which can be solved by a greedy algorithm.
Submodular functions, such as Graph Cut, Facility Location, and Log Determinant, which measure the diversity of information, have also been shown to be useful for data subset selection\,\cite{iyer2013submodular}. 
Meanwhile, some recent works reported that existing data pruning methods do not work well at high pruning ratios\,\cite{park2022active, zheng2022coverage}.
To alleviate this drawback, AL4DP\,\cite{park2022active} shows that mixing various levels of uncertain examples is better for data scarcity, Moderate\,\cite{xia2022moderate} aims to select examples with the distances close to the median, and CCS\,\cite{zheng2022coverage} proposes a coverage-based method that jointly considers data coverage with sample importance.
Refer to \,\cite{guo2022deepcore} for a detailed survey of data pruning for deep learning.

However, in realistic scenarios where there are noisy examples present, these existing methods may not always be applicable, because such samples may also exhibit high uncertainty and could potentially be considered informative for training\,\cite{toneva2018empirical}.
A few works attempted to improve the robustness of sample selection against label noise\,\cite{killamsetty2021glister}, but it requires an additional clean validation set which is difficult to obtain in practice. 
Also, to the best of our knowledge, there has been no work to consider the effect of data pruning on state-of-the-art noise-robust learners such as Re-labeling.

\section{Prioritizing Informative Examples from Labeled Text Noisy Data}
\label{sec:related_work_paper4}

\vspace*{0.1cm}
\subsection{Instruction Tuning for Large Language Models (LLMs)}
\label{sec:instruction_tuning_LLMs}

As scaling of transformer-based language models leads to a significant improvement in model capacity on many downstream tasks, many \emph{large language models (LLMs)} such as GPT-3\,\cite{brown2020language} and LLaMA\,\cite{touvron2023llama} have been proposed and demonstrated powerful performance for generating text on a wide range of natural language tasks\,\cite{zhang2023instruction}.
Despite their success, one major issue with LLMs in practice is the mismatch between their training objective and the user's objective.
That is, LLMs are mostly trained to just predict the \emph{next} word tokens on a large text corpus, while users want the models to understand their instructions accurately and to provide responses properly.
To fine-tune LLMs, early works use a collection of instruction-answer pair datasets constructed from public benchmark datasets in NLP tasks\,\cite{workshop2022bloom, raffel2020exploring, almazrouei2023falcon}.
To scale the instruction datasets, recent works generate instruction-answer pairs by leveraging proprietary models such as ChatGPT\,\cite{ouyang2022training}, and then fine-tune LLMs on the synthetic instruction datasets\,\cite{wang2022self, alpaca, chiang2023vicuna}.
Meanwhile, to further guide the LLMs to follow human preferences, some works directly incorporate human feedback into instruction-tuning by using reinforcement learning techniques such as proximal policy optimization\,\cite{ouyang2022training, iyer2022opt}.

Instruction datasets consisting of question-answer pairs from wide domains of NLP tasks are key to the success of instruction-following LLMs.
Many instruction datasets are being actively released and their size is growing exponentially.
Alpaca\,\cite{alpaca} and Unnatural Instruction\,\cite{honovich2022unnatural} datasets respectively contain 52K and 240K instruction examples generated from InstructGPT\,\cite{ouyang2022training}.
Natural Instructions\,\cite{honovich2022unnatural}, UnifiedQA\,\cite{khashabi2020unifiedqa}, Dolly\,\cite{conover2023free} datasets respectively contain 193K, 800K, and 15K instruction examples crafted by humans.
Some instruction datasets, such as Super-Natural Instructions\,\cite{wang2022super} and xP3\,\cite{muennighoff2022crosslingual}, support multi-lingual instructions.
To enhance LLM's generalization ability to many unseen tasks, the size of instruction datasets continues to increase containing a wide range of knowledge from many tasks.
With their exponentially increasing size, one main challenge is to handle the data quality issue; large-scale instruction collections inevitably contain uninformative, redundant, and noisy instructions which may degrade the accuracy and efficiency of instruction tuning.
This calls for developing approaches to reduce the size of the instruction dataset by selecting informative instructions and using only the reduced instructions for fine-tuning.

\vspace*{0.1cm}
\subsection{Instruction Selection}
\label{sec:instruction_selection}

To handle the data quality issue in instruction-tuning, some studies have been focused on instruction selection.
LIMA\,\cite{zhou2023lima} shows that an instruction dataset with 1K examples carefully crafted by humans can perform on par with decent aligned LLMs including LLaMA-65B trained on full Alpaca dataset for human preference test open-ended question benchmarks.
ALPAGASUS\,\cite{chen2023alpagasus} propose an automatic data selection strategy leveraging ChatGPT to score the quality of each instruction. With 9k high-quality data filtered from Alpaca-52k, LLaMA trained on 9k instructions outperforms that on the full dataset for the preference test.
InstructionMining\,\cite{cao2023instruction} uses natural language indicators to measure the data quality, and select informative subsets with a BlendSearch algorithm.
While these instruction selection approaches increase the performance of LLMs for the preference test, it remains a question that these selection approaches do not affect and decrease the factuality of LLMs.

\chapter{Prioritizing Informative Features for Model Training using Unlabeled Noisy Data}
\label{chap:part_1}
\section{Overview}
\label{sec:overview1}

\emph{Undesirable features}, which are informally defined as those not relevant to a target task, frequently appear in training data; for example, the background is an undesirable feature for classifying animals in images. The undesirable features are mainly caused by the statistical bias in \textit{in-distribution} training data. In fact, many undesirable features are statistically correlated with labels, even though they are unnecessary and sometimes even harmful for the target task\,\cite{bahng2020learning}; for example,  the ``desert'' background feature is correlated with ``camels'' because the camels frequently appear in a desert. However, such undesirable features\,(e.g., desert background) rather yield unreliable predictions because they are easily shifted in other unseen data\,(e.g., images of the camels on the road).

Meanwhile, deep neural networks\,(DNNs) are known to overly capture any high-frequency data components which are even imperceptible to a human\,\cite{wang2020high, ilyas2019adversarial}. This property is attributed to the vulnerability of DNNs that can totally overfit to random labels or adversarial examples owing to their extremely high capacity\,\cite{ilyas2019adversarial, zhang2016understanding, song2020learning, raghu2017expressive}. Accordingly, DNNs are easily biased toward the \textit{undesirable} features as well, thereby often showing unsatisfactory generalization to unseen examples\,\cite{geirhos2018imagenet}. Thus, it is very important to prevent overfitting to the undesirable features.

In this regard, a few research efforts have been devoted to remove the negative influence of undesirable features by leveraging \emph{out-of-distribution\,(OOD)} data\,\cite{lee2021removing, tsipras2019robustness}. Under the assumption that in-distribution and OOD data \emph{share} undesirable features, OOD data is treated as a useful resource to alleviate the aforementioned undesirable bias. Notably, a recent study\,\cite{lee2021removing} proposed a \emph{softmax-level} calibration, which assigns uniform softmax probabilities to all possible labels for all examples in OOD data. Although this approach shows decent de-biasing performance in the classification task, the softmax-level calibration has \emph{two} limitations:

\begin{itemize}[leftmargin=9pt, noitemsep]
\item{\textbf{Lack of Flexibility:} The softmax-level calibration is designated only for the classification task. However, the bias toward undesirable features occurs in numerous tasks, such as object localization and bounding box regression. Therefore, a flexible, task-agnostic approach is required to easily support other downstream tasks too.
}
\item{\textbf{Feature Entanglement:} Even desirable features can be entangled with undesirable ones by assigning the uniform softmax probability invariably to all possible labels for OOD examples. Thus, the negative influence of the undesirable features is not perfectly removed because they still remain and affect the activation of desirable features (See $\S$ \ref{sec:theoretic_analysis} for details).}
\end{itemize}

\begin{figure*}[t!]
\includegraphics[width=\linewidth]{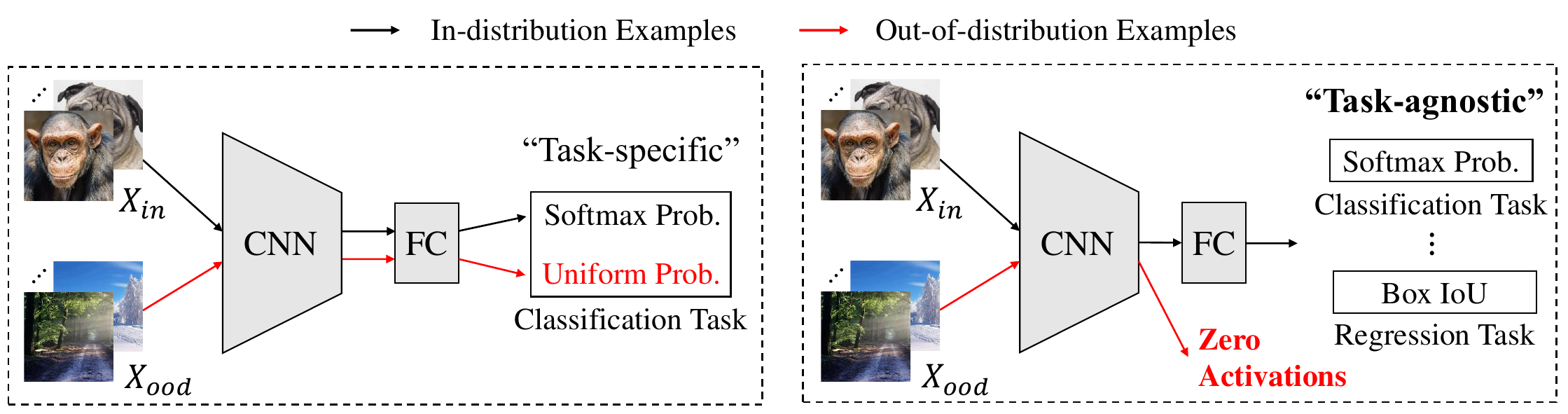}
\hspace*{1.3cm} \small{(a) Softmax-level calibration.} \hspace*{3cm} \small{(b) Feature-level calibration (\algnamea{}).}
\caption{Comparison of softmax-level and feature-level calibrations.}
\label{fig:intro_comparison}
\end{figure*}

In this chapter, we propose a novel \emph{task-agnostic} and \emph{feature-level} calibration method, called \algnamea{} (\underline{T}ask-\underline{A}gnostic \underline{U}ndesirable \underline{F}eature d\underline{E}activation), which explicitly forces a model to produce \emph{zero} values for many undesirable features in OOD examples. Differently from the softmax-level calibration that regularizes the \emph{classification} layer\,(Figure \ref{fig:intro_comparison}(a)), \algnamea{} exploits the \emph{penultimate} layer right before the classification layer and deactivates its activation only for OOD examples\,(Figure \ref{fig:intro_comparison}(b)). Thus, \algnamea{} is applicable to any task that requires another task-specific layer other than the classification layer, and the undesirable features are removed in the feature level without feature entanglement. The superiority of the proposed \emph{feature-level} calibration over the \emph{softmax-level} calibration is proven by theoretical and empirical analysis of the feature activation of the penultimate layer. 

To validate its general efficacy, we conducted extensive experiments through \emph{three} tasks: (i) image classification for classification; (ii) bounding box regression for regression; and (iii) weakly supervised object localization\,(WSOL) for a mix of them. We tested multiple pairs of in-distribution and OOD data: CIFAR-10, CIFAR-100, ImageNet, CUB200, and CAR for in-distribution; and SVHN, LSUN, and Places365 for OOD. The experiment results demonstrate that \algnamea{} consistently outperforms the softmax-level calibrator\,\cite{lee2021removing} by up to $9.88\%$ for classification and by up to $8.03\%$ for the mix of classification and regression.

Our main contributions are summarized as follows:
\begin{enumerate}[label={\arabic*.}, leftmargin=9pt] 
\item{We propose a simple yet effective method, \algnamea{}, to deactivate undesirable features in learning, which is easily applicable to any standard learning task with recent DNNs.}
\item{We provide an insight on how feature-level and softmax-level calibration differently affects feature extraction by theoretic and empirical analysis on the penultimate layer activation.}
\item{We validate the task-agnostic nature of \algnamea{} through three tasks and show its performance advantage over the state-of-the-art method.}
\end{enumerate} 
\section{Proposed Method: \algnamea{}}
\label{sec:OURS}

In this section, we first formulate the problem following the setup in the relevant literature\,\cite{ilyas2019adversarial, lee2021removing, tsipras2019robustness} and then describe our method \algnamea{}. Moreover, we provide a theoretical analysis with empirical evidence on how the softmax-level and feature-level calibrations work differently at the penultimate layer from the perspective of feature extraction.

\subsection{Problem Formulation}
\label{sec:problem_setup}


Let $\mathcal{D} = \{x_i,y_i\}_{i=1}^N$ be the target data obtained from a joint distribution over $\mathcal{X} \times \mathcal{Y}$, where $\mathcal{X}$ is the in-distribution example space and $\mathcal{Y}$ is the target label space. A DNN model consists of a general feature extractor $f_{\phi} : \mathcal{X} \rightarrow \mathcal{Z} \in \mathbb{R}^d$ and a task-specific layer $g_{\theta}: \mathcal{Z} \rightarrow \mathcal{Y}$. 
Then, the feature extractor is considered as a compound of $d$ sub-feature extractors $f_{\phi_{j}}$ such that $f_{\phi}(x) = \{ f_{\phi_{1}}(x), \dots, f_{\phi_{d}}(x) \}$ where $f_{\phi_{j}}: \mathcal{X} \rightarrow \mathbb{R}$.
A \emph{feature} is defined to be a function mapping from the example space $\mathcal{X}$ to a real number, and a set of the features is denoted by $\mathcal{F} = \{ f \in f_{\phi} : \mathcal{X} \rightarrow \mathbb{R} \}$.

We now formalize the desirableness of a feature. Let $\tilde{\mathcal{D}}=\{\tilde{x}_i\}_{i=1}^M$ be the \emph{out-of-distribution\,(OOD)} data obtained from a distribution over the OOD example space $\tilde{\mathcal{X}}$. Then, \emph{undesirable} and \emph{desirable} features are defined by Definitions \ref{def:undesirable} and \ref{def:desirable}, respectively.


\begin{definition}{\sc (Undesirable Feature)}.
For each example $\tilde{x}$ in the OOD data $\tilde{\mathcal{D}}$, we call a feature \emph{undesirable} if it is highly correlated with at least one true label in expectation. Thus, the set $\mathcal{F}_{undesirable}(\rho)$ of undesirable features is defined by
\begin{equation}
\mathcal{F}_{undesirable}(\rho) = \Big\{ f \in \mathcal{F} : \mathbb{E}_{\tilde{x}\in \tilde{\mathcal{D}}} \big[~ \text{max}_{y \in \mathcal{Y}} ~|\text{Corr}\big(f(\tilde{x}), y\big)|~ \big] \geq \rho \Big\},
\label{eq:undesirable_feature}
\end{equation}
where Corr is a function to produce the correlation between two given inputs\,(e.g., $R^2$) and $\rho$ is a constant threshold. $|~\cdot~|$ is an absolute value function to convert a negative correlation into a positive one. Intuitively speaking, an undesirable feature influences the model's decision-making even if it is not relevant to the target task\,(i.e., OOD examples).
\label{def:undesirable}
\end{definition}

\begin{definition} {\sc (Desirable Feature)}.
For each example $x$ and its corresponding label $y$ in the in-distribution data $\mathcal{D}$, we call a feature \emph{desirable} if it is highly correlated with the true label in expectation and does not belong to $\mathcal{F}_{undesirable}(\rho)$. Thus, the set $\mathcal{F}_{desirable}(\epsilon)$ of desirable features is defined by
\begin{equation}
\mathcal{F}_{desirable}(\epsilon) =  \Big\{ f \in \mathcal{F} / \mathcal{F}_{undesirable}(\rho) : \mathbb{E}_{{(x,y)}\in {\mathcal{D}}} \big[~ |\text{Corr}\big(f({x}), y\big)|~ \big] \geq \epsilon \Big\},
\label{eq:desirable_feature}
\end{equation}
where $\epsilon$ is a constant threshold; Corr and $\rho$ are the same as those for Definition \ref{def:undesirable}. 
\label{def:desirable}
\end{definition}

Note that Definitions \ref{def:undesirable} and \ref{def:desirable} are generally applicable to any supervised learning tasks including classification and regression.
By these definitions, a feature vector obtained by the feature extractor could be a mixture of desirable and undesirable features. DNNs can totally memorize even undesirable features owing to their high expressive power, leading to statistical bias in in-distribution training data. Therefore, the main challenge is to prevent the problem of biasing toward undesirable features, which will be discussed in the next section.

\subsection{Main Concept: Feature-Level Calibration}
\label{sec:method}

We introduce the notion of the \textit{feature-level} calibration, which directly manipulates the activations of the general feature extractor $f_{\phi}$. The key idea is to force the feature activations of all OOD examples to be zero vectors, thereby preventing the undesirable features from being carried over into the last task-specific layer $g_{\theta}$. 
Equation \ref{eq:objective} shows the difference in the objective function among standard learning, OAT\,(softmax-level calibration)\,\cite{lee2021removing}, and \algnamea{}\,(feature-level calibration): 
\begin{equation}
\begin{gathered}
\textsc{Standard:}~\mathop{\min}_{\phi, \theta}  \mathop{\mathbb{E}}_{(x,y)\in \mathcal{D}} \Big[\ell\Big(g_{\theta}\big(f_{\phi}(x)\big), y\Big)\Big],\\
\textsc{OAT:}~\mathop{\min}_{\phi, \theta}  \mathop{\mathbb{E}}_{(x,y)\in \mathcal{D}} \Big[\ell\Big(g_{\theta}\big(f_{\phi}(x)\big), y\Big)\Big] + \lambda \mathop{\mathbb{E}}_{\tilde{x}\in \tilde{\mathcal{D}}} \Big[ \ell\Big(g_{\theta}\big(f_{\phi}(x)\big), t_{\text{unif}}\Big) \Big],\\
\text{\algnamea{}:}~\mathop{\min}_{\phi, \theta}  \mathop{\mathbb{E}}_{(x,y)\in \mathcal{D}} \Big[\ell\Big(g_{\theta}\big(f_{\phi}(x)\big), y\Big)\Big] + \lambda \mathop{\mathbb{E}}_{\tilde{x}\in \tilde{\mathcal{D}}} \Big[||f_\phi(\tilde{x})||_2^2\Big],
\end{gathered}
\label{eq:objective}
\end{equation}
where $t_{\text{unif}} = [1/c, \dots, 1/c]$ and $\ell$ is the target loss for each original task\,(e.g., cross-entropy loss for classification or mean squared error\,(MSE) loss for regression). The first term is the same for all three methods, but there is a difference in the second term. Both OAT and \algnamea{} use the OOD examples\,(i.e., $\tilde{\mathcal{D}}$) to avoid the memorization of the undesirable features, but only \algnamea{} is not dependent on the task-specific layer $g_{\theta}$ in its regularization mechanism. Therefore, this feature-level calibration is easily applicable to any type of task for practical use and, at the same time, reduces the impact of undesirable features on the model's prediction. 

More importantly, \algnamea{} is remarkably simple. We contend that its simplicity should be a strong benefit because simple regularization often makes a huge impact and gains widespread use, as witnessed by weight decay and batch normalization. Algorithm \ref{alg:proposed_algorithm} describes the overall procedure of \algnamea{}, which is self-explanatory.

\algsetup{linenosize=\small}\newlength{\oldtextfloatsep}\setlength{\oldtextfloatsep}{\textfloatsep}
\setlength{\textfloatsep}{14pt}
\begin{algorithm}[!t]
\caption{\algnamea{}}
\label{alg:proposed_algorithm}
\begin{algorithmic}[1]
\REQUIRE{$\mathcal{D}$: target data, $\tilde{\mathcal{D}}$: OOD data, $epochs$: total number of epochs, $b$: batch size}
\ENSURE {$\phi_t, \theta_t$: network parameters}
\STATE $t \leftarrow 1; \phi_t, \theta_t \leftarrow$ Initialize the network parameters;
\STATE {{\bf for} $i=1$ {\bf to} $N$ {\bf do}}
\STATE ~~~~{{\bf for} $j=1$ {\bf to} $|{\mathcal{D}}|/b$ {\bf do}}
\STATE ~~~~~~~~{Draw a mini-batch ${\mathcal{B}}$ from ${\mathcal{D}}$;}~~\COMMENT{A target mini-batch.}
\STATE ~~~~~~~~{Draw a mini-batch $\tilde{\mathcal{B}}$ from $\tilde{\mathcal{D}}$;}~~\COMMENT{An OOD mini-batch.}
\STATE ~~~~~~~~{\COMMENT{Update for the feature extractor by the feature-level calibration.}}
\STATE ~~~~~~~~{$\phi_{t+1} = \phi_{t} -  \alpha \nabla_{\phi} \big(\mathop{\mathbb{E}}_{(x,y)\in \mathcal{B}} \big[\ell\big(g_{\theta_t}\big(f_{\phi_t}(x)\big), y\big)\big] + \lambda \mathop{\mathbb{E}}_{\tilde{x}\in \tilde{\mathcal{B}}} \big[||f_{\phi_t}(\tilde{x})||_2^2\big] \big)$}
\STATE ~~~~~~~~{\COMMENT{Update for the task-specific model by the standard manner.}}
\STATE ~~~~~~~~{$\theta_{t+1} = \theta_{t} - \alpha\nabla_{\theta} \mathop{\mathbb{E}}_{(x,y)\in \mathcal{B}} \big[\ell\big(g_{\theta_t}\big(f_{\phi_{t}}(x)\big), y\big)\big] $};
\STATE ~~~~~~~~{$t \leftarrow t+1;$}
\STATE {{\bf return} $\phi_{t}$, $\theta_{t}$}
\end{algorithmic}
\end{algorithm} 

\subsection{Theoretical and Empirical Analysis}
\label{sec:theoretic_analysis}

We analyze that the feature-level calibration works better than the softmax-level calibration in terms of feature disentanglement on the penultimate layer activations.
The use of OOD examples with the softmax-level calibration has been theoretically proven to remove undesirable feature contributions to the last linear classifier\,\cite{lee2021removing}.
However, the proof holds under the strong assumption that desirable and undesirable features should be disentangled perfectly before entering the last classifier layer. Because this assumption does not hold in practice, we provide an in-depth analysis on the use of OOD examples from the perspective of feature extraction without any assumption.

\textbf{Theoretic Analysis of Softmax-Level Calibration.}\hspace{0.1cm}
The effect of the softmax-level calibration is tightly related to label smoothing, which is a regularization technique\,\cite{muller2019does} that uses the target label combined with a uniform mixture over all possible labels. 
Let $z$ be the penultimate layer activation and $w_k$ be a weight row-vector of the last linear classifier assigned to the $k$-th class.
Then, the logit $z^{T}w_k$ for the $k$-th class can be thought of the negative  \textit{Euclidean distance} between $z$ and a weight template $w_k$, because $||z-w_k||^2=z^Tz+w^{T}_{k}w_k-2z^{T}w_k$ where $z^{T}z$ and $w^{T}_{k}w_k$ are usually constant across classes.
Therefore, when OAT assigns the uniform softmax probability to OOD examples, each logit $z^{T}w_k$ is forced into being the same value, which means that the penultimate layer activation $z$ is \textit{equally distant} to the templates\,(i.e., clusters) of all classes.

\begin{figure*}[t!]
\includegraphics[width=\linewidth]{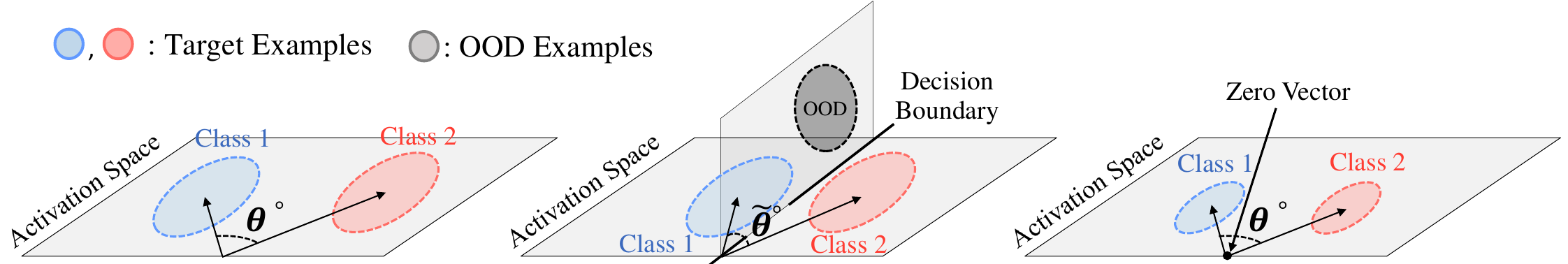}
\hspace*{1.3cm} \small{(a) Standard.} \hspace*{0.95cm} \small{(b) Softmax-level calibration\,(OAT).} \hspace*{0.20cm} \small{(c) Feature-level calibration\,(\algnamea{}).}
\caption{Effect of the softmax-level and feature-level calibrations on the penultimate layer activations.}
\label{fig:effect_on_penultimate}
\end{figure*}

As shown in Figure \ref{fig:effect_on_penultimate}(b), forcing all OOD examples into being equally distant to all class templates is mathematically equivalent to locating them on the \textrm{hyper-plane across the decision boundaries}. While the hyper-plane is orthogonal to the space composed of desirable features, it is likely onto a decision boundary. 
Accordingly, the undesirable features move the activations of the desirable features toward a decision boundary, and the two types of features are entangled.
Overall, although the softmax-level calibration helps remove undesirable features, it partially entangles the undesirable features with the desirable features, degrading the prediction performance.

\textbf{Theoretic Analysis of Feature-Level Calibration.}\hspace{0.1cm}
In contrast to the softmax-level calibration, the feature-level calibration explicitly forces the activations of all OOD examples into approaching the \emph{zero} vector\,\cite{park2020trap}, as shown in Figure~\ref{fig:effect_on_penultimate}(c).
This regularization reduces the norm of all target examples without changing the angle between the activations for different classes if they share undesirable features.
Since this angle plays a decisive role in classification\,\cite{chen2020addernet}, the feature-level calibration removes the undesirable features while effectively maintaining the disentanglement between desirable and undesirable features.
See Section \ref{sec:theoretic_analysis_appendix} for in-depth theoretical analysis.

\begin{table}
\small
\centering
\makeatletter\def\@captype{table}\makeatother
\caption{Average cosine similarity between all activation pairs across different classes on CIFAR-10 for Standard, OAT, and \algnamea{}.}
\begin{tabular}{c |c |c c c} \toprule
\multicolumn{2}{c|}{{Datasets}} & \multicolumn{3}{c}{Methods} \\\cline{1-2}\cline{3-5}\Tstrut
In-dist. & \!\!\!Out-of-dist.\!\!\! & \!\!\!Standard\!\!\! & OAT & \!\!\!\algnamea{}\!\!\! \\ \midrule
\!\!\!CIFAR-10\!\!\!  & LSUN & 0.116 & \!\!\!0.286\!\!\! & \!\!\!0.095\!\!\! \\ \bottomrule
\end{tabular}
\label{table:cosine_similarity}
\end{table}

\begin{figure*}[t!]
\centering
\includegraphics[width=\linewidth]{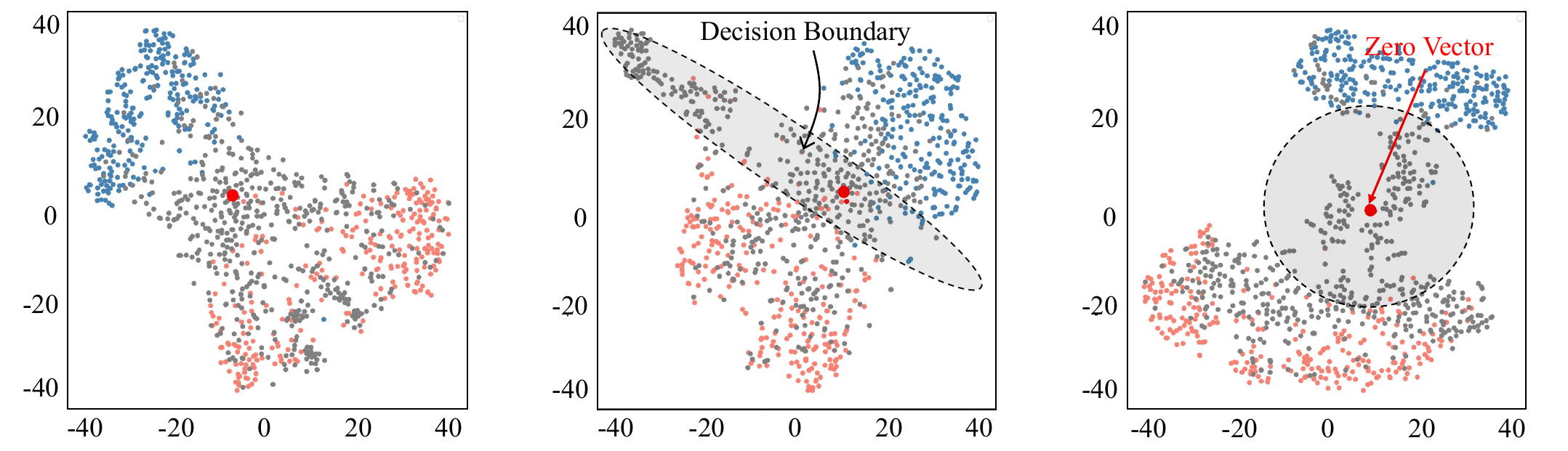} \\
\small{\hspace*{1.8cm}(a) Standard.} \hspace*{1.1cm} \small{(b) Softmax-level calibration\,(OAT).} \hspace*{0.05cm} \small{(c) Feature-level calibration\,(\algnamea{}).}
\caption{TSNE visualization of the penultimate layer activations. In-distribution examples are in pink for the automobile class and in blue for the bird class, while all OOD examples are in grey.}
\label{fig:tsne}
\end{figure*}

\textbf{Empirical Analysis.}\hspace{0.1cm}
To empirically support our analysis, in Table \ref{table:cosine_similarity}, we quantitatively calculate the cosine similarity of activations across all in-distribution classes. 
Compared with the standard learning method, \algnamea{}\,(feature-level calibration) decreases the cosine similarity between classes, whereas OAT\,(softmax-level calibration) rather increases the cosine similarity. That is, OAT is prone to move the activations of in-distribution examples toward the decision boundary, though it reduces the negative effect of the undesirable features on the classification task. In contrast, it is noteworthy that \algnamea{} renders the activations of different in-distribution classes more distinguishable.
Furthermore, we visualize the penultimate activations of the two in-distribution classes in CIFAR-10 together with those of OOD examples in Figure \ref{fig:tsne}. As shown in Figure \ref{fig:tsne}(b), OAT simply locates the activations of OOD examples around the decision boundary. However, \algnamea{} forces them into the zero vector without much change in the angles between different classes. Therefore, the empirical evidence confirms that \algnamea{} successfully reduces the negative effect of undesirable features on the classification task.

\subsection{In-Depth Theoretical Analysis}
\label{sec:theoretic_analysis_appendix}

Because a DNN extracts any type of feature if it is statistically correlated with the target label $y$, a feature vector $f_\phi$ of an in-distribution example $x$ contains both a desirable feature $f_{desirable}$ and an undesirable feature $f_{undesirable}$. That is, $f_\phi(x)=f_{desirable}(x)+f_{undesirable}(x)$, where $f\in \mathbb{R}^d$. On the other hand, because an OOD example $\tilde{x}$ does not contain any features that semantically indicate the target label $y$, $f_\phi(\tilde{x})=f_{undesirable}(\tilde{x})$.

For ease of exposition, let’s consider a binary classification setting. Let $x^{+}$ be an in-distribution example of the positive class and $x^{-}$ be an in-distribution example of the negative class. Then, via standard learning, $f_\phi(x^{+})=f_{desirable}(x^{+})+f_{undesirable}(x^{+})$ and $f_\phi(x^{-})=f_{desirable}(x^{-})+f_{undesirable}(x^{-})$. Because $f_{undesirable}(x^{+})$ and $f_{undesirable}(x^{-})$ are expected to share some features with $f_{undesirable}(\tilde{x})$, both OAT and TAUFE attempt to reduce their effect by the regularization on $f_\phi(\tilde{x})=f_{undesirable}(\tilde{x})$. Here, $f_\phi(x^{+})$ and $f_\phi(x^{-})$ correspond to the red and blue circles, respectively, in Figure \ref{fig:effect_on_penultimate}(a). 
For notational simplicity, we denote $f_\phi(x^{+})$ and $f_\phi(x^{-})$ as follows: 
\begin{equation}
\begin{gathered}
f_\phi(x^{+})=f_{desirable}^{+}+f_{undesirable}^{+}~~\text{and}~~
f_\phi(x^{-})=f_{desirable}^{-}+f_{undesirable}^{-}.
\end{gathered}
\label{eq:binary_classification_features}
\end{equation}

\textbf{OAT.}\hspace{0.1cm} As analyzed in Section \ref{sec:theoretic_analysis}, OAT regularizes the undesirable features from OOD examples being activated into the decision boundary. Thus, each class feature in Equation \ref{eq:binary_classification_features} is forced to be changed as follows:
\begin{equation}
\begin{gathered}
f_\phi^{OAT}(x^{+})=f_{desirable}^{+}+\left(\alpha \frac{(f_{desirable}^{+}+f_{desirable}^{-})}{2} + f_{\perp}^+ \right)~~\text{and}\\
f_\phi^{OAT}(x^{-})=f_{desirable}^{-}+\left(\beta \frac{(f_{desirable}^{+}+f_{desirable}^{-})}{2} + f_{\perp}^- \right),
\end{gathered}
\label{eq:binary_classification_features_OAT_1}
\end{equation}
where $\alpha, \beta \in \mathbb{R}$, $(f_{desirable}^{+}+f_{desirable}^{-})/2$ is a vector on the decision boundary, and $f_{\perp}$ is an orthogonal vector to the plane basis of $f_{desirable}^{+}$ and $f_{desirable}^{-}$. Then,
\begin{equation}
\begin{gathered}
f_\phi^{OAT}(x^{+})=\left(1+\frac{\alpha}{2}\right)f_{desirable}^{+}+\frac{\alpha}{2}f_{desirable}^{-} + f_{\perp}^+~~\text{and}\\
f_\phi^{OAT}(x^{-})=\frac{\beta}{2}f_{desirable}^{+}+\left(1+\frac{\beta}{2}\right)f_{desirable}^{-} + f_{\perp}^-.
\end{gathered}
\label{eq:binary_classification_features_OAT_2}
\end{equation}
Therefore, because the undesirable feature $f_{undesirable}$ moves the activation of the desirable feature toward the decision boundary, these two types of features (i.e., $f_{desirable}$ and $f_{undesirable}$) tend to be entangled, as illustrated in Figure \ref{fig:effect_on_penultimate}(b).

\textbf{\algnamea{}.}\hspace{0.1cm} As analyzed in Section \ref{sec:theoretic_analysis}, TAUFE regularizes the undesirable features from OOD examples being deactivated on the feature space (i.e., toward the zero vector). Thus, each class feature in Equation \eqref{eq:binary_classification_features} is forced to be changed as follows:
\begin{equation}
\begin{gathered}
f_\phi^{\algnamea{}}(x^{+})=f_{desirable}^{+}+\vec{0}~~\text{and}~~
f_\phi^{\algnamea{}}(x^{-})=f_{desirable}^{-}+\vec{0}.
\end{gathered}
\label{eq:binary_classification_features_TAUFE}
\end{equation}
Therefore, this regularization does not affect the activation of $f_{desirable}$, as illustrated in Figure \ref{fig:effect_on_penultimate}(c), thereby encouraging a prediction of a DNN to be solely based on the desirable features. This concludes the theoretical analysis of the novel L2 penalty term on \algnamea{}.

\section{Experiments}
\label{sec:experiments}

We compare \algnamea{} with the standard learning method\,(denoted as ``Standard'') and the state-of-the-art method OAT\,\cite{lee2021removing}. Standard  trains the network without any calibration process for OOD examples. In addition, we include the few-shot learning settings because DNNs are easily biased toward undesirable features especially when the number of training examples is small. All methods are implemented with PyTorch 1.8.0 and executed using four NVIDIA Tesla V100 GPUs. For reproducibility, we provide the source code at \url{https://github.com/kaist-dmlab/TAUFE}. 
In support of reliable evaluations, we repeat every test \emph{five} times and report the average. 

To show high flexibility in diverse types of tasks, we rigorously validate the efficacy of \algnamea{} for \emph{three} popular visual recognition tasks: (i) image classification, (ii) bounding box regression, and (iii) weakly supervised object localization\,(WSOL). Please note that OAT does not support the bounding box regression task because of the absence of the softmax layer.

\subsection{Task I: Image Classification}
\label{sec:image_classification}

\textbf{Dataset.}\hspace{0.1cm} We choose CIFAR-10, CIFAR-100\,\cite{krizhevsky2009learning}, and ImageNet\,\cite{deng2009imagenet} for the target in-distribution data. For the CIFAR datasets, two out-of-distribution datasets are carefully mixed for evaluation--- LSUN\,\cite{yu2015lsun}, a scene understanding dataset of 59M images with 10 classes such as bedroom and living room, and SVHN\,\cite{sermanet2012convolutional}, a real-world house numbers dataset of 70K images with 10 classes.
The ImageNet dataset is divided into 12K images of 10 randomly selected classes\,(ImageNet-10) and 1.1M images of the rest 990 classes\,(ImageNet-990); the former and the latter are used as in-distribution data and OOD data, respectively. A large-scale collection of place scene images with 365 classes, Places365\,\cite{zhou2017places}, is also used as another OOD data for ImageNet-10.

\textbf{Training Configuration.}\hspace{0.1cm} 
For CIFAR datasets, ResNet-18\,\cite{he2016deep} is trained from scratch for 200 epochs using SGD with a momentum of 0.9, a batch size of 64, and a weight decay of 0.0005. To support the original resolution, we drop the first pooling layer and change the first convolution layer with a kernel size of 3, a stride size of 1, and a padding size of 1. An initial learning rate of $0.1$ is decayed by a factor of $10$ at $100$-th and $150$-th epochs, following the same configuration in OAT\,\cite{lee2021removing}. For the ImageNet-10 dataset, ResNet-50 is used without any modification, but the resolution of ImageNet-10 is resized into 64$\times$64 and 224$\times$224 in order to see the effect of different resolutions. Resized random crops and random horizontal flips are applied for data augmentation. 

\algnamea{} requires only one additional hyperparameter, the scaling factor $\lambda$ for the feature-level calibration in Equation \ref{eq:objective}. The value of $\lambda$ is set to be $0.1$ and $0.01$ for CIFARs and ImageNet-10, respectively, where the best values are obtained via a grid search. The corresponding hyperparameter in OAT for softmax-level calibration is set to be $1$, following the original paper. In addition, both few-shot and full-shot learning settings are considered for evaluation.
Given the number $N$ of the examples for use in few-shot learning, $N$ examples are randomly sampled over all classes from both in-distribution and OOD data, and thus $2N$ examples in total are used for training. For full-shot learning, $N$ is set to be the total number of training examples in the target in-distribution data.

\begin{table}[t!]
\def\arraystretch{1.15}
\caption{Classification accuracy\,($\%$) of \algnamea{} compared with Standard and OAT on two CIFARs\,(32$\times$32), ImageNet-10\,(64$\times$64), and ImageNet-10\,(224$\times$224) under {few-shot} and {full-shot} learning settings. The highest values are marked in {bold}.} 

\begin{center}
\begin{tabular}{c|c|c|c c c c c}\toprule
\multicolumn{2}{c|}{{Datasets}} & \multirow{2}{*}{{Methods}} & \multicolumn{5}{c}{{\# Examples ($N$)}} \\\cline{1-2}\cline{4-8}\Tstrut
In-dist. & Out-of-dist. &  & 500 &  1,000 & 2,500  & 5,000 & Full-shot \\ \toprule
\multirow{5}{*}{\makecell[c]{CIFAR-10\\ (32$\times$32)}}
& -- & Standard & 38.58 & 52.63 & 72.94 & 82.38 & {94.22} \\ \cline{2-8}\Tstrut
& \multirow{2}{*}{SVHN} & OAT & 40.55 & 52.80 & 73.24 & 82.56 & {94.38} \\
& & \algnamea{} & 41.58 & 56.72 & 73.61 & 82.88 & {94.45} \\ \cline{2-8} \Tstrut
& \multirow{2}{*}{{LSUN}} & OAT & {40.73}& {53.16} & {73.51} & {82.71} & {94.61}\\
& & \algnamea{} & { \textbf{42.51} }& {\textbf{56.79}} & {\textbf{74.15}} & {\textbf{83.73}} & {\textbf{95.02}} \\ \midrule
\multirow{5}{*}{\makecell[c]{CIFAR-100\\ (32$\times$32)}}
& -- & Standard & 11.07 & 13.99 & 24.28 & 41.47 & {73.84} \\ \cline{2-8}\Tstrut
& \multirow{2}{*}{SVHN} & OAT & 10.92 & 14.56 & 24.67 & 42.21 & {74.82} \\
& & \algnamea{} & 11.30 & 15.13 & 24.91 & 43.61 & {75.38} \\ \cline{2-8}\Tstrut
& \multirow{2}{*}{{LSUN}} & OAT & {11.27} & {15.24} & {24.75} & {43.09} & {75.15} \\
& & \algnamea{} & {\textbf{12.26}} & {\textbf{15.97}} & {\textbf{25.36}} & {\textbf{44.50}} & {\textbf{75.69}} \\ \midrule
\multirow{5}{*}{\makecell[c]{\!\!\!ImageNet-10\!\!\!\\ (64$\times$64)}}
& -- & Standard & 38.82 & 43.66 & 56.17 & 66.80 & 78.30 \\ \cline{2-8}\Tstrut
& \multirow{2}{*}{\!\!\!ImageNet-990\!\!\!} & OAT & 38.95 & 44.09 & 57.29 & 69.41 & 79.29 \\
& & \algnamea{} & 42.80 & 46.04 & \textbf{60.40} & \textbf{70.51} & \textbf{81.09} \\ \cline{2-8}\Tstrut
& \multirow{2}{*}{Places365} & OAT & 41.06 & 43.81 & 56.47 & 67.20 & 79.30 \\
& & \algnamea{} & \textbf{43.25} & \textbf{47.61} & 60.02 & 68.25 & 80.89 \\ \midrule
\multirow{5}{*}{\makecell[c]{\!\!\!ImageNet-10\!\!\!\\ (224$\times$224)}}
& -- & Standard & 44.82 & 56.29 & 73.60 & 82.49 & 86.97 \\ \cline{2-8}\Tstrut
& \multirow{2}{*}{\!\!\!ImageNet-990\!\!\!} & OAT & 46.41 & 58.66 & 75.62 & 83.6 & 87.66 \\
& & \algnamea{} & 48.39 & 59.06 & 76.47 & \textbf{85.05} & \textbf{89.24} \\ \cline{2-8}\Tstrut
& \multirow{2}{*}{Places365} & OAT & 48.10 & 56.88 & 74.98 & 83.41 & 88.78 \\
& & \algnamea{} & \textbf{50.08} & \textbf{59.27} & \textbf{77.22} & 84.81 & 89.06 \\ \bottomrule
\end{tabular}
\end{center}
\label{table:classification_performance}
\end{table}

\textbf{Performance Comparison.}\hspace{0.1cm} Table~\ref{table:classification_performance} shows the classification accuracy of the three methods under few-shot and full-shot learning settings.
Overall, \algnamea{} shows the highest classification accuracy at any few-shot settings for all datasets. 
Specifically, \algnamea{} outperforms OAT by $0.07\%$ to $9.88\%$, though OAT also shows consistent performance improvement. OAT's lower performance is attributed to the property that it is prone to force the activations of in-distribution examples toward the decision boundary as analyzed in Section \ref{sec:theoretic_analysis}.
Adding {LSUN} as OOD for CIFARs is more effective than adding SVHN, {because LSUN is more similar to CIFARs than SVHN, thus sharing more undesirable features.}
For ImageNet-10, adding Places365 is more effective than adding ImageNet-990 when the number of training examples is not enough, but adding ImageNet-990 becomes more effective as the size of training data increases. 
Because ImageNet-990 has more diverse background scenes than Places365, we conjecture that the effect of Places365 saturates faster than that of ImageNet-990 as more OOD examples are exposed to the DNN model.
Besides, no significant difference is observed depending on the resolution of ImageNet-10.

\subsection{Task II: Bounding Box Regression}
\label{sec:box_regression} 

Bounding box regression is an essential sub-task for object localization and object detection. 
We compare \algnamea{} with only Standard because OAT does not work for regression.

\begin{table}[t!]
\def\arraystretch{1.15}
\caption{IoU\,($\%$) of \algnamea{} compared with Standard on CUB200\,(224$\times$224) and CAR\,(224$\times$224) under few-shot and full-shot learning settings. The highest values are marked in {bold}.}
\begin{center}
\begin{tabular}{c| c |c |c c c| c c c| c c c} \toprule
\multicolumn{2}{c|}{\multirow{2}{*}{Datasets}} & \multirow{3}{*}{{\!\!\!Methods\!\!\!}}& \multicolumn{3}{c|}{L1} & \multicolumn{3}{c|}{L1-IoU}  & \multicolumn{3}{c}{D-IoU} \\\cline{4-12}
\multicolumn{2}{c|}{} & & \multicolumn{3}{c|}{\makecell[c]{\# Examples ($N$)}} & \multicolumn{3}{c|}{\makecell[c]{\# Examples ($N$)}} & \multicolumn{3}{c}{\makecell[c]{\# Examples ($N$)}} \\ \cline{1-2}\cline{4-12}\Tstrut
In-dist. & \!Out-of-dist.\! &  & \!\!\!2,000\!\!\! &  \!\!\!4,000\!\!\! & \!\!\!Full\!\!\!  & \!\!\!2,000\!\!\! &  \!\!\!4,000\!\!\! & Full & \!\!\!2,000\!\!\! &  \!\!\!4,000\!\!\! & \!\!\!Full\!\!\! \\ \midrule
\multirow{3}{*}{\makecell[c]{CUB200\\ (224$\times$224)}}
& -- & \!\!\!Standard\!\!\! & \!\!66.41\!\! & \!\!73.10\!\! & \!\!76.42\!\! & \!\!66.57\!\! & \!\!73.28\!\! & \!\!76.67\!\! & \!\!66.82\!\! & \!\!73.18\!\! & \!\!76.57\!\! \\ \cline{2-12}\Tstrut
& \!\!\!{ImageNet}\!\!\! & \algnamea{} & \!\!\textbf{67.16}\!\! & \!\!\textbf{74.31}\!\! & \!\!\textbf{77.12}\!\! & \!\!\textbf{67.22}\!\! & \!\!\textbf{74.40}\!\! & \!\!\textbf{77.24}\!\! & \!\!\textbf{67.03}\!\! & \!\!\textbf{74.22}\!\! & \!\!\textbf{77.00}\!\! \\ \cline{2-12}\Tstrut
& \!\!\!{Places365}\!\!\! & \algnamea{} & \!\!66.70\!\! & \!\!73.55\!\! & \!\!76.86\!\! & \!\!66.87\!\! & \!\!73.63\!\! & \!\!77.01\!\! & \!\!66.88\!\! & \!\!73.66\!\! & \!\!76.88\!\! \\ \midrule
\multirow{3}{*}{\makecell[c]{CAR\\(224$\times$224)}}
& -- & \!\!\!Standard\!\!\! & \!\!83.06\!\! & \!\!85.50\!\! & \!\!90.56\!\! & \!\!83.52\!\! & \!\!86.54\!\! & \!\!91.25\!\! & \!\!83.62\!\! & \!\!87.93\!\! & \!\!91.09\!\!  \\ \cline{2-12}\Tstrut
& \!\!\!{ImageNet}\!\!\! & \algnamea{} & \!\!\textbf{85.23}\!\! & \!\!\textbf{87.82}\!\! & \!\!\textbf{91.32}\!\!&\!\!\textbf{85.82}\!\!& \!\!\textbf{89.11}\!\! & \!\!\textbf{91.40}\!\! &\!\!\textbf{85.30}\!\! & \!\!\textbf{89.06}\!\! & \!\!\textbf{91.35}\!\! \\ \cline{2-12}\Tstrut
& \!\!\!{Places365}\!\!\! & \algnamea{} & \!\!84.26\!\! & \!\!87.59\!\! & \!\!90.86\!\! & \!\!84.73\!\! & \!\!88.83\!\! & \!\!91.28\!\! &\!\!84.60\!\!& \!\!88.64\!\! & \!\!91.20\!\! \\ \bottomrule
\end{tabular}
\end{center}
\label{table:bbox_iou}
\end{table}

\textbf{Dataset.}\hspace{0.1cm} Two datasets are used as the target in-distribution data for the bounding box regression task---Caltech-UCSD Birds-200-2011\,(CUB200)\,\cite{wah2011caltech}, a collection of 6,033 bird images with $200$ classes, and Standford Cars\,(Car)\,\cite{krause20133d}, a collection of 8,144 car images with $196$ classes.
For each image of 224$\times$224 resolution, the two datasets contain a class label and bounding box coordinates of the top-left and bottom-right corners. ImageNet\footnote{All bird and vehicle relevant classes are excluded from the ImageNet dataset.} and Places365 are used as OOD data.

\textbf{Training Configuration.}\hspace{0.1cm} ResNet-50 is trained from scratch using SGD for 100 epochs. Following the prior work\,\cite{he2019bounding}, the last classification layer in ResNet-50 is converted to a box regressor that predicts the bounding box coordinates of the top-left and bottom-right corners. In addition, we use three types of different loss functions: (i) L1, a L1-smooth loss, (ii) L1-IoU, a combination of L1 and IoU, and (iii) D-IoU\,\cite{zheng2020distance}, a combination of L1, IoU, and the normalized distance between the predicted box and the target box. The remaining configurations are the same as those in $\S$ \ref{sec:image_classification}.

\textbf{Evaluation Metric.}\hspace{0.1cm} We adopt the Intersection over Union\,(IoU), which is the most widely-used metric for bounding box regression and defined by $\text{IoU}(b_i, \tilde{b}_i) = \frac{1}{k}\sum_{i=1}^{N}{|b_i \cap \tilde{b}_i|/|b_i\cup \tilde{b}_i}|$ where $b_i$ and $\tilde{b}_i$ are the ground-truth and predicted bounding boxes of the object in the $i$-th example.

\textbf{Performance Comparison.}\hspace{0.1cm} 
Table \ref{table:bbox_iou} shows the IoU accuracy of the two methods.
Overall, \algnamea{} consistently boosts the performance on bounding box regression for all datasets regardless of the loss type. Quantitatively, the box regression performance considerably improves with \algnamea{} by up to $2.97\%$ when using L1-IoU. This result indicates that the use of OOD examples with the feature-level calibration indeed alleviates the undesirable bias problem. 
Interestingly, adding ImageNet as OOD for both CUB200 and CAR is more effective than adding Places365, possibly because ImageNet contains a higher number of classes that reflect more diverse undesirable features.

\begin{table}[t!]
\caption{GT-known Loc of \algnamea{} compared with Standard on CUB200\,(224$\times$224) and CAR (224$\times$224) under few-shot and full-shot learning settings. The highest values are marked in {bold}.}
\begin{center}
\begin{tabular}{c |c |c |c c c} \toprule
\multicolumn{2}{c|}{{Datasets}} & \multirow{2}{*}{{Methods}} & \multicolumn{3}{c}{{\# Examples ($N$)}} \\\cline{1-2}\cline{4-6}\Tstrut
In-dist. & Out-of-dist. &  & 2,000 &  4,000 & Full-shot \\ \midrule
\multirow{5}{*}{\makecell[c]{CUB200\\ (224$\times$224)}}
& -- & Standard & 54.45 & 58.37 & 64.02 \\ \cline{2-6}\Tstrut
& \multirow{2}{*}{ImageNet} & OAT & 55.24 & 60.24 & 64.91 \\
& & \algnamea{} & \textbf{59.68} & \textbf{61.88} & \textbf{65.56} \\ \cline{2-6}\Tstrut 
& \multirow{2}{*}{Places365} & OAT & 56.97 & 60.01 & 64.27 \\
& & \algnamea{} & 58.24 & 60.90 & 64.84 \\ \midrule
\multirow{5}{*}{\makecell[c]{CAR\\ (32$\times$32)}}
& -- & Standard & 62.09 & 67.12 & 70.54 \\ \cline{2-6}\Tstrut
& \multirow{2}{*}{ImageNet} & OAT & 63.77 & 67.24 & 71.64 \\
& & \algnamea{} & \textbf{65.82} & \textbf{69.05} & \textbf{72.14} \\ \cline{2-6}\Tstrut
& \multirow{2}{*}{Places365} & OAT & 63.16 & 68.58 & 71.66 \\
& & \algnamea{} & 65.70 & 67.64 & 71.62 \\ \bottomrule
\end{tabular}
\end{center}
\label{table:GT_known_Loc}
\end{table}

\subsection{Task III: Weakly Supervised Object Localization (WSOL)}
\label{sec:wsol}

WSOL is a problem of localizing a salient foreground object in an image by using only weak supervision\,(i.e., image-level class labels). It can be considered as a mix of classification and regression because it uses class labels but aims at bounding box regression. The seminal WSOL work, class activation mapping\,(CAM)\,\cite{zhou2016learning}, has shown that the intermediate classifier activations focus on the most discriminative parts of the target object in the image. Thus, by simply averaging all local activations, we can estimate how much the corresponding pixels contribute to discriminating the object in the scene. CAM is used as the standard learning method.
Refer to the surveys\,\cite{zhang2021weakly} for more details about WSOL. 

\textbf{Dataset.}\hspace{0.1cm} Like the bounding box regression task, CUB200 and CAR are used as in-ditribution data, while Places365 and ImageNet are used as OOD data.

\textbf{Training Configuration.}\hspace{0.1cm} ResNet-50 is trained from scratch for 100 epochs using SGD with a batch size of 64. An initial learning rate of $0.1$ is decayed by a factor of 10 at the 50th and 75th epochs. 
The remaining configurations are the same as those in Section \ref{sec:image_classification}.

\textbf{Evaluation Metric.}\hspace{0.1cm} We adopt the localization accuracy
with  known ground-truth class\,(GT-known Loc), which is the most widely-used metric for WSOL and is defined by
$\text{GT\_known\_Loc}(b_i, \tilde{b}_i)=\frac{1}{N}\sum_{i=1}^{N}{\mathbb{1}\big({\text{IoU}(b_i, \tilde{b}_i)\geq \delta}\big)}$
where $b_i$ is the ground-truth box of the object in the $i$-th example and $\tilde{b}_i$ is the tightest box around the largest connected component of the activation mask for the $i$-th example.
The IoU threshold $\delta$ is set to be $0.5$, following the prior work\,\cite{zhou2016learning, zhang2021weakly}.

\textbf{Performance Comparison.}\hspace{0.1cm} 
Table~\ref{table:GT_known_Loc} shows the GT-known accuracy of the three methods.
Overall, \algnamea{} shows the best localization accuracy at any few-shot settings for all datasets.
Specifically, \algnamea{} outperforms OAT by $0.71\%$ to $8.03\%$, though OAT also shows consistent performance improvement.
This result indicates that \algnamea{} successfully removes undesirable features such as the background to locate an object in an image.
Adding ImageNet as OOD is more effective than adding Places365 for the same reason.
Besides, the performance gain of \algnamea{} over Standard is typically larger for few-shot learning than for full-shot learning, as observed in the other tasks.

\begin{table}[t!]
\def\arraystretch{1.15}
\caption{{Classification accuracy\,($\%$) of \algnamea{} under {few-shot} semi-supervised learning settings.}}
\centering
\begin{tabular}{c|c|c|c|c|c|c} \toprule
In-dist.&\multicolumn{3}{c|}{{CIFAR-10}}&\multicolumn{3}{c}{{CIFAR-100}}\\\midrule
Out-of-dist.& -- & SVHN & LSUN & -- & SVHN & LSUN\\ \midrule
Methods&MixMatch&$\text{\algnamea{}}_{\text{Mix}}$&$\text{\algnamea{}}_{\text{Mix}}$&MixMatch&$\text{\algnamea{}}_{\text{Mix}}$&$\text{\algnamea{}}_{\text{Mix}}$\\ \midrule
Accuracy& 88.32 & 90.02 & \textbf{90.10} & 51.38 & 52.32 & \textbf{52.58} \\ \bottomrule
\end{tabular}
\label{table:SSL_performance}
\end{table}

\subsection{Performance of \algnamea{} with Semi-Supervised Learning}
\label{sec:Performance_with_SSL}
We use a \emph{semi-supervised learning} framework for a baseline in addition to the standard supervised learning framework, because \algnamea{} can also improve the accuracy of a semi-supervised classifier. 

\textbf{Baseline.}\hspace{0.1cm} MixMatch\,\cite{berthelot2019mixmatch} is one of the state-of-the-art semi-supervised learning frameworks for image classification. By using unlabeled examples with automatic label guessing and mix-up, MixMatch nearly reaches the fully supervised learning accuracy with only a small number of labeled examples.

\textbf{Experiment Setting.}\hspace{0.1cm} CIFAR-10 and CIFAR-100 are used for in-distribution datasets, and LSUN and SVHN are used for two OOD datasets. We use the default or best hyperparameter values suggested by the authors\,\cite{berthelot2019mixmatch}. Specifically, the sharpening temperature $T$ is set to be 0.5, the number of augmentations $K$ to be 2, the Beta distribution parameter $\alpha$ to be 0.75, and the loss weight for unlabeled examples $\lambda_U$ to be 100. We fix the number of epochs to be 1,024 and the batch size to be 64, and linearly ramp up $\lambda_U$ in the first 16,000 optimization steps. We use 25 labeled examples per class as initially labeled data because MixMatch was shown to nearly reach the full-supervision accuracy on that setting\,\cite{berthelot2019mixmatch}.

\textbf{Result.}\hspace{0.1cm} Table \ref{table:SSL_performance} shows the classification accuracy of \algnamea{} combined with MixMatch on two CIFAR datasets under few-shot settings---i.e., $N\!\!=$250 for CIFAR-10 and $N\!\!=$2,500 for CIFAR-100; $\algnamea{}_{\text{Mix}}$ represents the \algnamea{} combined with MixMatch. $\algnamea{}_{\text{Mix}}$ consistently improves the performance of MixMatch on two CIFAR datasets. Similar to the supervised learning in Section \ref{sec:image_classification}, adding LSUN as OOD is more effective than adding SVHN; compared with MixMatch, the performance of $\algnamea{}_{\text{Mix}}$ is improved by up to $2.02\%$ on CIFAR-10 and by up to $2.34\%$ on CIFAR-100. This result shows that \algnamea{} successfully deactivates the negative effect of undesirable features even in the semi-supervised learning setting.

\subsection{Effect of \algnamea{} on Adversarial Robustness}
\label{sec:Effect_on_adversarial_robustness}

We further investigate the effect of \algnamea{} on \emph{adversarial robustness}, which is also known to highly rely on adversarial or undesirable features.

\textbf{Baseline.}\hspace{0.1cm} 
We use the projected gradient descent\,(PGD)\,\cite{madry2017towards} attack\,/\,learning method, which employs an iterative procedure of the fast gradient sign method\,(FGSM)\,\cite{goodfellow2014explaining} to find the worst-case examples having the maximum training loss.

\textbf{Experiment Setting.}\hspace{0.1cm} CIFAR-10 and CIFAR-100 are used for in-distribution datasets, and LSUN and SVHN are used for two OOD datasets which are exposed in the training phase. The hyperparameters of PGD are favorably set to be the best values reported in the original paper.
The attack learning rate $\epsilon$ is set to be 2, and $\text{PGD}_n$ indicate the PGD attacks with $n$ iterative FGSM procedures.
For adversarial learning, the adversarial examples generated by $\text{PGD}_7$ are used as the input of training.
To measure the adversarial accuracy, the adversarial examples generated by $\text{PGD}_{100}$ are used for testing.
This combination of step numbers was also used in the PGD work\,\cite{madry2017towards}.

\textbf{Evaluation Metric.}\hspace{0.1cm} The \emph{clean accuracy} is the classification accuracy on the original test data, while the \emph{adverserial accuracy} is that on the $\text{PGD}_{100}$ perturbed adversarial examples of the test data.

\textbf{Result.}\hspace{0.1cm} Table \ref{table:adversarial_robusteness} shows the adversarial robustness of \algnamea{} compared with the standard learning method. Overall, \algnamea{} improves the accuracy on adversarial examples by up to $2.76\%$ when adding the LSUN dataset as OOD examples. This result indicates that the undesirable feature deactivation of \algnamea{} is helpful for adversarial learning models.

\begin{table}[t!]
\def\arraystretch{1.15}
\centering
\caption{{Accuracy\,($\%$) of \algnamea{} under the PGD adversarial attacker.}} 
\begin{tabular}{c|c|c|c|c} \toprule
In-dist.&\multicolumn{4}{c}{{CIFAR-10}}\\\midrule
Out-of-dist.& -- & -- & SVHN & LSUN\\ \midrule
Methods&Standard&PGD&$\algnamea{}_{\text{PGD}}$&$\algnamea{}_{\text{PGD}}$\\ \midrule
\makecell{Clean.~acc.}&\makecell{94.22}&\makecell{71.22}&\makecell{72.35}&\makecell{72.31}\\ 
\makecell{Adv.~acc.}&\makecell{0.00}&\makecell{44.26}&\makecell{\textbf{44.37}}&\makecell{\textbf{45.48}}\\ \bottomrule
\end{tabular}
\label{table:adversarial_robusteness}
\end{table}

\begin{table}[t!]
\def\arraystretch{1.15}
\caption{{OOD detection performance\,($\%$) of \algnamea{} compared with Standard using uncertainty-based and energy-based OOD detection methods.}}
\vspace*{-0.2cm}
\begin{center}
\begin{tabular}{c |c |c |c | c |c | c | c} \toprule
\multicolumn{2}{c|}{{OOD detector}}&\multicolumn{3}{c|}{{Uncertainty}}&\multicolumn{3}{c}{{Energy}}\\\midrule
Dataset&Method&AUROC&\!\!AUPR$_{\text{out}}$\!\!&FPR95&AUROC&\!\!AUPR$_{\text{out}}$\!\!&FPR95\\\midrule
\!\!\multirow{2}{*}{CIFAR-10}\!\!&Standard&92.20&88.56&20.67&93.6&89.96&20.06\\ \cline{2-8}\Tstrut
&\algnamea{}&92.17&89.71&22.96&93.37&90.08&22.88\\ \midrule
\!\!\multirow{2}{*}{CIFAR-100}\!\!&Standard&83.31&79.37&45.76&88.46&86.04&35.35\\ \cline{2-8}\Tstrut
&\algnamea{}&82.03&79.21&47.89&88.42&88.56&37.82\\ \bottomrule
\end{tabular}
\end{center}
\label{table:OOD_detection}
\end{table}

\subsection{Effect of \algnamea{} on OOD detection}
\label{sec:Effect_on_OOD_detection}

We verify the effect of \algnamea{} on OOD detection, which aims at detecting out-of-distribution\,(OOD) examples in the test phase to support a trustworthy machine learning model. 

\textbf{Baseline.}\hspace{0.1cm} Numerous OOD detection methods have been proposed\,\cite{bulusu2020anomalous}. Here, we use two representative OOD detection methods---uncertainty-based\,\cite{hendrycks2016baseline} and energy-based\,\cite{liu2020energy}---to validate the effect of \algnamea{} on the OOD detection task.

\textbf{Experiment Setting.}\hspace{0.1cm} CIFAR-10 and CIFAR-100 are used for in-distribution datasets;  LSUN is used for exposing an OOD dataset in the training phase, and SVHN is used for measuring the detection performance in the test phase. The other training configurations are the same as those in Section \ref{sec:image_classification}.

\textbf{Evaluation Metric.}\hspace{0.1cm} The OOD detection performance is commonly quantified using three metrics\,\cite{hendrycks2016baseline, liu2020energy}. \emph{AUROC} is the area under the receiver operating characteristic, which is calculated by the area under the curve of the false positive rate\,(FPR) and the true positive rate\,(TPR). \emph{AUPR}$_\text{out}$ is the area under the curve of the precision and the recall, where they are calculated by considering OOD and in-distribution examples as positives and negatives, respectively. \emph{FPR95} is the FPR at $95\%$ of the TPR, which indicates the probability that an OOD example is misclassified as an in-distribution example when the TPR is $95\%$.

\textbf{Result.}\hspace{0.1cm} Table \ref{table:OOD_detection} shows the OOD detection performance of the two representative OOD detection methods without and with \algnamea{}. According to the three metrics, the performance with \algnamea{} is slightly higher than or just comparable to that without \algnamea{} in both OOD detection methods. Overall, as \algnamea{} is not geared for OOD detection, it does not significantly affect the OOD detection performance on two CIFAR datasets.

\subsection{Superiority of \algnamea{} over self-supervised learning}
\label{sec:superiority_over_ssl}

We compare \algnamea{} over the pre-training-fine-tuning approach, which performs pre-training on IN+OOD datasets and then performs fine-tuning on IN dataset, to solely validate whether the performance improvement of \algnamea{} genuinely comes from the regularization.

\textbf{Experiment Setting.}\hspace{0.1cm} 
We used CIFAR-10 as the IN dataset and SVHN as the OOD dataset. For the pre-training-fine-tuning approach, we first pre-train ResNet-18 by learning SimCLR\,\cite{chen2020simple} on both CIFAR-10 and SVHN datasets with the same training configurations in the original paper, and then fine-tune the last fully connected layer of the ResNet-18 on CIFAR-10 dataset only.

\textbf{Result.}\hspace{0.1cm} Table \ref{table:superiority_over_ssl} shows the performance superiority of \algnamea{} over the pre-training-fine-tuning approach. \algnamea{} consistently outperforms the standard classifier and the pre-training-fine-tuning approach throughout few- and full-shot settings, while the pre-training-fine-tuning approach shows worse performance in few-shot settings.

\begin{table}[t!]
\def\arraystretch{1.15}
\caption{Performance comparison between \algnamea{} and the pre-training-fine-tuning approach.} 
\begin{center}
\begin{tabular}{c|c|c|c c c c c}\toprule
\multicolumn{2}{c|}{{Datasets}} & \multirow{2}{*}{{Methods}} & \multicolumn{5}{c}{{\# Examples ($N$)}} \\\cline{1-2}\cline{4-8}\Tstrut
In-dist. & Out-of-dist. &  & 500 &  1,000 & 2,500  & 5,000 & Full-shot \\ \toprule
\multirow{3}{*}{\makecell[c]{CIFAR-10\\ (32$\times$32)}}
& -- & Standard & 38.58 & 52.63 & 72.94 & 82.38 & {94.22} \\ \cline{2-8}\Tstrut
& \multirow{2}{*}{SVHN} & SimCLR+Fine-tune & 30.45 & 47.70 & 71.66 & 82.36 & {94.40} \\
& & \algnamea{} & 41.58 & 56.72 & 73.61 & 82.88 & {94.45} \\ \bottomrule
\end{tabular}
\end{center}
\label{table:superiority_over_ssl}
\end{table}
\section{Conclusion and Future Work}
\label{sec:conclusion1}


In this chapter, we propose \algnamea{}, a novel \textit{task-agnostic} framework to reduce the bias toward undesirable features when training DNNs. Since the existing softmax-level calibration method confines its applicability to only the classification task, we overcome the limitation by introducing the \emph{feature-level} calibration that directly manipulates the feature output of a general feature extractor\,(e.g., a convolutional neural network). To remove the effect of undesirable features on the final task-specific module, \algnamea{} simply deactivates all undesirable features extracted from the OOD data by regularizing them as zero vectors. 
Moreover, we provide insight into how differently feature-level and softmax-level calibrations affect feature extraction by theoretic and empirical analysis of the penultimate layer activation.
The consistent performance improvement on three types of tasks clearly demonstrates the task-agnostic nature of \algnamea{}. 

Although \algname{} has shown consistent performance improvements in three types of real-world machine learning tasks, some issues need to be further discussed. 
First, the effectiveness of an OOD dataset for given a target dataset and a task needs to be formulated theoretically.
Owing to the transferability of undesirable features, any OOD dataset can be effective but its effectiveness varies as shown in Section \ref{sec:experiments}.
The difference in the effectiveness may come from the amount of shared undesirable features between the target dataset and each OOD dataset.
Therefore, formulating the effectiveness based on such factors is an interesting research direction.
Second, the applicability of \algname{} need to be verified for a wide range of learning frameworks including self-supervised learning, semi-supervised learning, and meta-learning, because the bias toward undesirable features is likely to be observed regardless of the learning frameworks.
Thus, we will clarify the outcome of \algname{} with varying learning frameworks as future work.

\chapter{Prioritizing Informative Examples for Active Learning from Unlabeled Noisy Data}
\label{chap:part_2}
\section{Overview}
\label{sec:overview2}

The success of deep learning in many complex tasks highly depends on the availability of massive data with well-annotated labels, which are very costly to obtain in practice\,\cite{song2022learning}.
\emph{Active learning\,(AL)} is one of the popular learning frameworks to reduce the high human-labeling cost, where a small number of maximally-informative examples are selected by a query strategy and labeled by an oracle repeatedly\,\cite{settles2009active}.
Numerous query\,(\textit{i.e.}, sample selection) strategies, mainly categorized into \emph{uncertainty}-based sampling\,\cite{wang2014active, Beluch2018Unc, yoo2019learning}  and \emph{diversity}-based sampling\,\cite{nguyen2004active, sener2018active, Ash2020badge}, have succeeded in effectively reducing the labeling cost while achieving high model performance.

Despite their success, most standard AL approaches rely on a strict assumption that all unlabeled examples should be cleanly collected from a pre-defined domain called \emph{in-distribution\,(IN)}, even before being labeled\,\cite{ren2021survey}.
This assumption is \emph{unrealistic} in practice since the unlabeled examples are mostly collected from rather \emph{casual} data curation processes such as web-crawling. 
Notably, in the Google search engine, the precision of image retrieval is reported to be 82$\%$ on average, and it is worsened to 48$\%$ for unpopular entities\,\cite{uyar2017investigating, cheshmehsohrabi2021performance}.
That is, such collected unlabeled data naturally involves \emph{open-set noise}, which is defined as a set of the examples collected from different domains called \emph{out-of-distribution\,(OOD)}\,\cite{geng2020recent}.

In general, standard AL approaches favor the examples either highly uncertain in predictions or highly diverse in representations as a query for labeling. However, the addition of open-set noise makes these two measures fail to identify informative examples; the OOD examples also exhibit high uncertainty and diversity because they share {neither} class-distinctive features {nor} other inductive biases with IN examples\,\cite{tseng2020cross, wei2021open}.
As a result, an active learner is confused and likely to query the OOD examples to a human-annotator for labeling.
Human annotators would disregard the OOD examples because they are unnecessary for the target task, thereby wasting the labeling budget.
Therefore, the problem of active learning with open-set noise, which we call \emph{open-set active learning}, has emerged as a new important challenge for real-world applications.

\begin{figure*}[t!]
\begin{center}
\includegraphics[width=\linewidth]{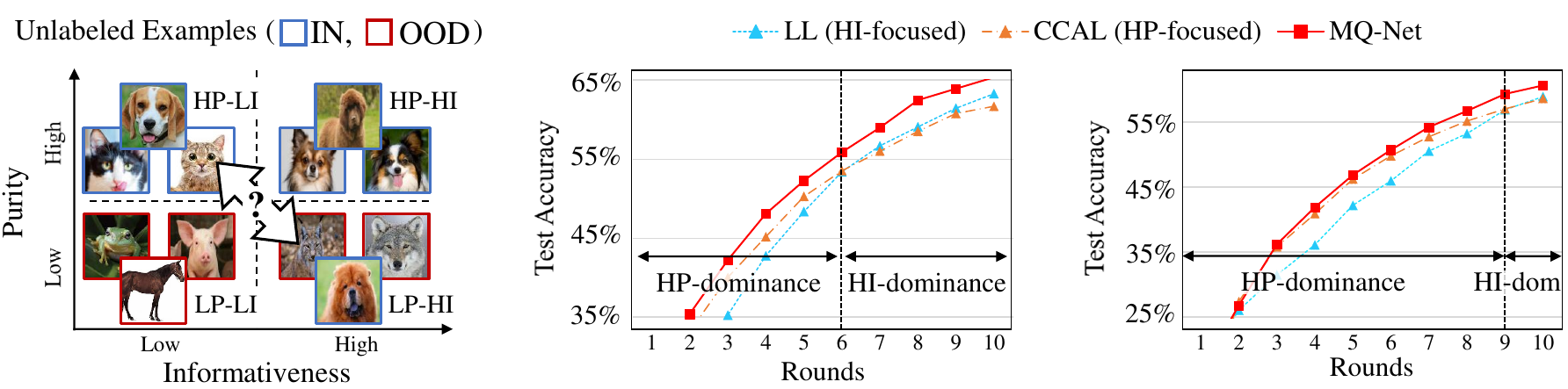}
\end{center}
\hspace*{0.cm} {\small (a) Purity-informativeness Dilemma.} \hspace*{0.85cm} {\small (b) $10\%$ Open-set Noise.} \hspace*{1.4cm} {\small (c) {$30\%$} Open-set Noise.}
\caption{
Motivation of \algname{}: (a) shows the purity-informativeness dilemma for query selection in open-set AL; (b) shows the AL performances of a standard AL method\,(HI-focused), LL\,\cite{yoo2019learning}, and an open-set AL method\,(HP-focused), CCAL\,\cite{du2022conal}, along with our proposed \algname{} for the ImageNet dataset with a noise ratio of $10\%$; (c) shows the trends with a noise ratio of {$30\%$}.
}
\label{fig:negative_effect_of_openset_noise}
\end{figure*}

Recently, a few studies have attempted to deal with the open-set noise for active learning\,\cite{du2022conal, kothawade2021similar}.
They commonly try to increase the purity of examples in a query set, which is defined as the proportion of IN examples, by effectively filtering out the OOD examples.
However, whether focusing on purity is needed \emph{throughout} the entire training period remains a question.
{In Figure \ref{fig:negative_effect_of_openset_noise}(a), let's consider an open-set AL task with a binary classification of cats and dogs, where the images of other animals, \emph{e.g.}, horses and wolves, are regarded as OOD examples.
It is clear that the group of high purity and high informativeness\,(HP-HI) is the most preferable for sample selection.
However, when comparing} the group of high purity and low informativeness\,(HP-LI) and that of low purity and high informativeness\,(LP-HI), the preference between these two groups of examples is \emph{not} clear, but rather contingent on the learning stage and the ratio of OOD examples. 
Thus, we coin a new term ``purity-informativeness dilemma'' to call attention to the best balancing of purity and informativeness. 

\newpage

Figures \ref{fig:negative_effect_of_openset_noise}(b) and \ref{fig:negative_effect_of_openset_noise}(c) illustrate the purity-informativeness dilemma. The standard AL approach, LL\cite{yoo2019learning}, puts more weight on the examples of high informativeness\,(denoted as HI-focused), while the existing open-set AL approach, CCAL\,\cite{du2022conal}, puts more weight on those of high purity\,(denoted as HP-focused). The HP-focused approach improves the test accuracy more significantly than the {HI}-focused one at earlier AL rounds, meaning that pure as well as easy examples are more beneficial. In contrast, the HI-focused approach beats the HP-focused one at later AL rounds, meaning that highly informative examples should be selected even at the expense of purity. Furthermore, comparing a low OOD\,(noise) ratio in Figure \ref{fig:negative_effect_of_openset_noise}(b) and a high OOD ratio in Figure \ref{fig:negative_effect_of_openset_noise}(c), the shift from HP-dominance to HI-dominance tends to occur later at a higher OOD ratio, which renders this dilemma more difficult.

In this chapter, to solve the purity-informativeness dilemma in open-set AL, we propose a novel meta-model \emph{Meta-Query-Net\,(\algname{})} that adaptively finds the best balancing between the two factors.
A key challenge is the best balancing is unknown in advance.
The meta-model is trained to assign higher priority for in-distribution examples over OOD examples as well as for more informative examples among in-distribution ones. The input to the meta-model, which includes the target and OOD labels, is obtained for free from each AL round's query set by the multi-round property of AL. Moreover, the meta-model is optimized more stably through a novel regularization inspired by the \emph{skyline} query\,\cite{kalyvas2017survey, deng2007multi} popularly used in multi-objective optimization.
As a result, \algname{} can guide the learning of the target model by providing the best balance between purity and informativeness throughout the entire training period.

Overall, our main contributions are summarized as follows:
\begin{enumerate}[label={\arabic*.}, leftmargin=9pt] 
\item{We formulate the \emph{purity-informativeness dilemma}, which hinders the usability of open-set AL in real-world applications.}
\item{As our answer to the dilemma, we propose a novel AL framework, \algname{}, which keeps finding the best trade-off between purity and informativeness.}
\item{Extensive experiments on CIFAR10, CIFAR100, and ImageNet show that \algname{} improves the classifier accuracy consistently when the OOD ratio changes from $10\%$ to $60\%$ by up to $20.14\%$.}
\end{enumerate}

\section{Purity-Informativeness Dilemma in Open-set Active Learning}
\label{sec:motivation}

In this section, we define the problem of open-set active learning and then introduce the purity-informativeness {dilemma} in query selection.

\subsection{Problem Statement: Open-set Active Learning}
\label{sec:problem_statement}

Let $\mathcal{D}_{IN}$ and $\mathcal{D}_{OOD}$ be the IN and OOD data distributions, where the label of examples from $\mathcal{D}_{OOD}$ does not belong to any of the $k$ known labels $Y=\{y_i\}_{i=1}^{k}$. Then, an unlabeled set is a mixture of IN and OOD examples, ${U}=\{{X}_{IN}, {X}_{OOD}\}$, \textit{i.e.}, ${X}_{IN} \sim \mathcal{D}_{IN}$ and ${X}_{OOD} \sim \mathcal{D}_{OOD}$. In the open-set AL, a human oracle is requested to assign a known label $y$ to an IN example $x \in {X}_{IN}$ with a labeling cost $c_{IN}$, while an OOD example $x \in {X}_{OOD}$ is marked as open-set noise with a labeling cost $c_{OOD}$.

AL imposes restrictions on the labeling budget $b$ every round. It starts with a small labeled set ${S}_{L}$, consisting of both labeled IN and OOD examples. The initial labeled set  ${S}_L$ improves by adding a small but maximally-informative labeled query set ${S}_{Q}$ per round, \emph{i.e.}, ${S}_L\!\leftarrow\!{S}_L\!\cup\!{S}_{Q}$, where the labeling cost for ${S}_{Q}$ by the oracle does not exceed the labeling budget $b$. Hence, the goal of open-set AL is defined to construct the optimal query set ${S}_{Q}^{*}$, minimizing the loss for the \emph{unseen} target IN data. 
The difference from standard AL is that the labeling cost for OOD examples is introduced, where the labeling budget is wasted when OOD examples are misclassified as informative ones. 

Formally, let $C(\cdot)$ be the labeling cost function for a given unlabeled set; then, each round of open-set AL is formulated to find the best query set ${S}_{Q}^{*}$ as
\begin{equation}
\begin{gathered}
{S}_Q^{*} = \argminB_{{S}_Q\!:\ C({S}_Q)\leq b} ~\mathbb{E}_{(x,y) \in {T}_{IN}} \Big[ \ell_{cls} \big( f(x; \Theta_{{S_L}\cup{S}_Q}), y \big) \Big], \\[-0.2pt]
\text{where}~~C({S}_Q)=\sum_{x \in {S}_Q} \big( \mathbbm{1}_{[x \in {X}_{IN}]} c_{IN}+ \mathbbm{1}_{[x \in {{X}_{OOD}}]} c_{OOD} \big).
\end{gathered}
\label{eq:single}
\end{equation}

Here, $f(\cdot;\Theta_{{S}_{L}\cup {S}_{Q}})$ denotes the target model trained on only IN examples in ${S}_{L}\cup{S}_{Q}$, and $\ell_{cls}$ is a certain loss function, \emph{e.g.}, cross-entropy, for classification. For each AL round, all the examples in ${S}_{Q}^{*}$ are removed from the unlabeled set ${U}$ and then added to the accumulated labeled set ${S}_{L}$ with their labels. This procedure repeats for the total number $r$ of rounds.

\subsection{Purity-Informativeness Dilemma}
\label{sec:purity_informativeness_dilemma}

An ideal approach for open-set AL would be to increase both the purity and informativeness of a query set by completely suppressing the selection of OOD examples and accurately querying the most informative examples among the remaining IN examples.
However, the ideal approach is infeasible because overly emphasizing purity in query selection does not promote example informativeness and \textit{vice versa}.
Specifically, OOD examples with low purity scores mostly exhibit high informativeness scores because they share neither class-distinctive features nor other inductive biases with the IN examples\,\cite{tseng2020cross, wei2021open}.
We call this trade-off in query selection the \emph{purity-informativeness dilemma}, which is our new finding expected to trigger a lot of subsequent work.

To address this dilemma, we need to consider the proper weights of a purity score and an informative score when they are combined.
Let $\mathcal{P}(x)$ be a purity score of an example $x$ which can be measured by any existing OOD scores, \textit{e.g.}, negative energy\,\cite{ENERGY}, and $\mathcal{I}(x)$ be an informativeness score of an example $x$ from any standard AL strategies, \textit{e.g.}, uncertainty\,\cite{ wang2014active} and diversity\,\cite{citovsky2021batch}. 
Next, supposing $z_{x}=\langle \mathcal{P}(x), \mathcal{I}(x) \rangle$ is a tuple of available purity and informativeness scores for an example $x$. Then, a score combination function $\Phi(z_{x})$, where $z_{x}=\langle \mathcal{P}(x), \mathcal{I}(x) \rangle$, is defined to return an overall score that indicates the necessity of $x$ being included in the query set.

Given two unlabeled examples $x_i$ and $x_j$, if $\mathcal{P}(x_i)>\mathcal{P}(x_j)$ and $\mathcal{I}(x_i)>\mathcal{I}(x_j)$, it is clear to favor $x_i$ over $x_j$ based on $\Phi(z_{x_i})>\Phi(z_{x_j})$. However, due to the purity-informativeness dilemma, if $\mathcal{P}(x_i)>\mathcal{P}(x_j)$ and $\mathcal{I}(x_i)<\mathcal{I}(x_j)$ or $\mathcal{P}(x_i)<\mathcal{P}(x_j)$ and $\mathcal{I}(x_i)>\mathcal{I}(x_j)$, it is very challenging to determine the dominance between $\Phi(z_{x_i})$ and $\Phi(z_{x_j})$.
In order to design $\Phi(\cdot)$, we mainly focus on leveraging \emph{meta-learning}, which is a more {agnostic} approach to resolve the dilemma other than several heuristic approaches, such as linear combination and multiplication.

\section{Meta-Query-Net}
\label{sec:method2}

\begin{figure}[t!]
\begin{center}
\includegraphics[width=5.5cm]{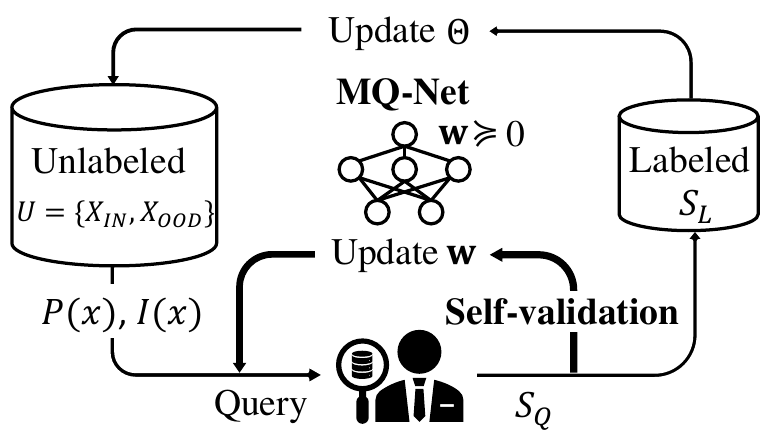}
\end{center}
\caption{Overview of \algname{}.}
\label{fig:method_overview}
\end{figure}

We propose a meta-model, named \emph{Meta-Query-Net\,(\algname{})}, which aims to learn a meta-score function for the purpose of identifying a query set.
In the presence of open-set noise, \algname{} outputs the meta-score for unlabeled examples to achieve the best balance between purity and informativeness in the selected query set.
In this section, we introduce the notion of a self-validation set to guide the meta-model in a supervised manner and then demonstrate the meta-objective of \algname{} for training. Then, we propose a novel skyline constraint used in optimization, which helps \algname{} capture the obvious preference among unlabeled examples when a clear dominance exists. Next, we present a way of converting the purity and informativeness scores estimated by existing methods for use in \algname{}. 
{Note that training \algname{} is \emph{not} expensive because it builds a light meta-model on a small self-validation set.}
The overview of \algname{} is illustrated in Figure \ref{fig:method_overview}.

\subsection{Training Objective with Self-validation Set}
\label{sec:objective}

The parameters $\textbf{w}$ contained in \algname{} $\Phi(\cdot;\textbf{w})$ is optimized in a supervised manner. For clean supervision, validation data is required for training. Without assuming a hard-to-obtain clean validation set, we propose to use a \emph{self-validation} set, which is instantaneously generated in every AL round. In detail, we obtain a labeled query set $S_Q$ by the oracle, consisting of a labeled IN set and an identified OOD set in every round. Since the query set $S_Q$ is unseen for the target model $\Theta$ and the meta-model $\textbf{w}$ at the current round, we can exploit it as a self-validation set to train \algname{}. This self-validation set eliminates the need for a clean validation set in meta-learning.

Given the ground-truth labels in the self-validation set, it is feasible to guide \algname{} to be trained to resolve the purity-informativeness dilemma by designing an appropriate meta-objective. It is based on {the} cross-entropy loss for classification because {the} loss value of training examples has been proven to be effective in identifying high informativeness examples\,\cite{yoo2019learning}. The conventional loss value by a target model $\Theta$ is masked to be \textit{zero} if $x \in X_{OOD}$ since OOD examples are useless for AL, 
\begin{equation}
\ell_{mce}(x) =  \mathbbm{1}_{[l_x = 1]}\ell_{ce}\big(f(x; \Theta), y\big),
\label{eq:mask_info}
\end{equation}
where $l$ is a true binary IN label, \emph{i.e.}, $1$ for IN examples, and $0$ for OOD examples, which can be reliably obtained from the self-validation set. This \emph{masked} loss, $\ell_{mce}$, preserves the informativeness of IN examples while excluding OOD examples. Given a self-validation data $S_Q$, the meta-objective is defined such that \algname{} parameterized by $\textbf{w}$ outputs a high\,(or low) meta-score $\Phi(z_{x}; \textbf{w})$ if an example $x$'s masked loss value is large\,(or small), 
\begin{equation}
\begin{gathered}
\!\!\!\!\!\!\mathcal{L}(S_Q) \!=\!\sum_{i\in S_Q} \sum_{j\in S_Q} \max\!\Big(0, - \text{Sign}\big(\ell_{mce}(x_i),\ell_{mce}(x_j)\big) \cdot \big(\Phi(z_{x_i}; \textbf{w})-\Phi(z_{x_j}; \textbf{w})+ \eta\big) \Big) \\
s.t. ~~\forall x_i, x_j,~~ \mbox{if} ~~ \mathcal{P}(x_i)>\mathcal{P}(x_j) ~~ \mbox{and} ~~ \mathcal{I}(x_i)>\mathcal{I}(x_j),~~ \mbox{then} ~~\Phi(z_{x_i}; \textbf{w})>\Phi(z_{x_j}; \textbf{w}),
\end{gathered}
\label{eq:ranking_loss}
\end{equation}
where $\eta>0$ is a constant margin for the ranking loss, and $\text{Sign}(a, b)$ is an indicator function that returns $+1$ if $a>b$, $0$ if $a=b$, and $-1$ otherwise. Hence, $\Phi(z_{x_i}; \textbf{w})$ is forced to be higher than $\Phi(z_{x_j}; \textbf{w})$ if $\ell_{mce}(x_i) > \ell_{mce}(x_j)$; in contrast, $\Phi(z_{x_i}; \textbf{w})$ is forced to be lower than $\Phi(z_{x_j}; \textbf{w})$ if $\ell_{mce}(x_i) < \ell_{mce}(x_j)$. Two OOD examples do not affect the optimization because they do not have any priority between them, \emph{i.e.}, $\ell_{mce}(x_i) = \ell_{mce}(x_j)$.

In addition to the ranking loss, we add a regularization term named the \emph{skyline} constraint (\emph{i.e.}, the second line) in the meta-objective Equation \ref{eq:ranking_loss}, which is inspired by the skyline query which aims to narrow down a search space in a large-scale database by keeping only those items that are not worse than any other\,\cite{kalyvas2017survey, deng2007multi}. 
Specifically, in the case of $\mathcal{P}(x_i)>\mathcal{P}(x_j)$ and $\mathcal{I}(x_i)>\mathcal{I}(x_j)$, the condition $\Phi(z_{x_i}; \textbf{w}) > \Phi(z_{x_j}; \textbf{w})$ must hold in our objective, and hence we make this proposition as the skyline constraint. This simple yet intuitive regularization is very helpful for achieving a meta-model that better judges the importance of purity or informativeness. 

\subsection{Architecture of \algname{}}
\label{sec:architecture}

\algname{} is parameterized by a multi-layer perceptron\,(MLP), a widely-used deep learning architecture for meta-learning\,\cite{shu2019meta}.
A challenge here is that the proposed skyline constraint in Equation \ref{eq:ranking_loss} does not hold with a standard MLP model. To satisfy the skyline constraint, the meta-score function $\Phi(\cdot; \textbf{w})$ should be a monotonic non-decreasing function because the output\,(meta-score) of \algname{} for an example $x_i$ must be higher than that for another example $x_j$ if the two factors\,(purity and informativeness) of $x_i$ are both higher than those of $x_j$.
The MLP model consists of multiple matrix multiplications with non-linear activation functions such as ReLU and Sigmoid. 
In order for the MLP model to be monotonically non-decreasing, all the parameters in $\textbf{w}$ for $\Phi(\cdot; \textbf{w})$ should be \emph{non-negative}, as proven by Theorem \ref{theorem:order_preserve}.

\smallskip\smallskip
\begin{theorem}
For any MLP meta-model $\textbf{w}$ {with non-decreasing activation functions}, a meta-score function $\Phi(z; \textbf{w})\!: \mathbb{R}^d \rightarrow \mathbb{R}$ holds the skyline constraints if $\textbf{w}\succeq0$ and $z (\in \mathbb{R}^d) \succeq 0$, where $\succeq$ is the component-wise inequality.
\label{theorem:order_preserve}
\begin{proof}
An MLP model is involved with matrix multiplication and composition with activation functions, which are characterized by three basic operators: \emph{(1) addition}: $h(z)=f(z)+g(z)$, \emph{(2) multiplication}: $h(z)=f(z) \times g(z)$, and \emph{(3) composition}: $h(z)=f\circ g(z)$. These three operators are guaranteed to be non-decreasing functions if the parameters of the MLP model are all non-negative because the non-negative weights guarantee all decomposed scalar operations in MLP to be non-decreasing functions. Combining the three operators, the MLP model $\Phi(z;\textbf{w})$, where $\textbf{w}\succeq0$, naturally becomes a monotonic non-decreasing function for each input dimension. Refer to Section \ref{sec:complete_proof} for the complete proof. 
\end{proof}
\end{theorem}

In the implementation,  non-negative weights are guaranteed  by applying a ReLU function to meta-model parameters. Since the ReLU function is differentiable, \algname{} can be trained with the proposed objective in an end-to-end manner. Putting this simple modification, the skyline constraint is preserved successfully without introducing any complex loss-based regularization term. The only remaining condition is that each input of \algname{} must be a vector of non-negative entries.

\subsection{Complete Proof} 
\label{sec:complete_proof}

Let $z_{x}=\{z_x^{\langle 1 \rangle}, \ldots, z_x^{\langle d \rangle}\}$ be the $d$-dimensional meta-input for an example $x$ consisting of $d$ available purity and informativeness scores.\footnote{We  use only two scores\,($d=2$) in \algname{}, one for purity and another for informativeness.}
A non-negative-weighted MLP $\Phi_{\textbf{w}}$ can be formulated as
\begin{equation}
h^{[l]}=\sigma\big( W^{[l]} \cdot h^{[l-1]} + b^{[l]} \big),~ l \in \{1,\ldots,L\},
\label{eq:mlp}
\end{equation}
where $h^{[0]}=z_{x}, h^{[L]}\in\mathbb{R}, W^{[l]}\succeq 0$, and $b^{[l]} \succeq 0$; $L$ is the number of layers and $\sigma$ is a non-linear non-decreasing activation function.

We prove Theorem~\ref{theorem:order_preserve} by mathematical induction, as follows: (1) the first layer's output satisfies the skyline constraint by Lemmas~\ref{lemma:skyline_layer_1} and \ref{lemma:skyline_non_linear}; and (2) the $k$-th layer's output ($k\geq2$) also satisfies the skyline constraint if the $(k-1)$-th layer's output satisfies the skyline constraint. 
Therefore, we conclude that the skyline constraint holds for any non-negative-weighted MLP $\Phi(z;\textbf{w})\!: \mathbb{R}^d \rightarrow \mathbb{R}$ by Theorem\,\ref{theorem:complete_skyline}.

\begin{lemma}
Let $g^{[1]}(z_{x})\!=\!W^{[1]}\!\cdot\!z_{x}\!+\!b^{[1]}$ be a non-negative-weighted single-layer MLP with $m$ hidden units and an identity activation function, where $W^{[1]}\!\in\mathbb{R}^{m\times d}\succeq 0$ and $b^{[1]}\in\mathbb{R}^m\succeq 0$. Given the meta-input of two different examples $z_{x_i}$ and $z_{x_j}$, the function $g^{[1]}(z_{x})$ satisfies the skyline constraint as 
\begin{equation}
z_{x_i} \succeq z_{x_j} \implies g^{[1]}(z_{x_i}) \succeq g^{[1]}(z_{x_j}).
\label{eq:skyline_layer_1}
\end{equation}
\begin{proof}
Let $g^{[1]}(z_{x})$ be $g(z_{x})$ and $W^{[1]}$ be $W$ for notation simplicity.
Consider each dimension's scalar output of $g(z_{x})$, and it is denoted as $g^{\langle p \rangle}(z_{x})$ where $p$ is an index of the output dimension. Similarly, let $W^{\langle p, n\rangle}$ be a scalar element of the matrix $W$ on the $p$-th row and $n$-th column.
With the matrix multiplication, the scalar output $g^{\langle p \rangle}(z_x)$ can be considered as the sum of multiple scalar linear operators $W^{\langle p, n\rangle}\!\cdot\!z_{x}^{\langle n \rangle}$.
By this property, we show that $g^{\langle p \rangle}(z_{x_i})\!-\!g^{\langle p \rangle}(z_{x_j})\!\geq\!0$ if $z_{x_i}\!\succeq\!z_{x_j}$ by
\begin{equation}
\begin{split}
g^{\langle p \rangle}(z_{x_i})-g^{\langle p \rangle}(z_{x_j}) &= W^{\langle p,\cdot \rangle} \cdot z_{x_i} - W^{\langle p,\cdot \rangle} \cdot z_{x_j} =  \sum_{n=1}^d{\big(W^{\langle p,n\rangle}\cdot z^{\langle n \rangle}_{x_i} - W^{\langle p,n\rangle}\cdot z^{\langle n \rangle}_{x_j}  \big)}\\
&=  \sum_{n=1}^d{\big(W^{\langle p,n\rangle}\cdot (z^{\langle n \rangle}_{x_i} -z^{\langle n \rangle}_{x_j})  \big)} \ge 0. 
\end{split}
\label{eq:layer1_scalar_output}
\end{equation}
Therefore, without loss of generality, $g(z_{x_i})-g(z_{x_j}) \succeq 0$ if $z_{x_i} \succeq z_{x_j}$. This concludes the proof.
\end{proof}
\label{lemma:skyline_layer_1}
\end{lemma}


\begin{lemma}
Let $h(z_{x})\!=\!\sigma (g(z_{x}))$ where $\sigma$ is a non-decreasing non-linear activation function. If the skyline constraint holds by $g(\cdot)\in\mathbb{R}^d$, 
the function $h(z_{x})$ also satisfies the skyline constraint as
\begin{equation}
z_{x_i} \succeq z_{x_j} \implies h(z_{x_i}) \succeq h(z_{x_j}).
\label{eq:skyline_non_linear}
\end{equation}
\begin{proof}
By the {composition} rule of the non-decreasing function, applying any non-decreasing function does not change the order of its inputs.
Therefore, $\sigma(g(z_{x_i}))-\sigma(g(z_{x_j}))\succeq 0$ if $g(z_{x_i}) \succeq g(z_{x_j})$.
\end{proof}
\label{lemma:skyline_non_linear}
\end{lemma}

\begin{lemma}
Let $h^{[k]}(z_x)\!=\sigma\big(W^{[k]}\cdot h^{[k-1]}(z_{x})\!+\!b^{[k]}\big)$ be the $k$-th layer of a non-negative-weighted MLP $(k\geq 2)$, where $W^{[k]}\!\in\mathbb{R}^{m^\prime\times m}\succeq 0$ and $b^{[k]}\in\mathbb{R}^{m^\prime}\succeq 0$. If $h^{[k-1]}(\cdot)\in\mathbb{R}^m$ satisfies the skyline constraint, 
the function $h^{[k]}(z_x)$ also holds the skyline constraint as
\begin{equation}
z_{x_i} \succeq z_{x_j} \implies h^{[k]}(z_{x_i}) \succeq h^{[k]}(z_{x_j}).
\label{eq:skyline_layer_k}
\end{equation}

\begin{proof}
Let $W^{[k]}$ be $W$, $h^{[k]}(z_{x})$ be $h(z_{x})$, and $h^{[k-1]}(z_{x})$ be $h_{input}(z_{x})$ for notation simplicity.
Since an intermediate layer uses $h_{input}(z_{x})$ as its input rather than $z$, Equation \ref{eq:layer1_scalar_output} changes to 
\begin{equation}
\begin{split}
g^{\langle p \rangle}(z_{x_i})-g^{\langle p \rangle}(z_{x_j}) &= \sum_{n=1}^d{\big(W^{\langle p,n\rangle}\cdot \big(h_{input}^{\langle n \rangle}(z_{x_i}) -h_{input}^{\langle n \rangle}(z_{x_j})\big)  \big)} \ge 0, 
\end{split}
\label{eq:layer_k_scalar_output}
\end{equation}
where $g^{\langle p \rangle}(z_{x_i})$ is the $p$-th dimension's output before applying non-linear activation $\sigma$.
Since $h_{input}(\cdot)$ satisfies the skyline constraint, $h_{input}^{\langle n \rangle}(z_{x_i})>h_{input}^{\langle n \rangle}(z_{x_j})$ when $z_{x_i} \succeq z_{x_j}$, $g^{\langle p \rangle}(z_{x_i}) > g^{\langle p \rangle}(z_{x_j})$ for all $p\in\{1,\ldots,m^\prime\}$.
By Lemma\,\ref{lemma:skyline_non_linear}, $h^{\langle p \rangle}(z_{x_i})-h^{\langle p \rangle}(z_{x_j})=\sigma(g^{\langle p \rangle}(z_{x_i}))-\sigma(g^{\langle p \rangle}(z_{x_j})) \ge 0$ for all $p$.
Therefore, $z_{x_i} \succeq z_{x_j} \implies h^{[k]}(z_{x_i}) \succeq h^{[k]}(z_{x_j})$.
\end{proof}
\label{lemma:skyline_layer_k}
\end{lemma}

\begin{theorem}
For any non-negative-weighted MLP $\Phi(z; \textbf{w})\!: \mathbb{R}^d \rightarrow \mathbb{R}$ where $\textbf{w}\succeq0$, the skyline constraint holds such that $z_{x_i}\succeq z_{x_j} \implies \Phi(z_{x_i}) \ge \Phi(z_{x_j})~\forall z_{x_i}, z_{x_j} \in \mathbb{R}^d \succeq 0$.
\label{theorem:complete_skyline}
\begin{proof}
By mathematical induction, where Lemmas~\ref{lemma:skyline_layer_1} and \ref{lemma:skyline_non_linear} constitute the base step, and Lemma~\ref{lemma:skyline_layer_k} is the inductive step, any non-negative-weighted MLP satisfies the skyline constraint.
\end{proof}
\end{theorem}

\subsection{Active Learning with \algname{}}

\textbf{Meta-input Conversion.}\hspace{0.1cm}
\algname{} receives $z_x = \langle \mathcal{P}(x), \mathcal{I}(x) \rangle$ and then returns a meta-score for query selection. All the scores for the input of \algname{} should be positive to preserve the skyline constraint, \emph{i.e.}, $z \succeq 0$. Existing OOD and AL query scores are converted to the meta-input.
The methods used for calculating the scores are orthogonal to our framework.
The OOD score $\mathcal{O}(\cdot)$ is conceptually the opposite of purity and varies in its scale; hence, we convert it to a purity score by $\mathcal{P}(x)={\rm Exp}({\rm Normalize}(-\mathcal{O}(x)))$, where ${\rm Normalize}(\cdot)$ is the z-score normalization. 
This conversion guarantees the purity score to be positive. Similarly, for the informativeness score, we convert an existing AL query score $\mathcal{Q}(\cdot)$ to $\mathcal{I}(x)={\rm Exp}({\rm Normalize}(\mathcal{Q}(x)))$.
{For the z-score normalization, we compute the mean and standard deviation of $\mathcal{O}(x)$ or $\mathcal{Q}(x)$ over the all unlabeled examples. Such mean and standard deviation are iteratively computed before the meta-training and used for the z-score normalization at that round.}

\textbf{Mini-batch Optimization.}\hspace{0.1cm} Mini-batch examples are sampled from the labeled query set $S_Q$ which contains both IN and OOD examples.
Since the meta-objective in Equation \ref{eq:ranking_loss} is a ranking loss, a mini-batch $\mathcal{M}$ is a set of meta-input pairs such that $\mathcal{M}=\{(z_{x_i}, z_{x_j})|~x_i,x_j\in S_Q\}$ where $z_x=\langle \mathcal{P}(x), \mathcal{I}(x) \rangle$.
To construct a paired mini-batch $\mathcal{M}$ of size $M$, we randomly sample $2M$ examples from $S_Q$ and pair the $i$-th example with the $(M\!+\!i)$-th one for all $i \in \{1,\ldots\!,M\}$.
Then, the loss for mini-batch optimization of \algname{} is defined as
\begin{equation}
\begin{gathered}
\small
\!\!\!\!\!\!\mathcal{L}_{meta}(\mathcal{M}) \!=\!\!\!\sum_{(i,j)\in \mathcal{M} }\!\!\max\!\Big(0, - \text{Sign}\big(\ell_{mce}(x_i),\ell_{mce}(x_j)\big) \!\cdot \!\big(\Phi(z_{x_i}; \textbf{w})\!-\!\Phi(z_{x_j}; \textbf{w})\!+\!\eta\big) \Big) \!:\!\textbf{w}\succeq 0.
\end{gathered}
\label{eq:mini_batch_loss}
\end{equation}

\textbf{Overall Procedure.}\hspace{0.1cm}
For each AL round, a target model is trained via stochastic gradient descent\,(SGD) on mini-batches sampled from the IN examples in the current labeled set $S_L$. Based on the current target model, the purity and informative scores are computed by using certain OOD and AL query scores. The querying phase is then performed by selecting the examples $S_{Q}$ with the highest meta-scores within the labeling budget $b$. The query set $S_Q$ is used as the self-validation set for training \algname{} at the current AL round. The trained \algname{} is used at the next AL round. The alternating procedure of updating the target model and the meta-model repeats for a given number $r$ of AL rounds.

\textbf{Algorithm Pseudocode.}\hspace{0.1cm} Algorithm \ref{alg:algorithm_pseudocode} describes the overall active learning procedure with \algname{}, which is self-explanatory.
For each AL round, a target model $\Theta$ is trained via stochastic gradient descent\,(SGD) using IN examples in the labeled set $S_L$ (Lines 3--5).
This trained target model is saved as the final target model at the current round.
Next, the querying phase is performed according to the order of meta-query scores from $\Phi$ given the budget $b$ (Lines 6--13). 
Then, the meta-training phase is performed, and the meta-model $\textbf{w}$ is updated via SGD using  the labeled query set $S_Q$ as a self-validation set (Lines 14--17). 
Lines 3--17 repeats for the given number  $r$ of rounds.
In the first round, because there is no meta-model trained in the previous round, the query set is constructed by choosing the examples whose sum of purity and informativeness scores is the largest (Lines 9--10).

\newcommand{\INDSTATE}[1][1]{\STATE\hspace{#1\algorithmicindent}}
\begin{algorithm}[t]
\small
\caption{AL Procedure with \algname{}}
\label{alg:algorithm_pseudocode}
\begin{algorithmic}[1]
\REQUIRE {${S}_L$: labeled set, ${U}$: unlabeled set, $r$: number of rounds, $b$: labeling budget, $C$: cost function, $\Theta$: parameters of the target model, \textbf{w}: parameters of \algname{}}
\ENSURE {Final target model $\Theta_{*}$}
\INDSTATE[0] $\textbf{w} \leftarrow$ Initialize the meta-model parameters;
\INDSTATE[0] {\bf for} $r=1$ {\bf to} $r$ {\bf do} 
\INDSTATE[1] /* {\color{blue} Training the target model parameterized by  $\Theta$}*/ 
\INDSTATE[1] $\Theta \leftarrow$ Initialize the target model parameters;
\INDSTATE[1] $\Theta \leftarrow {\rm TrainingClassifier}({S}_L, \Theta)$;
\INDSTATE[1] /* {\color{blue} Querying for the budget $b$} */
\INDSTATE[1] {${S}_Q \leftarrow \emptyset$}; 
\INDSTATE[1] {\bf while} $C({S}_Q) \leq b$ {\bf do}
\INDSTATE[2] {{\bf if} $r=1$ {\bf do}}
\INDSTATE[3] {${S}_Q\leftarrow {S}_Q \cup \argmaxA_{x\in U}(\mathcal{P}(x)+\mathcal{I}(x))$}; 
\INDSTATE[2] {{\bf else} {\bf do}}
\INDSTATE[3] {${S}_Q\leftarrow {S}_Q \cup \argmaxA_{x\in U}(\Phi(x; \textbf{w}))$}; 
\INDSTATE[1] {${S}_L\leftarrow {S}_L \cup {S}_Q$};\ \ \ {${U}\leftarrow {U}\!\setminus\!{S}_Q$}
\INDSTATE[1] /* {\color{blue} Training \algname{} $\Phi$ parameterized by $\textbf{w}$}  */ 
\INDSTATE[1] {\bf for} $t=1$ {\bf to} meta-train-steps {\bf do}
\INDSTATE[2] {Draw a mini-batch $\mathcal{M}$ and from $S_Q$;}
\INDSTATE[2] $\textbf{w} \leftarrow \textbf{w} -\alpha \nabla_{\textbf{w}}\big( \mathcal{L}_{meta}(\mathcal{M}) \big)$;
\INDSTATE[0] {\bf return} $\Theta$;
\end{algorithmic}
\end{algorithm}

\section{Experiments}
\label{sec:experiment}

\subsection{Experiment Setting}
\label{sec:experiment_setting}

\noindent\textbf{Datasets.} We perform the active learning task on three benchmark datasets; CIFAR10 \cite{krizhevsky2009learning}, CIFAR100 \cite{krizhevsky2009learning}, and ImageNet \cite{deng2009imagenet}.
Following the `split-dataset' setup in open-world learning literature\,\cite{du2022conal, kothawade2021similar, saito2021openmatch}, we divide each dataset into two subsets: (1) the target set with IN classes and (2) the noise set with OOD classes. Specifically, CIFAR10 is split into the target set with four classes and the noise set with the rest six classes; CIFAR100 into the two sets with 40 and 60 classes; and ImageNet into the two sets with 50 and 950 classes. 
The entire target set is used as the unlabeled IN data, while only a part of the classes in the noise set is selected as the unlabeled OOD data according to the given noise ratio. 

In addition, following OOD detection literature\,\cite{MSP, ren2019likelihood}, we also consider the `cross-dataset' setup, which mixes a certain dataset with two external OOD datasets collected from different domains, such as LSUN\,\cite{yu2015lsun} and Places365\,\cite{zhou2017places}. 
Each of CIFAR10, CIFAR100, and ImageNet is mixed with OOD examples sampled from an OOD dataset combined from two different domains---LSUN \cite{yu2015lsun}, an indoor scene understanding dataset of 59M images with 10 classes, and Places365 \cite{zhou2017places}, a large collection of place scene images with 365 classes.
The resolution of LSUN and Places365 is resized into 32$\times$32 after random cropping when mixing with CIFAR10 and CIFAR100.
For ImageNet, as in the split-dataset setup, we use 50 randomly-selected classes as IN examples, namely ImageNet50.

\smallskip\smallskip
\textbf{Algorithms.}\hspace{0.1cm} We compare \algname{} with {a random selection}, four standard AL, and two recent open-set AL approaches.

\begin{itemize}[leftmargin=9pt, noitemsep] 
\item \emph{Standard AL}: The four methods perform AL {without} any processing for open-set noise: (1) CONF\,\cite{wang2014active} queries the most uncertain examples with the lowest softmax confidence in the prediction, (2) CORESET\,\cite{sener2018active} queries the most diverse examples with the highest coverage in the representation space, (3) LL\,\cite{yoo2019learning} queries the examples having the largest predicted loss by jointly learning a loss prediction module, and (4) BADGE \cite{Ash2020badge} considers both uncertainty and diversity by querying the most representative examples in the gradient via $k$-means{++} clustering\,\cite{arthur2006k}.
\item \emph{Open-set AL}: The two methods tend to put more weight on the examples with high purity: (1) CCAL\,\cite{du2022conal} learns two contrastive coding models for calculating informativeness and OODness, and then it combines the two scores into one using a heuristic balancing rule, and (2) SIMILAR\,\cite{kothawade2021similar} selects a pure and core set of examples that maximize the distance coverage on the entire unlabeled data while minimizing the distance coverage to the already labeled OOD data. 
\end{itemize}

For all the experiments, regarding the two inputs of \algname{}, we mainly use CSI\,\cite{tack2020csi} and LL\,\cite{yoo2019learning} for calculating the purity and informativeness scores, respectively. 
For CSI, as in CCAL, we train a contrastive learner on the entire unlabeled set with open-set noise since the clean in-distribution set is not available in open-set AL.
The ablation study in Section \ref{sec:ablation_study} shows that \algname{} is also effective with other OOD and AL scores as its input.

\smallskip\smallskip
\textbf{Implementation Details.}\hspace{0.1cm}
We repeat the three steps---training, querying, and labeling---of AL. The total number $r$ of rounds  is set to 10.
Following the prior open-set AL setup\,\cite{du2022conal, kothawade2021similar}, for the split-dataset setup, we set the labeling cost $c_{IN}=1$ for IN examples and $c_{OOD}=1$ for OOD examples. For the class-split setup, the labeling budget $b$ per round is set to $500$ for CIFAR10/100 and $1,000$ for ImageNet.
For the cross-dataset setup, the budget $b$ is set to $1,000$ for CIFAR-10 and ImageNet50 and $2,000$ for CIFAR-100 following the literature\,\cite{yoo2019learning}.
Regarding the open-set noise ratio $\tau$, we configure four different levels from light to heavy noise in $\{10\%, 20\%, 40\%, 60\% \}$. 
In the case of $\tau=0\%$ (no noise), \algname{} naturally discards the purity score and only uses the informativeness score for query selection, since the self-validation set does not contain any OOD examples.
The initial labeled set is randomly selected uniformly at random from the entire unlabeled set within the labeling budget $b$. 
For instance, when $b$ is $1,000$ and $\tau$ is $20\%$, 800 IN examples and 200 OOD examples are expected to be selected as the initial set.
{For the architecture of \algname{}, we use a 2-layer MLP with a hidden dimension size of 64 and the Sigmoid activation function.}
We report the average results of five runs with different class splits. 
{We did \emph{not} use any pre-trained networks.}

\textbf{Training Configurations.}\hspace{0.1cm} We train ResNet-18 using SGD with a momentum of 0.9 and a weight decay of 0.0005, and a batch size of 64. The initial learning rate of $0.1$ is decayed by a factor of 0.1 at 50$\%$ and 75$\%$ of the total training iterations. In the setup of open-set AL, the number of IN examples for training differs depending on the query strategy. We hence use a fixed number of training iterations instead of epochs for fair optimization. The number of training iterations is set to 20,000 for CIFAR10/100 and 30,000 for ImageNet. We set $\eta$ to 0.1 for all cases. We train \algname{} for 100 epochs using SGD with a weight decay of 0.0005, and a mini-batch size of 64. An initial learning rate of $0.01$ is decayed by a factor of 0.1 at 50$\%$ of the total training iterations. 
Since \algname{} is not trained at the querying phase of the first AL round, we simply use the linear combination of purity and informativeness as the query score, \textit{i.e.}, $\Phi(x)=\mathcal{P}(x)+\mathcal{I}(x)$.
For calculating the CSI-based purity score, we train a contrastive learner for CSI with 1,000 epochs under the LARS optimizer with a batch size of 32. Following CCAL\,\cite{du2021contrastive}, we use the distance between each unlabeled example to the closest OOD example in the labeled set on the representation space of the contrastive learner as the OOD score.
The hyperparameters for other algorithms are favorably configured following the original papers. 
{All methods are implemented with PyTorch 1.8.0 and executed on a single NVIDIA Tesla V100 GPU.}
{The code is available at \url{https://github.com/kaist-dmlab/MQNet}}.

\begin{figure*}[t!]
\begin{center}
\includegraphics[width=\linewidth]{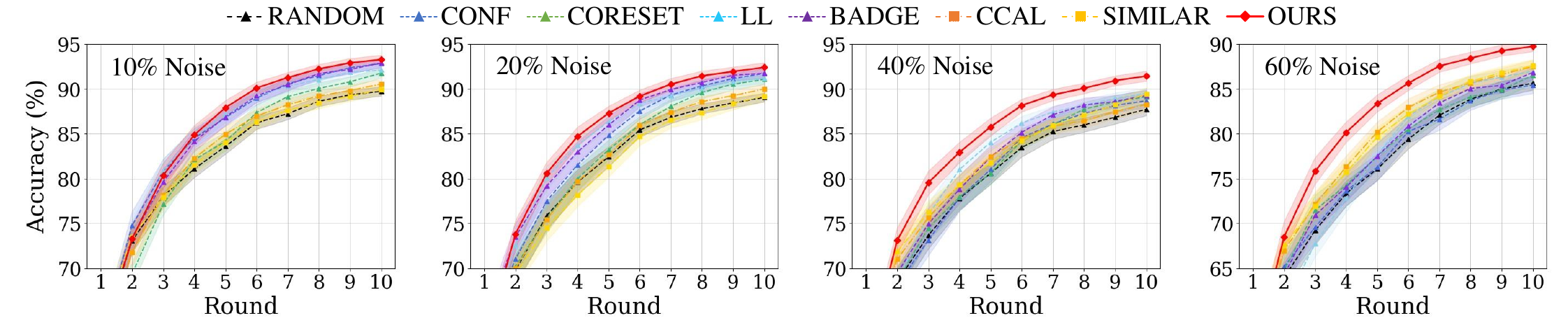}
\end{center}
\hspace*{0.5cm} {\small (a) Accuracy comparison over AL rounds on CIFAR10 with open-set noise of $10\%$, $20\%$, $40\%$, and $60\%$.}
\begin{center}
\includegraphics[width=\linewidth]{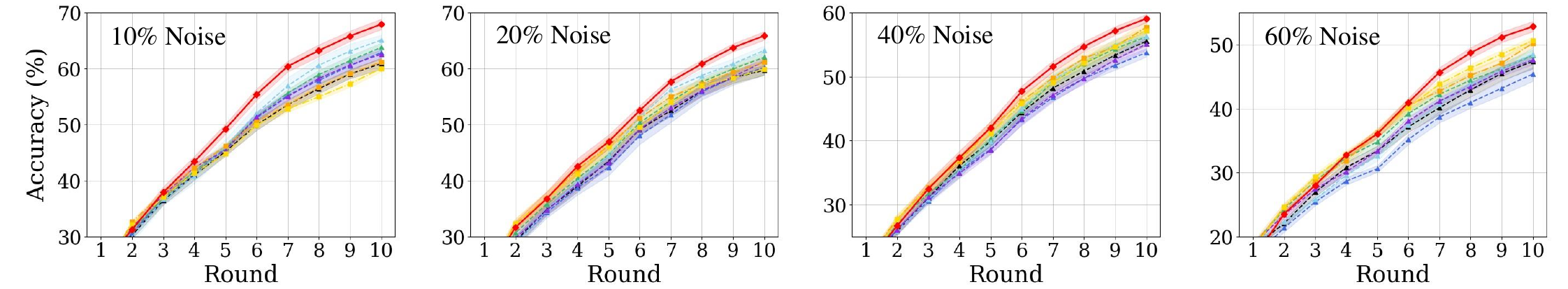}
\end{center}
\hspace*{0.5cm} {\small (b) Accuracy comparison over AL rounds on CIFAR100 with open-set noise of $10\%$, $20\%$, $40\%$, and $60\%$.}
\begin{center}
\includegraphics[width=\linewidth]{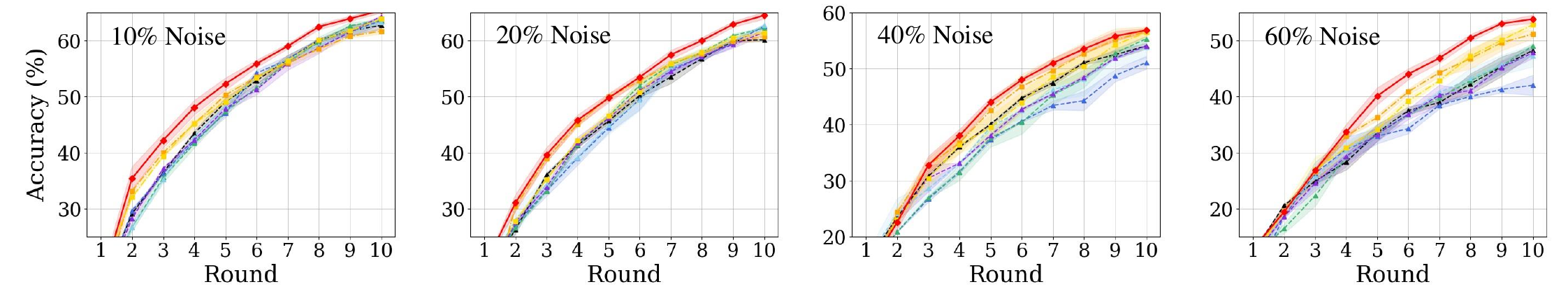}
\end{center}
\hspace*{0.5cm} {\small (c) Accuracy comparison over AL rounds on ImgeNet with open-set noise of $10\%$, $20\%$, $40\%$, and $60\%$.}
\caption{
Test accuracy over AL rounds for CIFAR10, CIFAR100, and ImageNet with varying open-set noise ratios.
}
\label{fig:performance_over_rounds}
\end{figure*}

\begin{table*}[t!]
\def\arraystretch{1.3}
\caption{Last test accuracy ($\%$) at the final round for CIFAR10, CIFAR100, and ImageNet. The best results are in bold, and the second best results are underlined.}
\label{table:overall_performance}
\begin{center}
\scriptsize
\begin{tabular}{c|c|c c c c|c c c c|c c c c} \toprule
\multicolumn{2}{c|}{Datasets}&\multicolumn{4}{c|}{CIFAR10 (4:6 split)}&\multicolumn{4}{c|}{CIFAR100 (40:60 split)}&\multicolumn{4}{c}{ImageNet (50:950 split)}\\\hline \addlinespace[0.15ex]
\multicolumn{2}{c|}{Noise Ratio} &\!10$\%$\!&\!20$\%$\!&\!40$\%$\!&\!60$\%$\!&\!10$\%$\!&\!20$\%$\!&\!40$\%$\!&\!60$\%$\!&\!10$\%$\!&\!20$\%$\!&\!40$\%$\!&\!60$\%$\!\\ \addlinespace[0.15ex]\hline\addlinespace[0.3ex]
{\hspace{-0.2cm}Non-AL}&{\hspace{-0.2cm}RANDOM\hspace{-0.2cm}}&\!{89.83}\!&\!{89.06}\!&\!{87.73}\!&\!{85.64}\!&\!{60.88}\!&\!{59.69}\!&\!{55.52}\!&\!{47.37}\!&\!{62.72}\!&\!{60.12}\!&\!{54.04}\!&\!{48.24}\!\\\cline{1-14}\Tstrut
\multirow{4}{*}{\makecell[c]{\hspace{-0.2cm}Standard\\AL}}&\hspace{-0.2cm}CONF\hspace{-0.2cm}&\!\underline{92.83}\!&\!91.72\!&\!88.69\!&\!85.43\!&\!62.84\!&\!60.20\!&\!53.74\!&\!45.38\!&\!63.56\!&\!\underline{62.56}\!&\!51.08\!&\!45.04\!\\\cline{2-14}\Tstrut
&\hspace{-0.2cm}CORESET\hspace{-0.2cm}&\!91.76\!&\!91.06\!&\!89.12\!&\!86.50\!&\!63.79\!&\!62.02\!&\!56.21\!&\!48.33\!&\!63.64\!&\!{62.24}\!&\!55.32\!&\!49.04\!\\\cline{2-14}\Tstrut
&\hspace{-0.2cm}LL\hspace{-0.2cm}&\!92.09\!&\!91.21\!&\!\underline{89.41}\!&\!86.95\!&\!\underline{65.08}\!&\!\underline{64.04}\!&\!56.27\!&\!48.49\!&\!63.28\!&\!61.56\!&\!55.68\!&\!47.30\!\\\cline{2-14}\Tstrut
&\hspace{-0.2cm}BADGE\hspace{-0.2cm}&\!{92.80}\!&\!\underline{91.73}\!&\!89.27\!&\!86.83\!&\!62.54\!&\!61.28\!&\!55.07\!&\!47.60\!&\!\underline{64.84}\!&\!61.48\!&\!54.04\!&\!47.80\!\\\cline{1-14}\Tstrut
\multirow{2}{*}{\makecell[c]{\hspace{-0.2cm}Open-set\\AL}}&\hspace{-0.2cm}CCAL\hspace{-0.2cm}&\!90.55\!&\!89.99\!&\!88.87\!&\!\underline{87.49}\!&\!61.20\!&\!61.16\!&\!\underline{56.70}\!&\!50.20\!&\!61.68\!&\!60.70\!&\!\underline{56.60}\!&\!51.16\!\\\cline{2-14}\Tstrut
&\hspace{-0.2cm}SIMILAR\hspace{-0.2cm}&\!89.92\!&\!89.19\!&\!88.53\!&\!87.38\!&\!60.07\!&\!59.89\!&\!56.13\!&\!\underline{50.61}\!&\!63.92\!&\!61.40\!&\!56.48\!&\!\underline{52.84}\!\\\cline{1-14} \Tstrut
\hspace{-0.2cm}Proposed&\hspace{-0.2cm}\textbf{MQ-Net}\hspace{-0.2cm}&\!\textbf{93.10}\!&\!\textbf{92.10}\!&\!\textbf{91.48}\!&\!\textbf{89.51}\!&\!\textbf{66.44}\!&\!\textbf{64.79}\!&\!\textbf{58.96}\!&\!\textbf{52.82}\!&\!\textbf{65.36}\!&\!\textbf{63.08}\!&\!\textbf{56.95}\!&\!\textbf{54.11}\!\\\hline\addlinespace[0.3ex]
\multicolumn{2}{c|}{\hspace{-0.2cm}\emph{$\%$ improve over 2nd best}\hspace{-0.3cm}}&0.32&0.40&2.32&2.32&2.09&1.17&3.99&4.37&0.80&1.35&0.62&2.40\\\hline
\multicolumn{2}{c|}{\hspace{-0.2cm}\emph{$\%$ improve over the least}\hspace{-0.3cm}}&3.53&3.26&3.33&4.78&10.6&8.18&9.71&16.39&5.97&3.92&11.49&20.14\\\bottomrule
\end{tabular}
\end{center}
\end{table*}

\subsection{Experiment Results on Split-datasets}
\label{sec:robustness}

\textbf{Results over AL Rounds.}\hspace{0.1cm}
Figure \ref{fig:performance_over_rounds} illustrates the test accuracy of the target model over AL rounds on the two CIFAR datasets. 
\algname{} achieves the highest test accuracy in most AL rounds, thereby reaching the best test accuracy at the final round in every case for various datasets and noise ratios.
Compared with the two existing open-set AL methods, CCAL and SIMILAR, \algname{} shows a steeper improvement in test accuracy over rounds by resolving the purity-informativeness dilemma in query selection.
%
For example, the performance gap between \algname{} and the two open-set AL methods gets larger after the sixth round, as shown in Figure \ref{fig:performance_over_rounds}(b), 
because CCAL and SIMILAR mainly depend on purity in query selection, which conveys less informative information to the classifier. For a better classifier, informative examples should be favored at a later AL round due to the sufficient number of IN examples in the labeled set.
In contrast, \algname{} keeps improving the test accuracy even in a later AL round by finding the best balancing between purity and informativeness in its query set. More analysis of \algname{} associated with the purity-informativeness dilemma is discussed in Section \ref{sec:how_resolve_dilemma}.

\textbf{Results with Varying Noise Ratios.}\hspace{0.1cm}
Table \ref{table:overall_performance} summarizes the last test accuracy at the final AL round for three datasets with varying levels of open-set noise. Overall, the last test accuracy of \algname{} is the best in every case.
This superiority concludes that \algname{} successfully finds the best trade-off between purity and informativeness in terms of AL accuracy regardless of the noise ratio.
In general, the performance improvement becomes larger with the increase in the noise ratio.
On the other hand, the two open-set AL approaches are even worse than the four standard AL approaches when the noise ratio is less than or equal to 20$\%$. 
Especially, in CIFAR10 relatively easier than others, CCAL and SIMILAR are inferior to the non-robust AL method, LL, even with 40$\%$ noise. 
This trend confirms that increasing informativeness is more crucial than increasing purity when the noise ratio is small; highly informative examples are still beneficial when the performance of a classifier is saturated in the presence of open-set noise.
{An in-depth analysis of the low accuracy of the existing open-set AL approaches in a low noise ratio is presented in Section \ref{sec:why_low_challenging_appendix}.}

\subsection{Experiment Results on Cross-datasets}
\label{sec:cross_dataset_results}

\begin{figure*}[t!]
\begin{center}
\includegraphics[width=\linewidth]{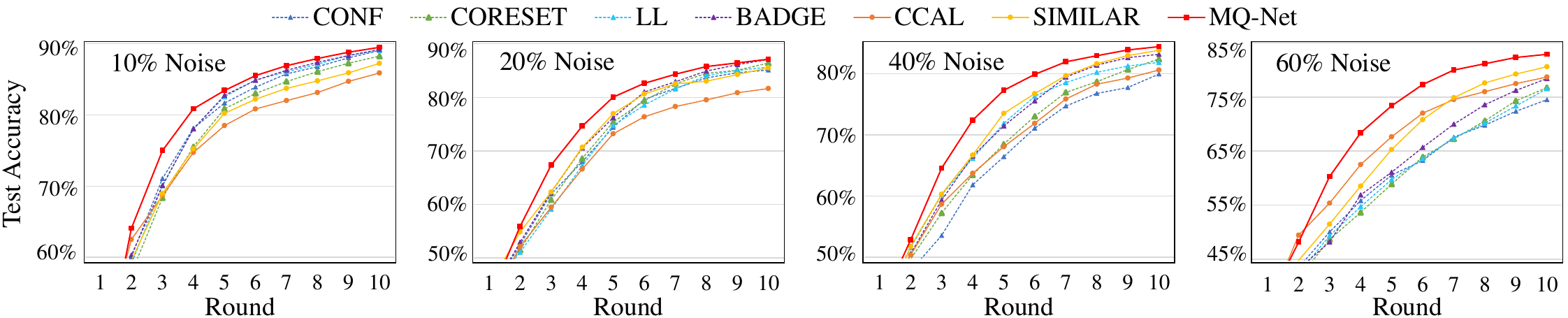}
\end{center}
\hspace*{1cm} {\small (a) Accuracy over AL rounds on cross-CIFAR10 with open-set noise of $10\%$, $20\%$, $40\%$, and $60\%$.}
\begin{center}
\includegraphics[width=\linewidth]{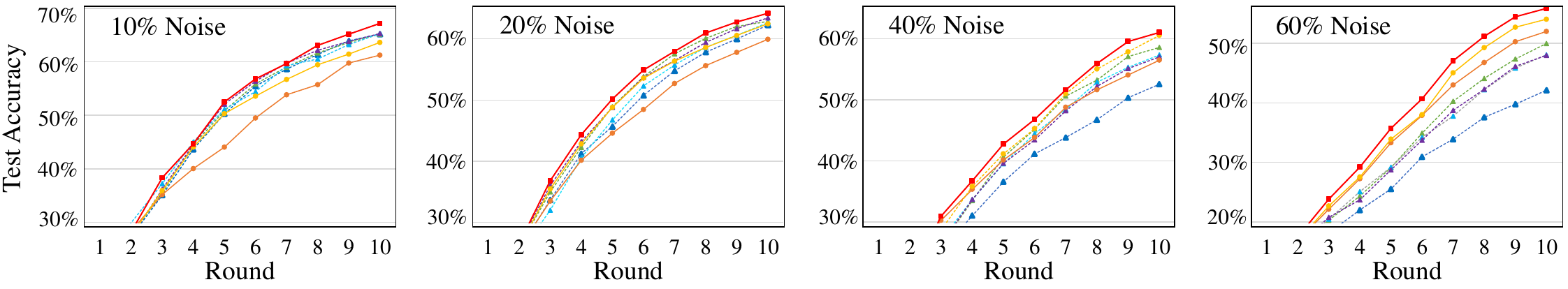}
\end{center}
\hspace*{0.5cm} {\small (b) Accuracy comparison over AL rounds on CIFAR100 with open-set noise of $10\%$, $20\%$, $40\%$, and $60\%$.}
\begin{center}
\includegraphics[width=\linewidth]{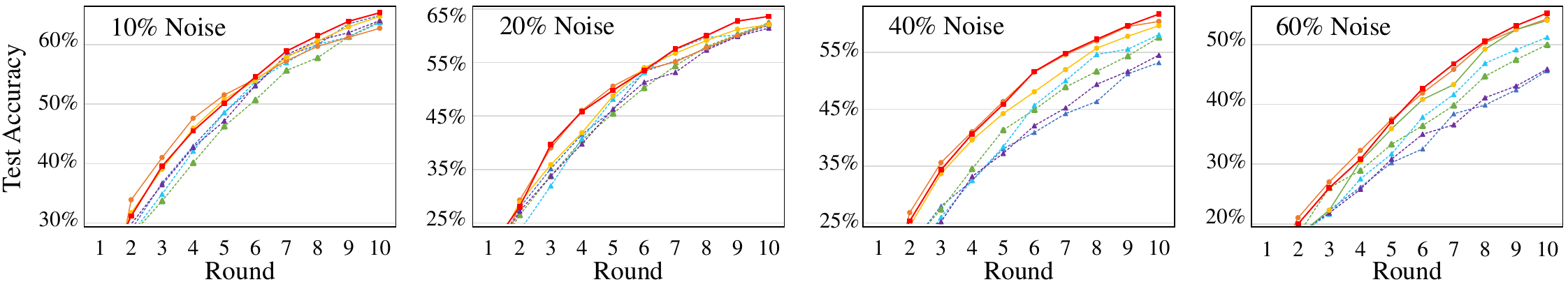}
\end{center}
\hspace*{0.5cm} {\small (c) Accuracy comparison over AL rounds on ImageNet with open-set noise of $10\%$, $20\%$, $40\%$, and $60\%$.}
\caption{
Test accuracy over AL rounds for the three \emph{cross-datasets}, CIFAR10, CIFAR100, and ImageNet, with varying open-set noise ratios.
}
\label{fig:performance_over_rounds_cross_datasets}
\end{figure*}

\textbf{Results over AL Rounds.}\hspace{0.1cm}
Figure \ref{fig:performance_over_rounds_cross_datasets} shows the test accuracy of the target model throughout AL rounds on the three cross-datasets. 
Overall, as analyzed in Section \ref{sec:robustness}, \algname{} achieves the highest test accuracy in most AL rounds, thereby reaching the best test accuracy at the final round in every case of various datasets and noise ratios.
Compared with the two existing open-set AL methods, CCAL and SIMILAR, \algname{} shows a steeper improvement in test accuracy over rounds by resolving the purity-informativeness dilemma in query selection, which shows that \algname{} keeps improving the test accuracy even in a later AL round by finding the best balancing between purity and informativeness in its query set. Together with the results in Section \ref{sec:robustness}, we confirm that \algname{} is robust to the two different distributions---`split-dataset' and `cross-dataset'---of open-set noise.

\textbf{Results with Varying Noise Ratios.}\hspace{0.1cm}
Table \ref{table:last_test_accuracy_cross_datasets} summarizes the last test accuracy at the final AL round for three cross-datasets with varying levels of open-set noise. 
Overall, the last test accuracy of \algname{} is the best in every case, which shows that \algname{} keeps finding the best trade-off between purity and informativeness in terms of AL accuracy regardless of the noise ratio. The performance improvement becomes larger as the noise ratio increases. Meanwhile, CCAL and SIMILAR are even worse than the four standard AL approaches when noise ratio is less than or equal to 20$\%$. 
This trend indicates that focusing on informativeness is more beneficial than focusing on purity when the noise ratio is small.

\begin{table*}[t!]
\def\arraystretch{1.3}
\caption{Last test accuracy ($\%$) at the final round for three cross-datasets: CIFAR10, CIFAR100, and ImageNet50 mixed with the merger of  LSUN and Places365. The best results are in bold, and the second best results are underlined.}
\label{table:last_test_accuracy_cross_datasets}
\begin{center}
\scriptsize
\begin{tabular}{c|c|c c c c|c c c c|c c c c} \toprule
\multicolumn{2}{c|}{Datasets}&\multicolumn{4}{c|}{Cross-CIFAR10}&\multicolumn{4}{c|}{Cross-CIFAR100}&\multicolumn{4}{c}{Cross-ImageNet50}\\\hline \addlinespace[0.1ex]
\multicolumn{2}{c|}{Noise Ratio} &\!10$\%$\!&\!20$\%$\!&\!40$\%$\!&\!60$\%$\!&\!10$\%$\!&\!20$\%$\!&\!40$\%$\!&\!60$\%$\!&\!10$\%$\!&\!20$\%$\!&\!40$\%$\!&\!60$\%$\!\\ \addlinespace[0.15ex]\hline\addlinespace[0.3ex]
\multirow{4}{*}{\makecell[c]{Standard\\AL}}&CONF&\!89.04\!&\!85.09\!&\!79.92\!&\!74.48\!&\!65.17\!&\!62.24\!&\!52.52\!&\!42.13\!&\!\underline{64.92}\!&\!61.92\!&\!53.60\!&\!45.64\!\\\cline{2-14}\Tstrut
&CORESET&\!88.26\!&\!86.38\!&\!82.36\!&\!76.71\!&\!65.13\!&\!62.83\!&\!58.56\!&\!49.98\!&\!63.88\!&\!\underline{62.40}\!&\!57.60\!&\!50.02\!\\\cline{2-14}\Tstrut
&LL&\!89.06\!&\!85.65\!&81.81\!&\!76.52\!&\!65.23\!&\!62.64\!&\!57.32\!&\!48.07\!&\!63.68\!&\!62.32\!&\!58.08\!&\!51.24\!\\\cline{2-14}\Tstrut
&BADGE&\!\underline{89.2}\!&\!\underline{87.07}\!&\!83.14\!&\!78.38\!&\!\underline{65.27}\!&\!\underline{63.42}\!&\!57.01\!&\!48.07\!&\!{64.04}\!&\!61.40\!&\!54.48\!&\!45.92\!\\\cline{1-14}\Tstrut
\multirow{2}{*}{\makecell[c]{Open-set\\AL}}&CCAL&\!85.89\!&81.62\!&\!80.55\!&\!78.68\!&\!61.22\!&\!59.91\!&\!56.47\!&\!52.01\!&\!62.72\!&\!62.20\!&\!\underline{60.40}\!&\!\underline{54.32}\!\\\cline{2-14}\Tstrut
&SIMILAR&\!87.24\!&\!85.50\!&\!\underline{83.80}\!&\!\underline{80.58}\!&\!63.61\!&\!62.46\!&\!\underline{60.52}\!&\!\underline{54.05}\!&\!64.72\!&\!62.04\!&\!59.68\!&\!{54.05}\!\\\cline{1-14} \Tstrut
Proposed&\textbf{MQ-Net}&\!\textbf{89.49}\!&\!\textbf{87.12}\!&\!\textbf{84.39}\!&\!\textbf{82.88}\!&\!\textbf{67.17}\!&\!\textbf{64.17}\!&\!\textbf{61.01}\!&\!\textbf{55.87}\!&\!\textbf{65.36}\!&\!\textbf{63.60}\!&\!\textbf{61.68}\!&\!\textbf{55.28}\!\\\bottomrule
\end{tabular}
\end{center}
\end{table*}

\begin{figure*}[t!]
\begin{center}
\includegraphics[width=0.95\linewidth]{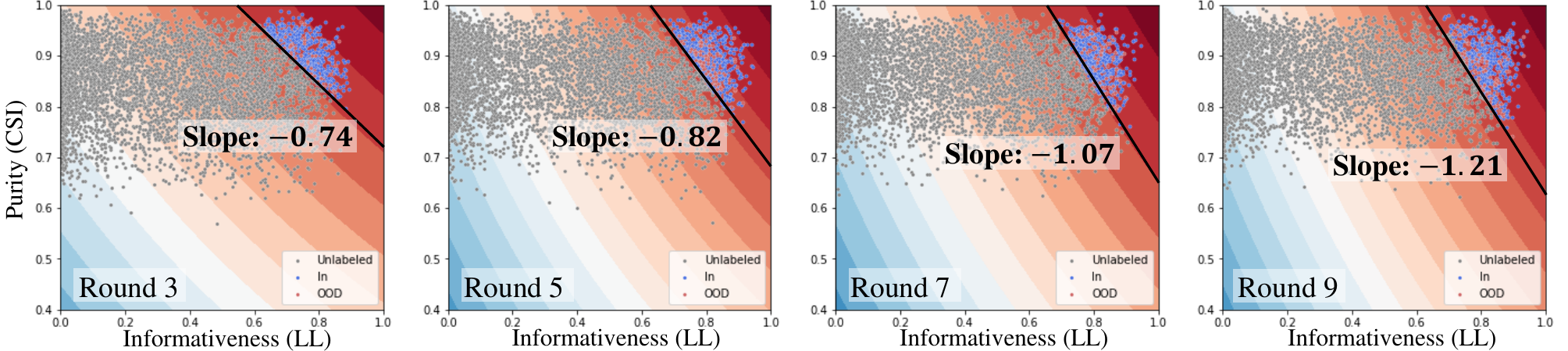}
\end{center}
\hspace*{2.2cm} {\small (a) The output of \algname{} over AL rounds (Round 3, 5, 7, and 9) with 10$\%$ noise.}
\begin{center}
\includegraphics[width=0.95\linewidth]{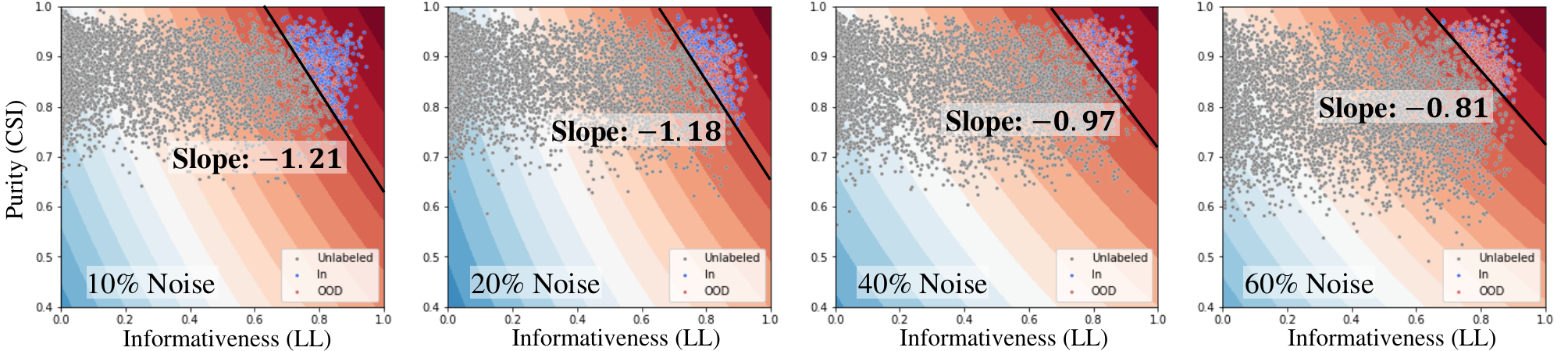}
\end{center}
\hspace*{1cm} {\small (b) The final round's output of \algname{} with varying noise ratios ($10\%$, $20\%$, $40\%$, and $60\%$).}
\caption{
Visualization of the query score distribution of \algname{} on CIFAR100. $x$- and $y$-axis indicate the normalized informativeness and purity scores, respectively. The background color represents the query score of \algname{}; the red is high, and the blue is low. Gray points represent unlabeled data, and blue and red points are the IN and OOD examples in the query set, respectively. The slope of the tangent line on the lowest-scored example in the query set is displayed together; the steeper the slope, the more informativeness is emphasized in query selection.
}
\label{fig:visualization}
\end{figure*}

\subsection{Answers to the Purity-Informativeness Dilemma}
\label{sec:how_resolve_dilemma}
The high robustness of \algname{} in Table \ref{table:overall_performance} and Figure \ref{fig:performance_over_rounds} is mainly attributed to its ability to keep finding the best trade-off between purity and informativeness.
Figure \ref{fig:visualization}(a) illustrates the preference change of \algname{} between purity and informativeness throughout the AL rounds.
As the round progresses, \algname{} automatically raises the importance of informativeness rather than purity; the slope of the tangent line keeps steepening from $-0.74$ to $-1.21$.
This trend implies that more informative examples are required to be labeled when the target classifier becomes mature.
That is, as the model performance increases, ‘fewer but highly-informative’ examples are more impactful than ‘more but less-informative’ examples in terms of improving the model performance.
Figure \ref{fig:visualization}(b) describes the preference change of \algname{} with varying 
noise ratios. Contrary to the trend over AL rounds, as the noise ratio gets higher, \algname{} prefers purity more over informativeness.

\begin{table*}[!t]
\centering
\caption{Effect of the meta inputs on \algname{}.}
\label{table:effect_of_meta_input}
\begin{tabular}{c|c|c c c c}\toprule
\multicolumn{2}{c|}{Dataset}&\multicolumn{4}{c}{CIFAR10 (4:6 split)}\\\hline \addlinespace[0.4ex]
\multicolumn{2}{c|}{Noise Ratio}&$10\%$&$20\%$&$40\%$&$60\%$\\\hline \addlinespace[0.4ex]
Standard AL&BADGE&92.80&91.73&89.27&86.83\\\hline
Open-set AL&CCAL&90.55&89.99&88.87&87.49\\\hline
\multirow{4}{*}{\algname{}} &CONF-ReAct&93.21&91.89&89.54&87.99\\\cline{2-6}\Tstrut
&CONF-CSI&\textbf{93.28}&\textbf{92.40}&91.43&89.37\\\cline{2-6}\Tstrut
&LL-ReAct&92.34&91.85&90.08&88.41\\\cline{2-6}\Tstrut
&LL-CSI&93.10&92.10&\textbf{91.48}&\textbf{89.51}\\\bottomrule
\end{tabular}
\end{table*}

\begin{table*}[!t]
\caption{Efficacy of the self-validation set.} 
\label{table:efficacy_of_query_set}
\centering
\begin{tabular}{c|c|c c c c}\toprule
\multicolumn{2}{c|}{Dataset}&\multicolumn{4}{c}{CIFAR10 (4:6 split)}\\\hline \addlinespace[0.15ex]
\multicolumn{2}{c|}{Noise Ratio}&$10\%$&$20\%$&$40\%$&$60\%$\\\hline \addlinespace[0.15ex]
\multirow{2}{*}{\algname{}} 
&Query set&\textbf{93.10}&\textbf{92.10}&\textbf{91.48}&\textbf{89.51}\\\cline{2-6}\Tstrut
&Random&92.10&91.75&90.88&87.65\\\bottomrule
\end{tabular}
\end{table*}

\subsection{Ablation Studies}
\label{sec:ablation_study}
\textbf{Various Combination of Meta-input.}\hspace{0.1cm} \algname{} can design its purity and informativeness scores by leveraging diverse metrics in the existing OOD detection and AL literature.
Table \ref{table:effect_of_meta_input} shows the final round test accuracy on CIFAR10 for the four variants of score combinations, each of which is constructed by a combination of two purity scores and two informativeness scores; each purity score is induced by the two recent OOD detection methods, ReAct\,\cite{sun2021react} and CSI\,\cite{tack2020csi}, while each informativeness score is converted from the two existing AL methods, CONF and LL. 
``CONF-ReAct'' denotes a variant that uses ReAct as the purity score and CONF as the informativeness score.

Overall, all variants perform better than standard and open-set AL baselines in every noise level. Refer to Table {\ref{table:effect_of_meta_input}} for detailed comparison.
This result concludes that \algname{} can be generalized over different types of meta-input owing to the learning flexibility of MLPs.
Interestingly, the variant using CSI as the purity score is consistently better than those using ReAct. ReAct, a classifier-dependent OOD score, performs poorly in earlier AL rounds.
A detailed analysis of the two OOD detectors, ReAct and CSI, over AL rounds can be found in Section \ref{sec:analysis_purity_scores}.

\smallskip 
\textbf{Efficacy of Self-validation Set.}\hspace{0.1cm}
\algname{} can be trained with an independent validation set, instead of using the proposed self-validation set.
We generate the independent validation set by randomly sampling the same number of examples as the self-validation set with their ground-truth labels from the entire data not overlapping with the unlabeled set used for AL.
As can be seen from Table~\ref{table:efficacy_of_query_set}, it is of interest to see that our self-validation set performs better than  the random validation set.
The two validation sets have a major difference in data distributions; 
the self-validation set mainly consists of the examples with the highest meta-scores among the remaining unlabeled data per round, while the random validation set consists of random examples. 
We conclude that the meta-score of \algname{} has the potential for constructing a high-quality validation set in addition to query selection.

\smallskip 
\textbf{Efficacy of Skyline Constraint.}\hspace{0.1cm}
Table~\ref{table:efficacy_of_skyline} demonstrates the final round test accuracy of \algname{} with or without the skyline constraint. For the latter, a standard 2-layer MLP is used as the meta-network architecture without any modification.
The performance of \algname{} degrades significantly without the skyline constraint, meaning that the non-constrained MLP can easily overfit to the small-sized self-validation set, thereby assigning high output scores on less-pure and less-informative examples.
Therefore, the violation of the skyline constraint in optimization makes \algname{} hard to balance between the purity and informativeness scores in query selection.

\begin{table*}[!t]
\caption{Efficacy of the skyline constraint.} 
\label{table:efficacy_of_skyline}
\centering
\begin{tabular}{c|c|c c c c}\toprule
\multicolumn{2}{c|}{Noise Ratio}&$10\%$&$20\%$&$40\%$&$60\%$\\\hline \addlinespace[0.15ex]
\multirow{2}{*}{\algname{}} 
&w/ skyline&\textbf{93.10}&\textbf{92.10}&\textbf{91.48}&\textbf{89.51}\\ \addlinespace[0.1ex] \cline{2-6} \addlinespace[0.2ex]
&w/o skyline&87.25&86.29&83.61&81.67\\\bottomrule
\end{tabular}
\vspace{0.3cm}
\end{table*}

\begin{table*}[t]
\def\arraystretch{1.3}
\caption{Efficacy of the meta-objective in \algname{}. We show the AL performance of two alternative balancing rules compared with \algname{} for the split-dataset setup on CIFAR10 with the open-set noise ratios of 20$\%$ and $40\%$.}
\label{table:efficacy_of_meta_objective}
\begin{center}
\scriptsize
\begin{tabular}{c|c|c|c c c c c c c c c c} \toprule
Dataset&Noise\,Ratio&Round& 1 & 2 & 3 & 4 & 5 & 6 & 7 & 8 & 9 & 10 \\\hline \addlinespace[0.15ex]
&\multirow{3}{*}{20$\%$}&$\mathcal{P}(x)+\mathcal{I}(x)$\!\!\!&\textbf{61.93}&\textbf{73.82}&76.16&80.65&82.61&85.73&87.44&88.86&89.21&89.49\\\Tstrut
& &\!\!\!$\mathcal{P}(x)\cdot\mathcal{I}(x)$\!\!\!&\!\!\!\textbf{61.93}\!\!\!&\!\!\!71.79\!\!\!&\!\!\!78.09\!\!\!&\!\!\!81.32\!\!\!&\!\!\!84.16\!\!\!&\!\!\!86.39\!\!\!&\!\!\!88.74\!\!\!&\!\!\!89.89\!\!\!&\!\!\!90.54\!\!\!&\!\!\!91.20\!\!\!\\\Tstrut
CIFAR10& &\!\!\!\!\algname{}\!\!\!\!&\!\!\!\textbf{61.93}\!\!\!&\!\!\!\textbf{73.82}\!\!\!&\!\!\!\textbf{80.58}\!\!\!&\!\!\!\textbf{84.72}\!\!\!&\!\!\!\textbf{87.31}\!\!\!&\!\!\!\textbf{89.20}\!\!\!&\!\!\!\textbf{90.52}\!\!\!&\!\!\!\textbf{91.46}\!\!\!&\!\!\!\textbf{91.93}\!\!\!&\!\!\!\textbf{92.10}\!\!\!\\\cline{2-13}\Tstrut
(4:6 split)&\multirow{3}{*}{\!\!\!40$\%$\!\!\!}&\!\!\!$\mathcal{P}(x)+\mathcal{I}(x)$\!\!\!&\!\!\!\textbf{59.31}\!\!\!&\!\!\!\textbf{72.50}\!\!\!&\!\!\!75.67\!\!\!&\!\!\!78.78\!\!\!&\!\!\!81.70\!\!\!&\!\!\!83.74\!\!\!&\!\!\!85.08\!\!\!&\!\!\!86.48\!\!\!&\!\!\!87.47\!\!\!&\!\!\!88.86\!\!\!\\\Tstrut
& &\!\!\!$\mathcal{P}(x)\cdot\mathcal{I}(x)$\!\!\!&\!\!\!\textbf{59.31}\!\!\!&\!\!\!66.37\!\!\!&\!\!\!73.57\!\!\!&\!\!\!77.85\!\!\!&\!\!\!81.37\!\!\!&\!\!\!84.22\!\!\!&\!\!\!86.80\!\!\!&\!\!\!88.04\!\!\!&\!\!\!88.73\!\!\!&\!\!\!89.11\!\!\!\\\Tstrut
& &\!\!\!\!\algname{}\!\!\!\!&\!\!\!\textbf{59.31}\!\!\!&\!\!\!\textbf{72.50}\!\!\!&\!\!\!\textbf{79.54}\!\!\!&\!\!\!\textbf{82.94}\!\!\!&\!\!\!\textbf{85.77}\!\!\!&\!\!\!\textbf{88.16}\!\!\!&\!\!\!\textbf{89.34}\!\!\!&\!\!\!\textbf{90.07}\!\!\!&\!\!\!\textbf{90.92}\!\!\!&\!\!\!\textbf{91.48}\!\!\!\\\bottomrule
\end{tabular}
\end{center}
\end{table*}

\textbf{Efficacy of Meta-objective.}\hspace{0.1cm}
{\algname{} keeps finding the best balance between purity and informativeness over multiple AL rounds by repeatedly minimizing the meta-objective in Equation \ref{eq:ranking_loss}. To validate its efficacy, we compare it with two simple alternatives based on heuristic balancing rules such as \emph{linear combination} and \emph{multiplication}, denoted as $\mathcal{P}(x)+\mathcal{I}(x)$ and $\mathcal{P}(x)\cdot\mathcal{I}(x)$, respectively.
Following the default setting of \algname{}, we use LL for $\mathcal{P}(x)$ and CSI for $\mathcal{I}(x)$.}

{Table \ref{table:efficacy_of_meta_objective} shows the AL performance of the two alternatives and \algname{} for the split-dataset setup on CIFAR10 with the noise ratios of 20$\%$ and $40\%$.
\algname{} beats the two alternatives 
after the second AL round where \algname{} starts balancing purity and informativeness with its meta-objective.
This result implies that our meta-objective successfully finds the best balance between purity and informativeness by emphasizing informativeness over purity at the later AL rounds.}

\subsection{Effect of Varying OOD Labeling Cost}
{
The labeling cost for OOD examples could vary with respect to data domains. To validate the robustness of \algname{} on diverse labeling scenarios, we conduct an additional study of adjusting the labeling cost $c_{OOD}$ for the OOD examples. Table \ref{table:effect_of_varying_cost} summarizes the performance change with four different labeling costs ({\em i.e.}, 0.5, 1, 2, and 4). The two standard AL methods, CONF and CORESET, and two open-set AL methods, CCAL and SIMILAR, are compared with \algname{}. Overall, 
\algname{} consistently outperforms the four baselines regardless of the labeling cost. Meanwhile, CCAL and SIMILAR are more robust to the higher labeling cost than CONF and CORESET; CCAL and SIMILAR, which favor high purity examples, query more IN examples than CONF and CORESET, so they are less affected by the labeling cost, especially when it is high.}

\subsection{In-depth Analysis of Various Purity Scores}
\label{sec:analysis_purity_scores}
The OSR performance of classifier-dependent OOD detection methods, {\em e.g.}, ReAct, degrades significantly if the classifier performs poorly\,\cite{vaze2021open}. Also, the OSR performance of self-supervised OOD detection methods, {\em e.g.}, CSI, highly depends on the sufficient amount of clean IN examples\,\cite{DGM, tack2020csi}.
Table~\ref{table:ood_detection_performance} shows the OOD detection performance of two OOD detectors, ReAct and CSI, over AL rounds with \algname{}.
Notably, at the earlier AL rounds, CSI is better than ReAct, meaning that self-supervised OOD detection methods are more robust than classifier-dependent methods when the amount of labeled data is small.
Thus, the versions of \algname{} using CSI as the purity score is better than those using ReAct, as shown in Section \ref{sec:ablation_study}.

\begin{table*}[t]
\caption{{Effect of varying the labeling cost.}}
\label{table:effect_of_varying_cost}
\centering
\begin{tabular}{c|c c c c}\toprule
$c_{OOD}$&0.5&1&2&4\\\midrule
CONF&91.05&88.69&86.25&80.06\\\cline{2-5}
CORESET&90.59&89.12&85.32&81.22\\\midrule 
CCAL&90.25&88.87&88.16&87.25\\\cline{2-5}
SIMILAR&91.05&88.69&87.95&86.52\\\midrule
\algname{}&\textbf{92.52}&\textbf{91.48}&\textbf{89.53}&\textbf{87.36}\\\bottomrule
\end{tabular}
\end{table*}

\begin{table*}[t]
\centering
\caption{OOD detection performance (AUROC) of two different OOD scores with \algname{}.}
\label{table:ood_detection_performance}
\centering
\begin{tabular}{c|c|c c c c c}\toprule
\multicolumn{2}{c|}{Dataset}&\multicolumn{5}{c}{CIFAR10 (4:6 split), 40$\%$ Noise}\\\hline\addlinespace[0.1ex]
\multicolumn{2}{c|}{Round}&2&4&6&8&10\\\addlinespace[0.15ex]\hline\addlinespace[0.3ex]
\multirow{2}{*}{\algname{}}&ReAct&0.615&0.684&0.776&0.819&0.849\\\cline{2-7}    &CSI&0.745&0.772&0.814&0.849&0.870\\\bottomrule
\end{tabular}
\end{table*}

\subsection{In-depth Analysis of CCAL and SIMILAR in a Low-noise Case}
\label{sec:why_low_challenging_appendix}
{In the low-noise case, the standard AL method, such as CONF, can query many IN examples even without careful consideration of purity. As shown in Table \ref{table:acc_in_count}, with 10$\%$ noise, the ratio of IN examples in the query set reaches 75.24$\%$ at the last AL round in CONF. This number is farily similar to 88.46$\%$ and 90.24$\%$ in CCAL and SIMILAR, respectively. In contrast, with the high-noise case (60$\%$ noise), the difference between CONF and CCAL or SIMILAR becomes much larger ({\em i.e.}, from 16.28$\%$ to 41.84$\%$ or 67.84$\%$). That is, considering mainly on purity (not informativeness) may not be effective with the low-noise case. Therefore, especially in the low-noise case, the two purity-focused methods, SIMILAR and CCAL, have the potential risk of overly selecting less-informative IN examples that the model already shows high confidence, leading to lower generalization performance than the standard AL methods.}

{In contrast, \algname{} outperforms the standard AL baselines by controlling the ratio of IN examples in the query set to be very high at the earlier AL rounds but moderate at the later AL rounds; \algname{} achieves a higher ratio of IN examples in the query set than CONF at every AL round, but the gap keeps decreasing.
Specifically, with 10$\%$ noise, the ratio of IN examples in the query set reaches 94.76$\%$ at the first AL round in \algname{}, which is higher than 87.52$\%$ in CONF, but it becomes 75.80$\%$ at the last AL round, which is very similar to 75.24$\%$ in CONF.
This observation means that \algname{} succeeds in maintaining the high purity of the query set and avoiding the risk of overly selecting less-informative IN examples at the later learning stage.
}

\begin{table*}[t!]
\def\arraystretch{1.15}
\caption{Test accuracy and ratio of IN examples in a query set for the split-dataset setup on CIFAR10 with open-set noise of 10$\%$ and 60$\%$. ``$\%$IN in $S_Q$'' means the ratio of IN examples in the query set.}
\label{table:acc_in_count}
\begin{center}
\small
\begin{tabular}{c|c|c|c c c c c c c c c c} \toprule
\!\!\!\!N.\,Ratio\!\!\!\!&\!\!\!\!Method\!\!\!\!&\!\!\!Round\!\!\!& 1 & 2 & 3 & 4 & 5 & 6 & 7 & 8 & 9 & 10 \\\hline \addlinespace[0.15ex]
\multirow{8}{*}{\makecell[c]{\!\!\!10$\%$\!\!\!}}&\multirow{2}{*}{\!\!\!CONF\!\!\!}&\!\!\!Acc\!\!\!&62.26&74.77&80.81&84.52&86.79&88.98&90.58&91.48&92.36&92.83\\
& &\!\!\!$\%$IN in $S_Q$\!\!\!&87.52&82.28&80.84&79.00&75.16&76.21&74.08&74.61&74.00&75.24\\\cline{2-13}\Tstrut
&\multirow{2}{*}{\!\!\!CCAL\!\!\!}&\!\!\!Acc\!\!\!&61.18&71.80&78.18&82.26&84.96&86.98&88.23&89.22&89.82&90.55\\
& &\!\!\!$\%$IN in $S_Q$\!\!\!&89.04&88.48&89.12&88.64&89.52&88.80&90.44&88.08&88.64&88.46\\\cline{2-13}\Tstrut
&\multirow{2}{*}{\!\!\!SIMILAR\!\!\!}&\!\!\!Acc\!\!\!&63.48&73.51&77.92&81.54&84.04&86.28&87.61&88.46&89.20&89.92\\
& &\!\!\!$\%$IN in $S_Q$\!\!\!&91.44&91.04&91.52&92.56&92.61&91.40&92.24&90.64&90.75&90.24\\\cline{2-13}\Tstrut
&\multirow{2}{*}{\!\!\!\algname{}\!\!\!}&\!\!\!Acc\!\!\!&61.59&73.30&80.36&84.88&87.91&90.10&91.26&92.23&92.90&93.10\\
& &\!\!\!$\%$IN in $S_Q$\!\!\!&94.76&93.28&88.84&86.96&82.04&79.60&77.24&76.92&79.00&75.80\\\hline \addlinespace[0.15ex]
\multirow{8}{*}{\makecell[c]{\!\!\!60$\%$\!\!\!}}&\multirow{2}{*}{\!\!\!CONF\!\!\!}&\!\!\!Acc\!\!\!&56.14&65.17&69.60&73.63&76.28&80.27&81.63&83.69&84.88&85.43\\
& &\!\!\!$\%$IN in $S_Q$\!\!\!&37.44&32.20&28.16&25.40&25.64&20.08&20.88&17.00&18.04&16.28\\\cline{2-13}\Tstrut
&\multirow{2}{*}{\!\!\!CCAL\!\!\!}&\!\!\!Acc\!\!\!&56.54&66.97&72.16&76.32&80.21&82.94&84.64&85.68&86.58&87.49\\
& &\!\!\!$\%$IN in $S_Q$\!\!\!&41.92&38.52&39.76&41.20&38.64&42.16&42.24&40.32&42.24&41.84\\\cline{2-13}\Tstrut
&\multirow{2}{*}{\!\!\!SIMILAR\!\!\!}&\!\!\!Acc\!\!\!&57.60&67.58&71.95&75.70&79.67&82.20&84.17&85.86&86.81&87.58\\
& &\!\!\!$\%$IN in $S_Q$\!\!\!&56.08&61.08&67.12&66.56&67.32&67.28&68.08&67.00&68.16&67.84\\\cline{2-13}\Tstrut
&\multirow{2}{*}{\!\!\!\algname{}\!\!\!}&\!\!\!Acc\!\!\!&54.87&68.49&75.84&80.16&83.37&85.64&87.56&88.43&89.26&89.51\\
& &\!\!\!$\%$IN in $S_Q$\!\!\!&82.80&79.92&65.88&55.40&52.00&47.52&46.60&41.44&36.52&35.64\\\bottomrule
\end{tabular}
\end{center}
\end{table*}

\subsection{AL Performance with More Rounds}

\begin{figure*}[t!]
\begin{center}
\includegraphics[width=0.85\linewidth]{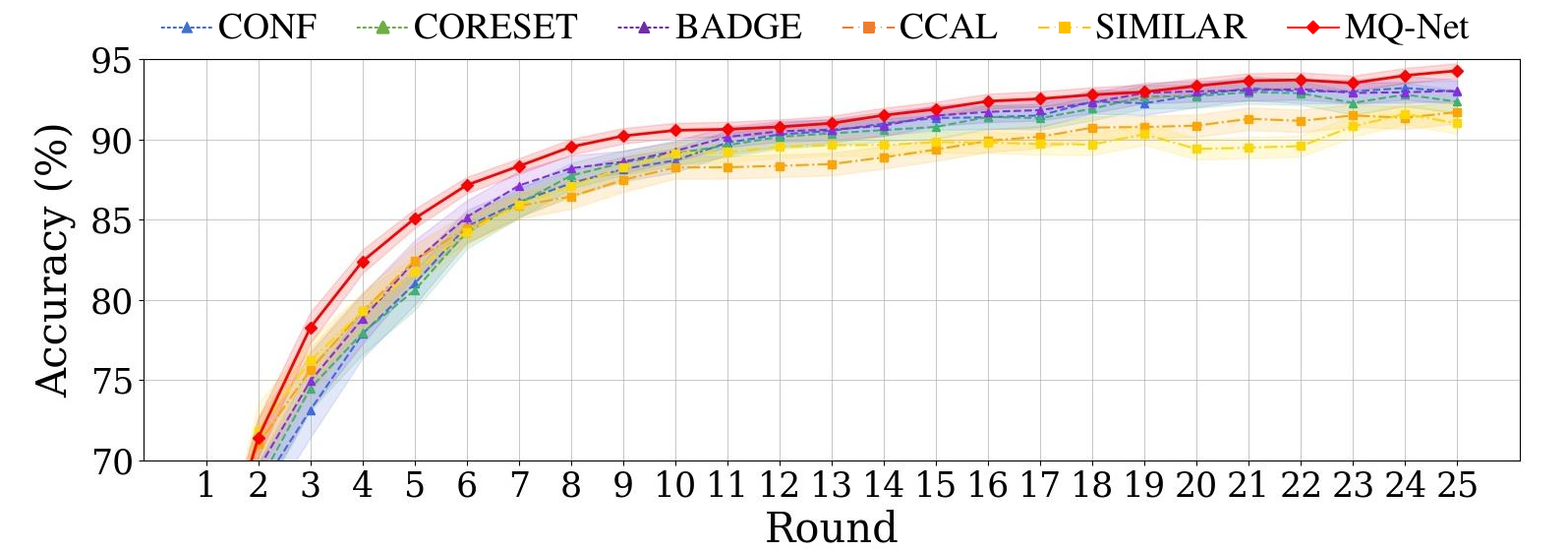}
\end{center}
\caption{Test accuracy over longer AL rounds for the split-dataset setup on CIFAR10 with an open-set noise ratio of 40$\%$. $500$ examples are selected as a query set in each AL round.}
\label{fig:AL_converge}
\end{figure*}

{Figure \ref{fig:AL_converge} shows the test accuracy over longer AL rounds for the split-dataset setup on CIFAR10 with an open-set noise ratio of 40$\%$. Owing to the ability to find the best balance between purity and informativeness, \algname{} achieves the highest accuracy on every AL round. The purity-focused approaches, CCAL and SIMILAR, lose their effectiveness at the later AL rounds, compared to the informativeness-focused approaches, CONF, CORESET, and BADGE; the superiority of CONF, CORESET, and BADGE over CCAL and SIMILAR gets larger as the AL round proceeds, meaning that fewer but highly-informative examples are more beneficial than more but less-informative examples for model generalization as the model performance converges.
However, with low\,({\em e.g.}, 20\%) open-set noise cases, most OOD examples are selected as a query set and removed from the unlabeled set in a few additional AL rounds, because the number of OOD examples in the unlabeled set is originally small. Thus, the situation quickly becomes similar to the standard AL setting.
}

\begin{figure*}[t!]
\begin{center}
\includegraphics[width=0.55\linewidth]{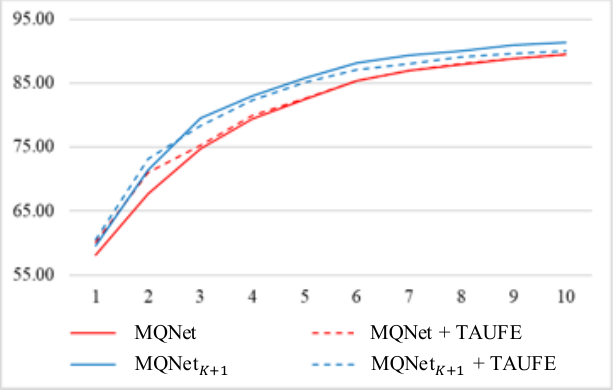}
\end{center}
\caption{Effect of using labeled OOD examples in the query set for model training of AL.}
\label{fig:AL_OOD_class}
\end{figure*}

\subsection{Effect of using labeled OOD examples in model training of AL}

Figure \ref{fig:AL_OOD_class} shows the AL performance when using the labeled OOD examples in the query set for model training of AL. We used the labeled OOD examples in two ways: (1) making the classifier be $k+1$-way classifier to predict the additional OOD class (blue lines), and (2) applying \algnamea{} in the penultimate layer activations (dotted lines).
Adding additional OOD class induces performance improvement throughout the AL rounds as it enhances the open-set recognition performance by leveraging the effect of outlier exposure (blue line).
Applying \algnamea{} induces performance improvement on the earlier AL rounds (red dotted line).
However, applying \algnamea{} on the $k+1$-way classifier with the OOD class degrades the performance as the effect of \algnamea{} conflicts with adding the OOD class.
\section{Conclusion and Future Work}
\label{sec:conclusion2}

\vspace{-0.2cm}
We propose \algname{}, a novel meta-model for open-set active learning that deals with the purity-informativeness dilemma.
\algname{} finds the best balancing between the two factors, being adaptive to the noise ratio and target model status.
A clean validation set for the meta-model is obtained for free by exploiting the procedure of active learning.
A ranking loss with the skyline constraint optimizes \algname{} to make the output a {legitimate} meta-score that keeps the obvious order of two examples.
\algname{} is shown to yield the best test accuracy throughout the entire active learning rounds, thereby empirically proving the correctness of our solution to the purity-informativeness dilemma. Overall, we expect that our work will raise the practical usability of active learning with open-set noise.

Although \algname{} outperforms other methods on multiple pairs of noisy datasets under the open-set AL settings, there are some issues that need to be further discussed. 
First, the performance gap between standard AL without open-set noise and open-set AL still exists. That is, we could not \emph{completely} eliminate the negative effect of open-set noise.
Second, although we validated \algname{} with many OOD datasets, its effectiveness may vary according to the types of the OOD datasets.
Formulating the effectiveness of \algname{} based on the characteristics of a given pair of IN and OOD datasets can be an interesting research direction.
Third, we regarded the OOD examples in a query set to be completely useless in training, but recent studies have reported that the OOD examples are helpful for model generalization\,\cite{park2021task, lee2021removing, lee2022weakly, wei2021open}.
Therefore, analyzing how to use OOD examples for model generalization and sample selection in AL can also be an interesting research direction.

\chapter{Prioritizing Informative Examples for Data Pruning from Labeled Noisy Data}
\label{chap:part_3}
\section{Overview}
\label{sec:overview3}
\vspace{-0.cm}

By virtue of ever-growing datasets and the neural scaling law\,\cite{hestness2017deep, kaplan2020scaling}, where the model accuracy often increases as a power of the training set size, modern deep learning has achieved unprecedented success in many domains, \emph{e.g.}, GPT\,\cite{brown2020language}, CLIP\,\cite{radford2021learning}, and ViT\,\cite{dosovitskiy2020image}.
With such massive datasets, however, practitioners often suffer from enormous computational costs for training models, tuning their hyper-parameters, and searching for the best architectures, which become the main bottleneck of development cycles. 
One popular framework to reduce these costs is \emph{data pruning}, which reduces a huge training set into a small subset while preserving model accuracy.
Notably, Sorscher et al.\,\cite{sorscher2022beyond} have shown that popular data pruning approaches can break down the neural scaling law from power-law to exponential scaling, meaning that one can reach a desired model accuracy with much fewer data.
Despite their great success, the impact of \emph{label noise} on data pruning has received little attention,
which is unavoidable in real-world data collection\,\cite{wei2021learning, li2017webvision, xiao2015learning}.

Noisy labels are widely known to severely degrade the generalization capability of deep learning, and thus numerous robust learning strategies have been developed to overcome their negative effect in deep learning\,\cite{song2022learning}.
Among them, \emph{Re-labeling}\,\cite{song2019selfie}, a family of methods that identify wrongly labeled examples and correct their labels during training by a self-correction module such as self-consistency regularization\,\cite{xie2020unsupervised}, has shown state-of-the-art performance.
For example, the performance of DivideMix\,\cite{li2020dividemix} trained on the CIFAR-10N\,\cite{wei2021learning} dataset containing real human annotation noise is nearly identical to that of a standard model trained on the clean CIFAR-10 dataset.
Consequently, it is evident that this excellent performance of re-labeling must be carefully considered when designing a framework for data pruning under label noise.

\begin{figure*}[t!]
\begin{center}
\includegraphics[width=0.95\textwidth]{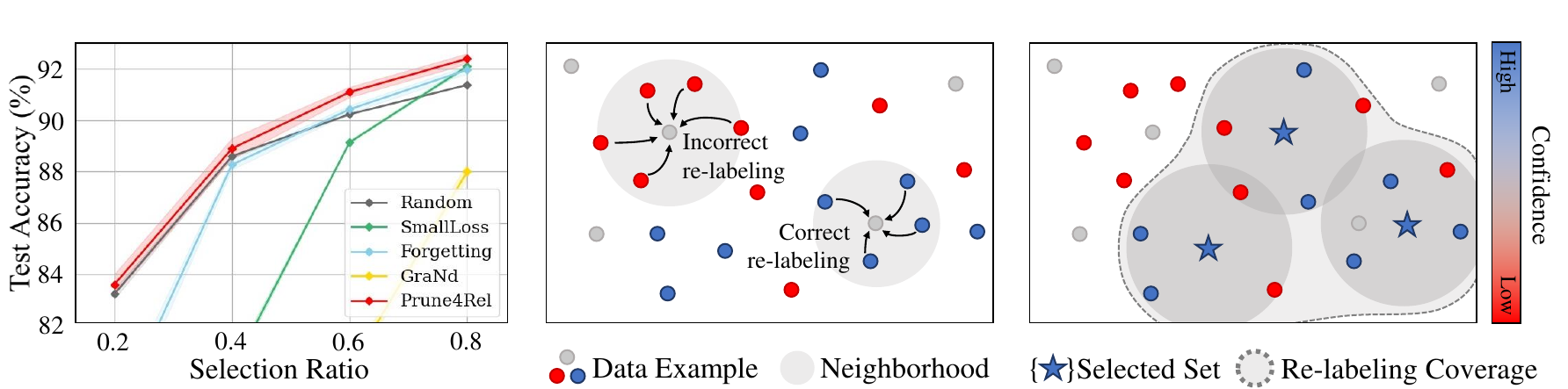}
\end{center}
\hspace*{0.8cm} {\small (a) Pruning Performance.} \hspace*{0.5cm} {\small (b) Re-labeling by Neighborhood.} \hspace*{0.3cm} {\small (c) Goal of \algnamec{}.}
\caption{Key idea of \algnamec{}: (a) shows data pruning performance of \algnamec{} and existing sample selection methods on CIFAR-10N with DivideMix; (b) shows how the neighborhood confidence affects the re-labeling correctness; (c) shows the goal of \algnamec{} that maximize the neighbor confidence coverage to the entire training set, thereby maximizing the re-labeling accuracy. }
\label{fig:key_idea}
\end{figure*}

In this paper, we formulate a new problem of \emph{data pruning with re-labeling} for a training set with noisy labels, which aims to maximize the generalization power of the selected subset with expecting that a large proportion of erroneous labels are self-corrected\,(\emph{i.e.}, re-labeled). 
Unfortunately, prior data pruning and sample selection algorithms are not suitable for our problem because the re-labeling capability is not taken into account, and have much room for improvement as shown in Figure \ref{fig:key_idea}(a). Popular data pruning approaches\,(denoted as Forgetting\,\cite{toneva2018empirical} and GraNd\,\cite{paul2021deep} in blue and yellow, respectively) favor hard\,(\emph{i.e.}, uncertain) examples because they are considered more beneficial for generalization\,\cite{guo2022deepcore}; however, because it is very difficult to distinguish between hard examples and incorrectly-labeled examples\,\cite{park2022meta}, many of the incorrectly-labeled examples can be included in the subset causing unreliable re-labeling. 
In addition, the small-loss trick\,\cite{han2018co}\,(denoted as SmallLoss in green) for sample selection favors easy examples because they are likely to be correctly labeled; however, they are not beneficial for generalization at a later stage of training. 
Therefore, this new problem necessitates the development of a new data pruning approach.

Accordingly, we suggest a completely novel approach of finding a subset of the training set such that \emph{the re-labeling accuracy of all training examples is preserved as much as possible with the model trained on the subset}. 
The first challenge in this direction is how to estimate whether each example can be re-labeled correctly even before fully training models on the candidate subset. 
The second challenge is how to find the subset that maximizes the overall re-labeling accuracy of the entire training set in an efficient manner.

Addressing these two challenges, we develop a novel framework, called \textbf{\algnamec{}}. 
For the first challenge, we define the concept of the \emph{neighborhood confidence} which is the sum of the prediction confidence of each neighbor example in the selected subset. 
We show that, as in Figure \ref{fig:key_idea}(b), an example with high neighborhood confidence is likely to be corrected by Re-labeling methods.
We further provide theoretical and empirical evidence of this argument.
For the second challenge, we show that the overall re-labeling accuracy is maximized by selecting a subset that maximizes the sum of its reachable neighborhood confidence for all training examples, as shown in Figure \ref{fig:key_idea}(c). 
Furthermore, because enumerating all possible subsets is a combinatorial optimization problem which is NP-hard\,\cite{killamsetty2021glister}, we provide an efficient greedy selection algorithm that expands the subset one by one by choosing the example that most increases the overall neighborhood confidence. 


Extensive experiments on {four} real noisy datasets, CIFAR-10N, CIFAR-100N, WebVision, and Clothing-1M, and one synthetic noisy dataset on ImageNet-1K show that \algnamec{} consistently outperforms the \emph{eight} data pruning baselines by up to 9.1\%.
Moreover, \algnamec{} with Re-labeling models significantly outperforms the data pruning baselines with a standard model by up to 21.6$\%$, which reaffirms the necessity of data pruning with re-labeling. 
\section{Methodology}
\label{sec:method3}


We formalize a problem of data pruning with re-labeling such that it finds the most informative subset $\mathcal{S}$, where a model $\theta_{\mathcal{S}}$ trained on $\mathcal{S}$ maximizes the re-labeling accuracy of the entire noisy training set $\mathcal{\tilde{D}}=\{(x_i, \tilde{y}_i)\}_{i=1}^{m}$\,\footnote{Maximizing the re-labeling accuracy is equivalent to correctly re-labeling all training examples. This formulation enables the model to exploit all clean labels for training, leading to a satisfactory generalization.}. Formally, we aim to find an optimal subset $\mathcal{S}^{*}$ such that
\begin{equation}
\begin{gathered}
\mathcal{S}^{*} = \argmaxB_{\mathcal{S}:\ |\mathcal{S}|\leq s} ~\sum_{(x,\tilde{y}) \in \mathcal{\tilde{D}}} \mathbbm{1}_ {[{f(x;{{\theta}_{\mathcal{S}}})~=~y^*}] } ~~ : ~~{\theta}_{\mathcal{S}}=\argminB_{{\theta}} ~~ \mathcal{L}_{Re\text{-}labeling}(\mathcal{S}; \theta,\!\mathcal{A}),
\end{gathered}
\label{eq:goal}
\end{equation}
where $y^*$ is the ground-truth label of a noisy example $x$, $f(x;{{\theta}_{\mathcal{S}}})\in \mathbb{R}^c$ is a $c$-way class prediction of the example $x$ from the Re-labeling model ${\theta}_{\mathcal{S}}$, and $s$ is the target subset size. 

Finding the optimal subset $\mathcal{S}^{*}$ through direct optimization of Eq.\,\eqref{eq:goal} is infeasible because the ground-truth label $y^*$ is unknown in practice. 
In addition, the subset should be found at the early stage of training, \emph{i.e.}, in a {warm-up} period, to reduce the computational cost\,\cite{guo2022deepcore}. 
To achieve these goals in an accurate and efficient way, in Section \ref{sec:label_correction_by_neighbor}, we first introduce a new metric, \emph{the reduced neighborhood confidence}, that enables estimating the re-labeling capacity of a subset even in the warm-up period. Then, in Section \ref{sec:ours}, we propose a new data pruning algorithm \emph{\algnamec{}} using this reduced neighborhood confidence to find a subset that maximizes the re-labeling accuracy.

\subsection{Reduced Neighborhood Confidence}
\label{sec:label_correction_by_neighbor}

As a measurement of estimating the re-labeling accuracy, we use the confidence of neighbor examples for each target noisy example $x$, because the noisy examples are known to be corrected by their \emph{clean neighbor} examples with self-consistency regularization\,\cite{englesson2021consistency}.
Specifically, once an augmentation of a noisy example has a similar embedding to those of other clean neighbors in the representation space, the self-consistency loss can force the prediction of the noisy example to be similar to those of other clean neighbors as a way of re-labeling. 
This property is also evidenced by a theory of re-labeling with a generalization bound\,\cite{wei2020theoretical}. 
Thus, the neighboring relationship among examples can be a clear clue to estimate the re-labeling accuracy even in the early stage of training.

We define a \emph{neighborhood} and its \emph{reduced neighborhood confidence} to utilize the relationship of neighboring examples in Definitions \ref{def:neighborhood} and \ref{def:pruned_neighbor_confidence}.

\smallskip
\begin{definition}{\sc (Neighborhood)}.
Let $\mathcal{B}(x_i)=\{x\!: ||\mathcal{A}(x_i)-x||\leq\epsilon \}$ be a set of all possible augmentations from the original example $x_i$ using an augmentation function $\mathcal{A}$. Then, given a noisy training set $\tilde{\mathcal{D}}$, a \emph{neighborhood} of $x_i$ is defined as $\mathcal{N}(x_i)=\{x\!\in\tilde{\mathcal{D}}\!:\mathcal{B}(x_i)\cap\mathcal{B}(x)\neq \emptyset \}$, which is the set of examples that are reachable by the augmentation $\mathcal{A}$. \qed
\label{def:neighborhood}
\end{definition}

\smallskip
\begin{definition}{\sc (Reduced Neighborhood Confidence)}.
The \emph{reduced neighborhood confidence} ${C}_{\mathcal{N}}(x_i; \mathcal{S})$ of an example $x_i$ is the sum of the prediction confidence ${C}(\cdot)$ of its neighbors $x_j\in\mathcal{N} (x_i)$ in a given reduced (\emph{i.e.}, selected) subset $\mathcal{S}$, which is formalized as
\begin{equation}
\begin{gathered}
{C}_{\mathcal{N}}(x_i;\mathcal{S}) = \sum_{x_j\in {\mathcal{S}}} \mathbbm{1}_{[x_j\in \mathcal{N}(x_i)]}\cdot {C}(x_j),
\end{gathered}
\label{eq:neighbor_confidence}
\end{equation} 
\label{def:pruned_neighbor_confidence}
\end{definition}
and its \emph{empirical reduced neighborhood confidence} is computed by using the cosine similarity among the augmentations of all possible pairs of example embeddings,
\begin{equation}
\begin{gathered}
\hat{{C}}_{\mathcal{N}}(x_i;\mathcal{S})=\sum_{x_j \in \mathcal{S}} \mathbbm{1}_{[{sim}(\mathcal{A}(x_i), \mathcal{A}(x_j)) \ge \tau]} \cdot{sim}\big(\mathcal{A}(x_i), \mathcal{A}(x_j)\big)\cdot {C}(x_j),
\end{gathered}
\label{eq:emp_rnc}
\end{equation}
where ${sim}(\cdot)$ is the cosine similarity between the augmentations $\mathcal{A}(x)$ of two different examples in the embedding space, and $\tau$ is a threshold to determine whether the two examples belong to the same neighborhood. Unlike Eq.\,\eqref{eq:neighbor_confidence}, Eq.\,\eqref{eq:emp_rnc} is calculated as a weighted sum of prediction confidences with cosine similarity to approximate the likelihood of belonging to the neighborhood. \qed

Based on these definitions, we investigate the theoretical evidence of employing reduced neighborhood confidence as a means to estimate the re-labeling capacity of a subset.


\smallskip
\noindent\textbf{Theoretical Evidence.} 
A subset $\mathcal{S}$ with a \emph{high} value of the \emph{total} reduced neighborhood confidence, the sum of the reduced neighborhood confidence of each example in ${\mathcal{S}}$, allows a Re-labeling model to maximize its re-labeling accuracy in the entire training set. 
We formally support this optimization by providing a theoretical analysis that extends the generalization bound in the prior re-labeling theory\,\cite{wei2020theoretical} to data pruning. 

\smallskip
\begin{assumption}{\sc (Expansion and Separation)}.
Following the assumption in \,\cite{wei2020theoretical}, the $\alpha$-expansion and $\beta$-separation assumptions hold for the training set $\tilde{\mathcal{D}}$. 
The $\alpha$-expansion means that an example is reachable to the $\alpha$ number of augmentation neighbors on average, \emph{i.e.}, $\mathbb{E}_{x\in\tilde{\mathcal{D}}} [ |\mathcal{N}(x)| ]=\alpha$.
The $\beta$-separation means that data distributions with different ground-truth classes are highly separable, such that the average proportion of the neighbors from different classes is as small as $\beta$. 
\label{assume:expansion}
\end{assumption}
Under these assumptions, we can obtain a training accuracy (error) bound of a Re-labeling model trained on a subset $S$ as in Theorem \ref{theorem:neighbor_confidence_to_label_correction}.

\smallskip
\begin{lemma}{\sc (Re-labeling Bound)}.
Suppose $\alpha$-expansion and $\beta$-separation assumptions hold for the training set $\tilde{\mathcal{D}}$. Then, for a Re-labeling minimizer $\theta_{\tilde{\mathcal{D}}}$ on $\tilde{\mathcal{D}}$, we have
\begin{equation}
{Err}(\theta_{\tilde{\mathcal{D}}}) \le {2\cdot {Err}(\theta_{\mathcal{M}}) \over \alpha-1} + {2 \cdot \alpha \over \alpha-1}\cdot\beta,
\label{eq:relabeling_bound}
\end{equation}
where ${Err}(\cdot)$ is a training error on ground-truth labels, and $\theta_{\mathcal{M}}$ is a model trained with the supervised loss in Eq.~\eqref{eq:relabeling} on a minimum (or given) clean set $\mathcal{M}\subset \mathcal{S}$.
\vspace*{-0.3cm}
\begin{proof}
Refer to \cite{wei2020theoretical} for the detailed concept and proof.
\end{proof}
\label{lemma:relabeling_bound}
\end{lemma}

\smallskip
\begin{theorem}
Assume that a subset $\mathcal{S}\in\tilde{\mathcal{D}}$ follows $\alpha_\mathcal{S}$-expansion and $\beta_\mathcal{S}$-separation, where $\alpha_\mathcal{S} \le \alpha$. Then, the training error of a Re-labeling model $\theta_{\mathcal{S}}$ trained on $\mathcal{S}$ is bounded by the inverse of the total reduced neighborhood confidence $\sum_{x\in{\tilde{\mathcal{D}}}} C_{\mathcal{N}}(x;\mathcal{S})$ such that,
\begin{equation}
\begin{gathered}
{Err}(\theta_{\mathcal{S}}) \le {{2 \cdot |\mathcal{S}| \cdot {Err}(\theta_{\mathcal{M}}) }\over \sum_{x\in{\tilde{\mathcal{D}}}} {C_\mathcal{N}(x;\mathcal{S})} } + {2 \cdot \alpha_{\mathcal{S}}\over {\alpha}_{\mathcal{S}}-1}\cdot{\beta_{\mathcal{S}}},
\end{gathered}
\label{eq:error_bound}
\end{equation}
where $\theta_{\mathcal{M}}$ is a model trained with the supervised loss in Eq.~\eqref{eq:relabeling} on a given clean set $\mathcal{M}\subset \mathcal{S}$.
\label{theorem:neighbor_confidence_to_label_correction}
\begin{proof}
The $\alpha$-expansion and $\beta$-separation assumptions hold for the training set $\tilde{\mathcal{D}}$. Then, following the re-labeling theory\,\cite{wei2020theoretical}, minimizing the self-consistency loss forces the classifier into correcting the erroneous labels and improving the training accuracy, as presented in Lemma~\ref{lemma:relabeling_bound}. 

This lemma is used for proving Theorem~\ref{theorem:neighbor_confidence_to_label_correction}.
Since $\alpha_{\mathcal{S}}$ indicates the average number of augmentation neighbors in $\mathcal{S}$, 
we can transform Eq.~\eqref{eq:relabeling_bound} using $\alpha_{\mathcal{S}}$,
\begin{equation}
\begin{gathered}
{Err}(\theta_{\mathcal{S}}) \le {{2 \cdot {Err}(\theta_{\mathcal{M}}) }\over (1/|\mathcal{S}|)\sum_{x\in{\mathcal{S}}} \mathbbm{1}_{[x^{\prime}\in\mathcal{N}(x)]} } + {2 \cdot \alpha_{\mathcal{S}}\over {\alpha}_{\mathcal{S}}-1}\cdot{\beta_{\mathcal{S}}}.
\end{gathered}
\label{eq:error_bound_eq10}
\end{equation}
Assume that the training error of the minimum clean set $\mathcal{M}$ in the selected subset $\mathcal{S}$ is proportional to the inverse of the confidence of $x\in\mathcal{S}$, since the performance of the standard learner is often correlated to the confidence of training examples. Then, Eq.~\eqref{eq:error_bound_eq10} becomes
\begin{equation}
\begin{gathered}
{Err}(\theta_{\mathcal{S}}) \le {{2\cdot |\mathcal{S}| \cdot {Err}(\theta_{\mathcal{M}}) }\over \sum_{x\in{\mathcal{S}}}C(x)  \sum_{x\in{\mathcal{S}}}  \mathbbm{1}_{[x^{\prime}\in\mathcal{N}(x)]} } + {2 \cdot \alpha_{\mathcal{S}}\over {\alpha}_{\mathcal{S}}-1}\cdot{\beta_{\mathcal{S}}} \\
\le {{2\cdot |\mathcal{S}| \cdot {Err}(\theta_{\mathcal{M}}) }\over \sum_{x\in{\mathcal{S}}} \mathbbm{1}_{[x^{\prime}\in\mathcal{N}(x)]} C(x)} + {2 \cdot \alpha_{\mathcal{S}}\over {\alpha}_{\mathcal{S}}-1}\cdot{\beta_{\mathcal{S}}},
\end{gathered}
\label{eq:error_bound_eq11}
\end{equation}
where the last inequality holds because of  Hölder's inequality with two sequence variables.
Therefore, ${Err}(\theta_{\mathcal{S}}) \le {{2 \cdot |\mathcal{S}| \cdot {Err}(\theta_{\mathcal{M}}) }\over \sum_{x\in{\tilde{\mathcal{S}}}} {C_\mathcal{N}(x;\mathcal{S})}  } + {2 \cdot \alpha_{\mathcal{S}}\over {\alpha}_{\mathcal{S}}-1}\cdot{\beta_{\mathcal{S}}}$, and this concludes the proof of Theorem~\ref{theorem:neighbor_confidence_to_label_correction}.
\end{proof}
\end{theorem}
Since $\beta_{\mathcal{S}}$ is usually very small, its effect on the error bound is negligible.
Then, the bound highly depends on the total reduced neighborhood confidence.
That is, as the total reduced neighborhood confidence increases, the error bound becomes tighter. 
This theorem supports that we can utilize the reduced neighborhood confidence for the purpose of maximizing the re-labeling accuracy.

\begin{figure}[t!]
\begin{center}
\includegraphics[width=7cm]{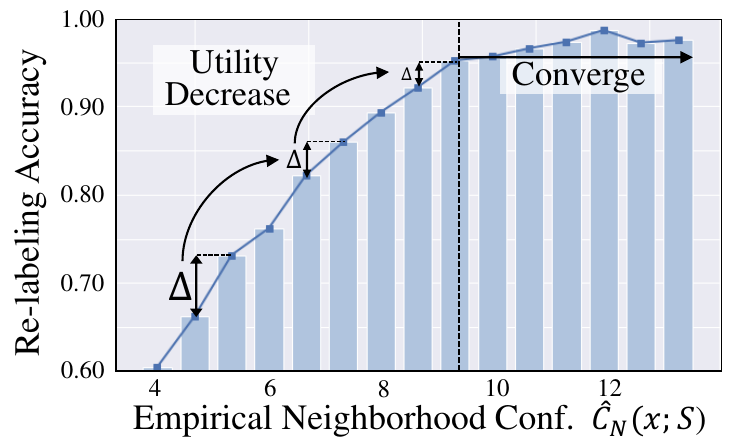}
\end{center}
\caption{Correlation between neighborhood confidence and re-labeling accuracy on a 20$\%$ randomly selected subset of CIFAR-10N.}
\label{fig:correlation}
\end{figure}

\smallskip
\noindent\textbf{Empirical Evidence.}
To empirically support Theorem \ref{theorem:neighbor_confidence_to_label_correction}, we validate the correlation between the empirical reduced neighborhood confidence\footnote{RandAug\,\cite{cubuk2020randaugment} is used as the augmentation function for the reduced neighborhood confidence.} and the re-labeling accuracy using CIFAR-10N, which is a real-world noisy benchmark dataset.

%
Specifically, we train DivideMix\,\cite{li2020dividemix} on the 20$\%$ randomly selected subset $\mathcal{S}$ for a warm-up training epoch of 10  and calculate the empirical reduced neighborhood confidence $\hat{{C}}_{\mathcal{N}}(x;\mathcal{S})$ for the entire training set.
Next, we fully train DivideMix\,\cite{li2020dividemix} on the random subset $\mathcal{S}$.
Last, we divide the entire training set into 15 bins according to the obtained $\hat{{C}}_{\mathcal{N}}(x;\mathcal{S})$ and verify the average re-labeling accuracy for each bin.

Figure \ref{fig:correlation} shows how the re-labeling accuracy changes according to the empirical reduced neighborhood confidence in Eq.~\eqref{eq:emp_rnc}. (The term ``empirical'' is simply omitted hereafter.)
%
The re-labeling accuracy shows a strong positive correlation with the reduced neighborhood confidence.
Interestingly, as the neighborhood confidence increases, its \emph{utility} in improving the re-labeling accuracy decreases, eventually reaching a convergence point after surpassing a certain threshold.

\subsection{Data Pruning by Maximizing Neighborhood Confidence Coverage}
\label{sec:ours}

We present a new data pruning algorithm called \emph{\algnamec{}} which optimizes the total reduced neighborhood confidence defined in Eq.\,\eqref{eq:emp_rnc}. This objective is equivalent to identifying a subset that maximizes the re-labeling accuracy on the entire training set, as justified in Theorem \ref{theorem:neighbor_confidence_to_label_correction}. 
%
Therefore, the objective of \algnamec{} is to find the subset $\mathcal{S}^{*}$, which is formulated as
\begin{equation}
\begin{gathered}
\mathcal{S}^{*} = \argmaxB_{\mathcal{S}:~|\mathcal{S}|\le s} \sum_{x_i \in \tilde{\mathcal{D}}} \bm{\sigma}\big( \hat{{C}}_{\mathcal{N}}(x_i;\mathcal{S}) \big),
\end{gathered}
\label{eq:objective3}
\end{equation}
where $\bm{\sigma}(z)$ is a \emph{utility} function of the reduced neighborhood confidence $\hat{{C}}_{\mathcal{N}}(x_i;\mathcal{S})$ in improving the re-labeling accuracy.
By the observation in Figure \ref{fig:correlation}, we define $\bm{\sigma}(z)$ as a \emph{non-decreasing} and \emph{concave} function where $\bm{\sigma}(0)=0$. 
In our implementation, we use the positive of the $tanh$ function as the utility function, \emph{i.e.}, $\bm{\sigma}(z)=tanh(z)$.
However, directly solving Eq.~\eqref{eq:objective3} is computationally expensive and impractical due to its NP-hard nature as a Set-Cover problem\,\cite{killamsetty2021glister}.
Accordingly, we employ an approximation solution to efficiently address this combinatorial optimization.

\begin{algorithm}[H]
\caption{Greedy Neighborhood Confidence ($\text{\algnamec{}}$)}
\label{alg:greedy}
    \begin{algorithmic}[1]
    {\small 
        \REQUIRE $\tilde{\mathcal{D}}$: training set, $s$: target subset size, and $C(x)$: confidence from warm-up classifier
        \STATE Initialize
        $\mathcal{S}\leftarrow\emptyset; \forall x\in \tilde{\mathcal{D}},~\hat{C}_{\mathcal{N}}(x)=0$
        \STATE \textbf{repeat}
        \STATE ~~$x\!=\!\argmaxB_{x\in\tilde{\mathcal{D}}\backslash\mathcal{S}}                \!\bm{\sigma}(\hat{C}_\mathcal{N}(x)\!+\!C(x))\!-\!\bm{\sigma}(\hat{C}_\mathcal{N}(x))$
        \STATE ~~{$\mathcal{S}\!=\!\mathcal{S}\cup\{x\}$}
        \STATE ~~\textbf{for all} $v\in \tilde{\mathcal{D}}$ ~\textbf{do}
        \STATE \quad ~$\hat{C}_\mathcal{N}(v)~+\!\!=\mathbbm{1}_{[sim(x,v)\ge \tau]}\cdot sim(x,v)\cdot C(x)$ 
        \STATE \textbf{until} $|\mathcal{S}|=s$
        \ENSURE Final selected subset $\mathcal{S}$
    }
    \end{algorithmic}
\end{algorithm}

\begin{algorithm}[H]
\caption{Greedy Balanced Neighborhood Confidence ($\text{\algnamec{}}_B$)}
\label{alg:greedy_balance}
    \begin{algorithmic}[1]
    {\small 
        \REQUIRE $\tilde{\mathcal{D}}$: training set, $\tilde{\mathcal{D}}_j(\subset\tilde{\mathcal{D}})$: set of training examples with a $j$-th class, $s$: target subset size, and $C(x)$: confidence of $x$ calculated from a warm-up classifier
        \STATE Initialize $\mathcal{S}\leftarrow\emptyset; \forall x\in \tilde{\mathcal{D}},~\hat{C}_{\mathcal{N}}(x)=0$ 
        \STATE \textbf{while} $|\mathcal{S}|<s$ \textbf{do}
        \STATE \quad\textbf{for} $j=1$ to $c$ \textbf{do}
        \STATE \quad\quad $x\!=\!\argmaxB_{x\in\tilde{\mathcal{D}}_j\backslash\mathcal{S}} ~ \bm{\sigma}(\hat{C}_\mathcal{N}(x)\!+\!C(x))\!-\!\bm{\sigma}(\hat{C}_\mathcal{N}(x))$
        \STATE \quad\quad {$\mathcal{S}\!=\!\mathcal{S}\cup\{x\}$}
        \STATE \quad\quad \textbf{for all} $v\in \tilde{\mathcal{D}}$ ~\textbf{do}
        \STATE \quad\quad \quad$\hat{C}_\mathcal{N}(v)~+\!\!=\mathbbm{1}_{[sim(x,v)\ge \tau]}\cdot sim(x,v)\cdot C(x)$ 
        \STATE \quad\quad \textbf{if} $|\mathcal{S}|=s$
        \STATE \quad\quad \quad \textbf{return} $\mathcal{S}$
        \STATE \textbf{end}
        \ENSURE Final selected subset $\mathcal{S}$
    }
    \end{algorithmic}
\end{algorithm}

\noindent\textbf{Optimization with Greedy Approximation.}
We present a practical solution for solving the optimization problem stated in Eq.~\eqref{eq:objective3} using a {greedy} approximation. The objective function satisfies both the \emph{monotonicity} and \emph{submodularity} conditions, indicating that the return of the objective function monotonically increases and the marginal benefit of adding an example decreases as the subset grows.
Therefore, a greedy sample selection can be employed as in Algorithm \ref{alg:greedy}.
In detail, we begin with an empty set $\mathcal{S}$ and initialize the reduced neighborhood confidence ${\hat{C}_{\mathcal{N}}}$ to $0$\,(lowest confidence) for all training examples (Line 1).
Next, at every step, we select an example $x$ that maximizes the marginal benefit $\bm{\sigma}(\hat{C}_\mathcal{N}(x)\!+\!C(x))\!-\!\bm{\sigma}(\hat{C}_\mathcal{N}(x))$ of Eq.~\eqref{eq:objective3}, and update the reduced neighborhood confidence ${\hat{C}_{\mathcal{N}}}$ based on the similarity scores (Lines 3--7).

To further improve robustness and efficiency, we introduce a class-balanced version, $\text{\algnamec{}}_B$, of which the detailed process is elaborated in Algorithm \ref{alg:greedy_balance}.
We first divide the entire training set into $c$ groups according to the noisy label of each example, under the assumption that the number of correctly labeled examples is much larger than that of incorrectly labeled examples in practice\,\cite{wei2021learning}.
Similar to Algorithm \ref{alg:greedy}, we begin with an empty set $\mathcal{S}$ and initialize the reduced neighborhood confidence ${\hat{C}_{\mathcal{N}}}$ to $0$ for each training example (Line 1). 
Then, by iterating class $j$, we select an example $x$ that maximizes the marginal benefit $\bm{\sigma}(\hat{C}_\mathcal{N}(x)\!+\!C(x))\!-\!\bm{\sigma}(\hat{C}_\mathcal{N}(x))$ within the set $\tilde{\mathcal{D}}_j(\subset\tilde{\mathcal{D}})$ and add it to the selected subset $\mathcal{S}$ (Lines 3--5).
Next, we update the reduced neighborhood confidence ${\hat{C}_{\mathcal{N}}}$ of each example in the entire training set by using the confidence and the similarity score to the selected example $x$ (Lines 6--7).
We repeat this procedure until the size of the selected subset $\mathcal{S}$ meets the target size $s$ (Lines 8--9).

In Theorem \ref{theorem:epsilon_guarantee}, we guarantee the selected subset $\mathcal{S}$ obtained by our greedy solution achieves a $(1-1/e)$-approximation of the optimum.

\begin{theorem}
Since Eq.~\eqref{eq:objective3}, denoted as $OBJ$, is a monotone, submodular, and non-negative function on $x$, the greedy solution provides a set with a $(1-1/e)$-approximation of the optimum. Formally,
\begin{equation}
OBJ(\mathcal{S})\ge (1-1/e)\cdot OBJ(\mathcal{S}^{*}).
\label{eq:epsilon_guarantee}
\end{equation}
\label{theorem:epsilon_guarantee}
\begin{proof}

We complete Theorem~\ref{theorem:epsilon_guarantee} by proving the \emph{monotonicity} and \emph{submodularity} of Eq.~\eqref{eq:objective3} in Lemmas~\ref{lemma:monotonicity} and \ref{lemma:submodularity}, under the widely proven fact that the monotonicity and submodularity of a combinatorial objective guarantee the greedy selection to get an objective value within $(1-1/e)$ of the optimum\,\cite{feige1998threshold}.

\medskip
\begin{lemma}{\sc (Monotonicity)}.
Our data pruning objective in Eq.~\eqref{eq:objective3}, denoted as $OBJ$, is monotonic. Formally,
\begin{equation}
\forall ~ \mathcal{S}\subset \mathcal{S}^{\prime}, ~~ OBJ(\mathcal{S}) \le OBJ(\mathcal{S}^{\prime}).
\label{eq:monotonicity}
\end{equation}
\vspace*{-0.7cm}
\begin{proof}
\begin{equation}
{\small
\begin{gathered}
OBJ(\mathcal{S}^{\prime}) = \sum_{x_i \in \tilde{\mathcal{D}}} \bm{\sigma}\big( \hat{{C}}_{\mathcal{N}}(x_i; \mathcal{S}^{\prime}) \big)=
\sum_{x_i \in \tilde{\mathcal{D}}} \bm{\sigma}\big( \sum_{x_j \in \mathcal{S}^{\prime}}\!\mathbbm{1}_{[{sim}(x_i, x_j) \ge \tau]} \cdot{sim}\big(x_i, x_j\big)\cdot {C}(x_j) ~~\big)
\\
=\!\sum_{x_i \in \tilde{\mathcal{D}}}\!\bm{\sigma}\big(\!\sum_{x_j \in \mathcal{S}}\!\mathbbm{1}_{[{sim}(x_i, x_j) \ge \tau]}\!\cdot{sim}\big(x_i, x_j\big)\!\cdot\!{C}(x_j)+\!\!\!\!\!\sum_{x_j \in \mathcal{S}^{\prime}\setminus\mathcal{S} }\!\!\!\!\mathbbm{1}_{[{sim}(x_i, x_j) \ge \tau]}\!\cdot\!{sim}\big(x_i, x_j\big)\!\cdot\!{C}(x_j)\big)
\\
\ge \sum_{x_i \in \tilde{\mathcal{D}}} \bm{\sigma}\big( \sum_{x_j \in \mathcal{S}}\mathbbm{1}_{[{sim}(x_i, x_j) \ge \tau]} \cdot{sim}\big(x_i, x_j\big)\cdot {C}(x_j)\big)=\sum_{x_i \in \tilde{\mathcal{D}}} \bm{\sigma}\big( \hat{{C}}_{\mathcal{N}}(x_i; \mathcal{S}) \big) = OBJ(\mathcal{S}),
\end{gathered}
}
\label{eq:monotonicity_proof}
\end{equation}
where the inequality holds because of the non-decreasing property of the utility function $\sigma$. 
Therefore, $OBJ(\mathcal{S}) \le OBJ(\mathcal{S}^{\prime})$.
\end{proof}
\label{lemma:monotonicity}
\end{lemma}

\begin{lemma}{\sc (Submodularity)}.
Our objective in Eq.~\eqref{eq:objective3} is submodular. Formally,
\begin{equation}
\forall ~ \mathcal{S}\subset \mathcal{S}^{\prime} ~and~~ \forall x\notin \mathcal{S}^{\prime}, ~~ OBJ(\mathcal{S}\cup \{x \})-OBJ(\mathcal{S}) \ge OBJ(\mathcal{S}^{\prime}\cup \{x \})-OBJ(\mathcal{S}^{\prime}).
\label{eq:submodularity}
\end{equation}
\vspace*{-0.7cm}
\begin{proof}
For notational simplicity, let $x_i$ be $i$, $x_j$ be $j$, and $\mathbbm{1}_{[{sim}(x_i, x_j) \ge \tau]}\!\cdot{sim}\big(x_i, x_j\big)\!\cdot\!{C}(x_j)$ be $C_{ij}$. Then, Eq.~\eqref{eq:submodularity} can be represented as
\begin{equation}
{\small
\begin{gathered}
\sum_{i \in \tilde{\mathcal{D}}} \bm{\sigma}\big( \sum_{j \in \mathcal{S}}C_{ij} +C_{ix}\big) - \sum_{i \in \tilde{\mathcal{D}}} \bm{\sigma}\big( \sum_{j \in \mathcal{S}}C_{ij}\big) \ge
\sum_{i \in \tilde{\mathcal{D}}} \bm{\sigma}\big( \sum_{j \in \mathcal{S}^{\prime}}C_{ij} +C_{ix}\big) - \sum_{i \in \tilde{\mathcal{D}}} \bm{\sigma}\big( \sum_{j \in \mathcal{S}^{\prime}}C_{ij}\big).
\end{gathered}
}
\label{eq:submodularity_simplified}
\end{equation}
Proving Eq.~\eqref{eq:submodularity_simplified} is equivalent to proving the decomposed inequality for each example $x_i\in\tilde{\mathcal{D}}$,
\begin{equation}
{\small
\begin{gathered}
\bm{\sigma}\big( \sum_{j \in \mathcal{S}}C_{ij} +C_{ix}\big) - \bm{\sigma}\big( \sum_{j \in \mathcal{S}}C_{ij}\big) \ge
\bm{\sigma}\big( \sum_{j \in \mathcal{S}^{\prime}}C_{ij} +C_{ix}\big) - \bm{\sigma}\big( \sum_{j \in \mathcal{S}^{\prime}}C_{ij}\big) \\
=\bm{\sigma}\big( \sum_{j \in \mathcal{S}}C_{ij}+ \sum_{j \in \mathcal{S}^{\prime}\setminus \mathcal{S}}C_{ij} +C_{ix}\big) - \bm{\sigma}\big( \sum_{j \in \mathcal{S}}C_{ij}+ \sum_{j \in \mathcal{S}^{\prime}\setminus \mathcal{S}}C_{ij}\big)
.
\end{gathered}
}
\label{eq:submodularity_decomposed}
\end{equation}
Since $\mathcal{S}$, $\mathcal{S}^{\prime}\!\setminus\!\mathcal{S}$, and $\{ x \}$ do not intersect each other, we can further simplify Eq.~\eqref{eq:submodularity_decomposed} with independent scala variables such that
\begin{equation}
{\small
\begin{gathered}
\bm{\sigma}\big( a + \epsilon \big) - \bm{\sigma}\big( a \big) \ge
\bm{\sigma}\big( a + b + \epsilon\big) - \bm{\sigma}\big( a + b \big),
\end{gathered}
}
\label{eq:submodularity_abe}
\end{equation}
where $a=\sum_{j \in \mathcal{S}}C_{ij}$, $b=\sum_{j \in \mathcal{S}^{\prime}\setminus \mathcal{S}}C_{ij}$, and $\epsilon=C_{ix}$. 

Since the utility function $\sigma$ is \emph{concave}, by the definition of concavity,
\begin{equation}
{\small
\begin{gathered}
{\bm{\sigma}\big( a + \epsilon \big) - \bm{\sigma}\big( a \big) \over (a+\epsilon - a) } \ge
{\bm{\sigma}\big( a + b + \epsilon\big) - \bm{\sigma}\big( a + b \big) \over (a+b+\epsilon - (a+b))}.
\end{gathered}
}
\label{eq:concavity}
\end{equation}
The denominators of both sides of the inequality become $\epsilon$, and Eq.\ \eqref{eq:concavity} can be transformed to Eq.~\eqref{eq:submodularity_abe}. Therefore, Eq.~\eqref{eq:submodularity_abe} should hold, and $OBJ(\mathcal{S}\cup \{x \})-OBJ(\mathcal{S}) \ge OBJ(\mathcal{S}^{\prime}\cup \{x \})-OBJ(\mathcal{S}^{\prime})$.
\end{proof}
\label{lemma:submodularity}
\end{lemma}

By Lemmas~\ref{lemma:monotonicity} and \ref{lemma:submodularity}, the monotonicity and submodularity of Eq.~\eqref{eq:objective3} hold. Therefore, Eq.~\eqref{eq:epsilon_guarantee} naturally holds, and this concludes the proof of Theorem \ref{theorem:epsilon_guarantee}.
\end{proof}
\end{theorem}

\textbf{Time Complexity Analysis.}
We analyze the time complexity of our greedy approximation in Algorithm \ref{alg:greedy}.
At each step, \algnamec{} takes the time complexity of $O(m\log m)$ + $O(m d)$, where $m$ is the training set size and $d$ is the embedding dimension size of the warm-up classifier.
Specifically, in Line 3, sampling an example with the largest marginal benefit of confidence takes $O(m\log m)$, and in Lines 5--6, updating the reduced neighborhood confidence of all training examples takes $O(m d)$.
In addition, with $\text{\algnamec{}}_B$, the time complexity is reduced to $O(m\log (m/c))$ + $O(m d)$ because it iteratively selects the example with the largest marginal benefit within each class subset, which is lower than the time complexity of a similar distance-based data pruning work, kCenterGreedy\,\cite{sener2018active} aiming to maximize the \emph{distance} coverage of a selected subset to the entire training set by a greedy approximation.
At each iteration, kCenterGreedy's runtime is $O(mk_t)$, where $k_t$ is the size of the selected set at iteration $t$\,\cite{citovsky2021batch}.
Note that, its time complexity increases as the subset size grows, which hinders its usability on a large-scale dataset.
In Section \ref{sec:results}, we empirically show that \algnamec{} is scalable to prune Clothing-1M, a large dataset with 1M examples, whereas kCenterGreedy is not.

\section{Experiments}
\label{sec:experiment3}

\subsection{Experiment Setting}
\label{sec:experiment_setting3}

\noindent\textbf{Datasets.} 
We first perform the data pruning task on four \emph{real} noisy datasets, CIFAR-10N, CIFAR-100N, Webvision, and Clothing-1M.
CIFAR-10N and CIFAR-100N\,\cite{wei2021learning} contain human re-annotations of 50K training images in the original CIFAR-10 and CIFAR-100\,\cite{krizhevsky2009learning}.
Specifically, each training image in CIFAR-10N contains three noisy labels, called Random 1,2,3, which are further transformed into the Worst-case label. Each image in CIFAR-100N contains one noisy label.
WebVision\,\cite{li2017webvision} and Clothing-1M\,\cite{xiao2015learning} are two large-scale noisy datasets. 
WebVision contains 2.4M images crawled from the Web using the 1,000 concepts in ImageNet-1K\,\cite{deng2009imagenet}. 
Following prior work\,\cite{chen2019understanding}, we use mini-WebVision consisting of the first 50 classes of the Google image subset with approximately 66K training images.
Clothing-1M consists of 1M training images with noisy labels and 10K clean test images collected from online shopping websites.

Additionally, a large-scale \emph{synthetic} noisy dataset, which we call ImageNet-N, is included in our experiments.
It consists of 1.2M training images, which are the training images of ImageNet-1K\,\cite{deng2009imagenet} with asymmetric label noise. 
Since ImageNet-1K is a clean dataset with no known real label noise, we inject the synthetic label noise to construct ImageNet-N.
Specifically, we inject \emph{asymmetric} label noise to mimic real-world label noise following the prior noisy label literature\,\cite{song2022learning}.
When a target noise ratio of ImageNet-N is $r\%$, we randomly select $r\%$ of the training examples for each class $c$ in ImageNet-1K and then flip their label into class $c+1$, \emph{i.e.}, class 0 into class 1, class 1 into class 2, and so on. 
This flipping is reasonable because consecutive classes likely belong to the same high-level category.
For the selected examples with the last class 1000, we flip their label into class 0.

\noindent\textbf{Algorithms.}
We compare \algnamec{} with a random selection from a uniform distribution, Uniform, a clean sample selection algorithm, SmallLoss\,\cite{jiang2018mentornet}, and six data pruning algorithms including Margin\,\cite{coleman2019selection}, $k$-CenterGreedy\,\cite{sener2018active}, Forgetting\,\cite{toneva2018empirical}, GraNd\,\cite{paul2021deep}, SSP\,\cite{sorscher2022beyond}, and Moderate\,\cite{xia2022moderate}.
SmallLoss favors examples with a small loss. For data pruning algorithms, (1) Margin selects examples in the increasing order of the difference between the highest and the second highest softmax probability; (2) $k$-CenterGreedy selects $k$ examples that maximize the distance coverage to the entire training set;
(3) Forgetting selects examples that are easy to be forgotten by the classifier throughout the warm-up training epochs;
(4) GraNd uses the average norm of the gradient vectors to measure the contribution of each example to minimizing the training loss;
(5) SSP leverages a self-supervised pre-trained model to select the most prototypical examples;
and (6) Moderate aims to select moderately hard examples using the distances to the median.


\noindent\textbf{Implementation Details.}
We train two representative Re-labeling models, DivideMix\,\cite{li2020dividemix} and SOP+\,\cite{liu2022robust} for our experiments. The hyperparameters for DivideMix and SOP+ are favorably configured following the original papers. 
Following the prior Re-labeling work\,\cite{li2020dividemix, liu2022robust}, for CIFAR-10N and CIFAR-100N, PreAct Resnet-18\,\cite{he2016identity} is trained for 300 epochs using SGD with a momentum of 0.9, a weight decay of 0.0005, and a batch size of 128. The initial learning rate is 0.02, and it is decayed with a cosine annealing scheduler.
For WebVision, InceptionResNetV2\,\cite{szegedy2017inception} is trained for 100 epochs with a batch size of 32.
For Clothing-1M, we use ResNet-50\,\cite{he2016deep} pre-trained on ImageNet and fine-tune it for 10 epochs with a batch size of 32. 
The initial learning rates of WebVision and Clothing-1M are 0.02 and 0.002, which are dropped by a factor of 10 at the halfway point of the training epochs.
For ImageNet-N, ResNet-50\,\cite{he2016deep} is trained for 50 epochs with a batch size of 64 and an initial learning rate of 0.02 decayed with a cosine annealing scheduler.

For data pruning algorithms, following prior work\,\cite{guo2022deepcore}, we perform sample selection after 10 warm-up training epochs for CIFAR-10N, WebVision, and ImageNet-N, and 30 warm-up epochs for CIFAR-100N. 
For Clothing-1M, we perform the sample selection after 1 warm-up training epoch from the ImageNet pre-trained ResNet-50.
The hyperparameters for all data pruning methods are favorably configured following the original papers.
For \algnamec{}, we set its hyperparameter $\tau$ to 0.975 for CIFAR-10N, to 0.95 for CIFAR-100N, WebVision, and ImageNet-N, and to 0.8 for Clothing-1M.
All methods are implemented with PyTorch 1.8.0 and executed on NVIDIA RTX 3080 GPUs.
The code is available at \url{https://github.com/kaist-dmlab/Prune4Rel}.

\noindent\textbf{Hyperparameter Configuration.}
\def\arraystretch{1.15}
\begin{table*}[t!]
\caption{Summary of the hyperparameters for training SOP+ and DivideMix on the CIFAR-10N/100N, Webvision, and Clothing-1M datasets.}
\scriptsize
\centering
\begin{tabular}{lc|c|c|c|c}
\toprule
\multicolumn{2}{c|}{\multirow{1}{*}{\textbf{Hyperparamters}}} & \multirow{1}{*}{\textbf{CIFAR-10N}} & \multirow{1}{*}{\textbf{CIFAR-100N}} & \multirow{1}{*}{\textbf{WebVision}} & \multirow{1}{*}{\textbf{Clothing-1M}} \\
\midrule
\multicolumn{1}{l|}{\multirow{7}{*}{\textbf{\begin{tabular}[c]{@{}c@{}}Training\\ Configuration\end{tabular}}}} & \textbf{architecture} & PreActResNet18 & PreActResNet18 & InceptionResNetV2 & \begin{tabular}[c]{@{}c@{}}ResNet-50 \\ (pretrained)\end{tabular} \\
\multicolumn{1}{l|}{} & \textbf{warm-up epoch} & 10 & 30 & 10 & 0 \\ 
\multicolumn{1}{l|}{} & \textbf{training epoch} & 300 & 300 & 100 & 10 \\
\multicolumn{1}{l|}{} & \textbf{batch size} & 128 & 128 & 32 & 32 \\
\multicolumn{1}{l|}{} & \textbf{learning rate\,(lr)} & 0.02 & 0.02 & 0.02 & 0.002 \\
\multicolumn{1}{l|}{} & \textbf{lr scheduler} & Cosine Annealing & Cosine Annealing & MultiStep-50th & MultiStep-5th \\
\multicolumn{1}{l|}{} & \textbf{weight decay} & $5 \times 10^{-4}$ & $5 \times 10^{-4}$ & $5 \times 10^{-4}$ & 0.001 \\
\midrule
\multicolumn{1}{c|}{\multirow{5}{*}{\textbf{DivideMix}}} & \textbf{$\lambda_{U}$} & 1 & 1 & \multicolumn{1}{l|}{} & 0.1 \\
\multicolumn{1}{c|}{} & {$\kappa$} & 0.5 & 0.5 & \multicolumn{1}{l|}{} & 0.5 \\
\multicolumn{1}{c|}{} & $T$ & 0.5 & 0.5 & \multicolumn{1}{c|}{--} & 0.5 \\
\multicolumn{1}{c|}{} & {$\gamma$} & 4 & 4 & \multicolumn{1}{l|}{} & 0.5 \\
\multicolumn{1}{c|}{} & $M$ & 2 & 2 & \multicolumn{1}{l|}{} & 2 \\
\midrule
\multicolumn{1}{c|}{\multirow{4}{*}{\textbf{SOP+}}}& \textbf{$\lambda_{C}$} & 0.9 & 0.9 & 0.1 & \multirow{4}{*}{--} \\
\multicolumn{1}{c|}{} & \textbf{$\lambda_{B}$} & 0.1 & 0.1 & 0 & \\
\multicolumn{1}{c|}{} & {lr for $u$} & 10 & 1 & 0.1 & \\
\multicolumn{1}{c|}{} & {lr for $v$} & 100 & 100 & 1 & \\
\bottomrule
\end{tabular}
\label{table:experiment_detail}
\end{table*}
Table \ref{table:experiment_detail} summarizes the overall training configurations and hyperparameters used to train the two Re-labeling models, DivideMix and SOP+.
The hyperparameters for DivideMix and SOP+ are favorably configured following the original papers. 
DivideMix\,\cite{li2020dividemix} has multiple hyperparameters: $\lambda_{U}$ for weighting the self-consistency loss, $\kappa$ for selecting confidence examples, $T$ for sharpening prediction probabilities, $\gamma$ for controlling the Beta distribution, and $M$ for the number of augmentations.
For both CIFAR-10N and CIFAR-100N, we use $\lambda_{U}=1$, $\kappa=0.5$, $T=0.5$, $\gamma=4$, and $M=2$.
For Clothing-1M, we use $\lambda_{U}=0.1$, $\kappa=0.5$, $T=0.5$, $\gamma=0.5$, and $M=2$.
SOP+\,\cite{liu2022robust} also involves several hyperparameters: $\lambda_{C}$ for weighting the self-consistency loss, $\lambda_{B}$ for weighting the class-balance, and learning rates for training its additional variables $u$ and $v$.
For CIFAR-10N, we use $\lambda_{C}=0.9$ and $\lambda_{B}=0.1$, and set the learning rates of $u$ and $v$ to 10 and 100, respectively.
For CIFAR-100N, we use $\lambda_{C}=0.9$ and $\lambda_{B}=0.1$, and set the learning rates of $u$ and $v$ to 1 and 100, respectively.
For WebVision, we use $\lambda_{C}=0.1$ and $\lambda_{B}=0$, and set the learning rates of $u$ and $v$ to 0.1 and 1, respectively.

Besides, the hyperparameters for all data pruning algorithms are also favorably configured following the original papers. 
For Forgetting\,\cite{toneva2018empirical}, we calculate the forgetting event of each example throughout the warm-up training epochs in each dataset.
For GraNd\,\cite{paul2021deep}, we train ten different warm-up classifiers and calculate the per-sample average of the norms of the gradient vectors obtained from the ten classifiers.

\noindent\textbf{Evaluation.}
For CIFAR-10N, CIFAR-100N, and WebVision, we select the subset with the selection ratios $\{$0.2, 0.4, 0.6, 0.8$\}$.
For Clothing-1M and ImageNet-N, we construct the subset with $\{$0.01, 0.05, 0.1, 0.2$\}$ and $\{$0.05, 0.1, 0.2, 0.4$\}$ selection ratios, respectively.
We measure the test accuracy of the Re-labeling models trained from scratch on the selected subset.
Every experiment is run three times, and the average of the last accuracy is reported. 
For CIFAR-10N with the Random noise, we average the test accuracy of the models trained using the three noisy labels.

\def\arraystretch{1.15}
\begin{table*}[t!]
\scriptsize
\centering
\caption{Performance comparison of sample selection baselines and \algnamec{} on CIFAR-10N and CIFAR-100N. The best results are in bold.}
\begin{tabular}[c]
{@{}c|c|cccc|cccc|ccccc@{}}
\toprule
\multirow{3}{*}{\hspace{-0.0cm}\makecell[c]{Relabel\\Models}\hspace{-0.10cm}}&\hspace{-0.20cm}\multirow{3}{*}{\makecell[c]{Selection\\Methods}}& \multicolumn{8}{c|}{{ \underline{CIFAR-10N}}} 
& \multicolumn{4}{c}{{ \underline{CIFAR-100N}}}\\
& & \multicolumn{4}{c|}{{Random}} & \multicolumn{4}{c|}{{Worst}} & \multicolumn{4}{c}{{Noisy}} \\
& & \multicolumn{1}{c}{\hspace{-0.0cm}{0.2}}
& \multicolumn{1}{c}{\hspace{-0.20cm}{0.4}}
& \multicolumn{1}{c}{\hspace{-0.20cm}{0.6}}
& \multicolumn{1}{c|}{\hspace{-0.20cm}{0.8}}
& \multicolumn{1}{c}{\hspace{-0.0cm}{0.2}}
& \multicolumn{1}{c}{\hspace{-0.20cm}{0.4}}
& \multicolumn{1}{c}{\hspace{-0.20cm}{0.6}}
& \multicolumn{1}{c|}{\hspace{-0.20cm}{0.8}}
& \multicolumn{1}{c}{\hspace{-0.0cm}{0.2}}
& \multicolumn{1}{c}{\hspace{-0.20cm}{0.4}}
& \multicolumn{1}{c}{\hspace{-0.20cm}{0.6}}
& \multicolumn{1}{c}{\hspace{-0.20cm}{0.8}}
\\ \midrule
\multirow{10}{*}{\hspace{-0.0cm}{DivMix}\hspace{-0.10cm}} 
& \begin{tabular}[c]{@{}c@{}}{\hspace{-0.20cm}Uniform\hspace{-0.20cm}}\end{tabular}
& \begin{tabular}[c]{@{}c@{}}{ \hspace{-0.20cm}{{87.5}}\scalebox{0.70}{$\pm$0.2}\hspace{-0.20cm} }\end{tabular}
& \begin{tabular}[c]{@{}c@{}}{ \hspace{-0.20cm}{91.9}\scalebox{0.70}{$\pm$0.2}\hspace{-0.20cm} }\end{tabular}
& \begin{tabular}[c]{@{}c@{}}{ \hspace{-0.20cm}{93.7}\scalebox{0.70}{$\pm$0.1}\hspace{-0.20cm} }\end{tabular}
& \begin{tabular}[c]{@{}c@{}}{ \hspace{-0.20cm}{94.8}\scalebox{0.70}{$\pm$0.1}\hspace{-0.10cm} }\end{tabular}
& \begin{tabular}[c]{@{}c@{}}{ \hspace{-0.20cm}{{83.2}}\scalebox{0.70}{$\pm$0.2}\hspace{-0.20cm} }\end{tabular}
& \begin{tabular}[c]{@{}c@{}}{ \hspace{-0.20cm}{{88.5}}\scalebox{0.70}{$\pm$0.1}\hspace{-0.20cm} }\end{tabular}
& \begin{tabular}[c]{@{}c@{}}{ \hspace{-0.20cm}{90.2}\scalebox{0.70}{$\pm$0.0}\hspace{-0.20cm} }\end{tabular}
& \begin{tabular}[c]{@{}c@{}}{ \hspace{-0.20cm}{91.4}\scalebox{0.70}{$\pm$0.0}\hspace{-0.10cm} }\end{tabular}
& \begin{tabular}[c]{@{}c@{}}{ \hspace{-0.20cm}{30.5}\scalebox{0.70}{$\pm$1.0}\hspace{-0.20cm} }\end{tabular}
& \begin{tabular}[c]{@{}c@{}}{ \hspace{-0.20cm}{55.3}\scalebox{0.70}{$\pm$0.5}\hspace{-0.20cm} }\end{tabular}
& \begin{tabular}[c]{@{}c@{}}{ \hspace{-0.20cm}{57.5}\scalebox{0.70}{$\pm$1.9}\hspace{-0.20cm} }\end{tabular}
& \begin{tabular}[c]{@{}c@{}}{ \hspace{-0.20cm}{58.6}\scalebox{0.70}{$\pm$0.9}\hspace{-0.10cm} }\end{tabular}
\\
& \begin{tabular}[c]{@{}c@{}}{\hspace{-0.20cm}SmallL\hspace{-0.20cm}}\end{tabular}
& \begin{tabular}[c]{@{}c@{}}{ \hspace{-0.20cm}{68.1}\scalebox{0.70}{$\pm$4.0}\hspace{-0.20cm} }\end{tabular}
& \begin{tabular}[c]{@{}c@{}}{ \hspace{-0.20cm}{82.4}\scalebox{0.70}{$\pm$0.8}\hspace{-0.20cm} }\end{tabular}
& \begin{tabular}[c]{@{}c@{}}{ \hspace{-0.20cm}{89.0}\scalebox{0.70}{$\pm$0.3}\hspace{-0.20cm} }\end{tabular}
& \begin{tabular}[c]{@{}c@{}}{ \hspace{-0.20cm}{93.1}\scalebox{0.70}{$\pm$0.1}\hspace{-0.10cm} }\end{tabular}
& \begin{tabular}[c]{@{}c@{}}{ \hspace{-0.20cm}{70.3}\scalebox{0.70}{$\pm$0.6}\hspace{-0.20cm} }\end{tabular}
& \begin{tabular}[c]{@{}c@{}}{ \hspace{-0.20cm}{80.3}\scalebox{0.70}{$\pm$0.2}\hspace{-0.20cm} }\end{tabular}
& \begin{tabular}[c]{@{}c@{}}{ \hspace{-0.20cm}{89.1}\scalebox{0.70}{$\pm$0.0}\hspace{-0.20cm} }\end{tabular}
& \begin{tabular}[c]{@{}c@{}}{ \hspace{-0.20cm}{92.1}\scalebox{0.70}{$\pm$0.1}\hspace{-0.10cm} }\end{tabular}
& \begin{tabular}[c]{@{}c@{}}{ \hspace{-0.20cm}{33.3}\scalebox{0.70}{$\pm$3.2}\hspace{-0.20cm} }\end{tabular}
& \begin{tabular}[c]{@{}c@{}}{ \hspace{-0.20cm}{47.4}\scalebox{0.70}{$\pm$1.1}\hspace{-0.20cm} }\end{tabular}
& \begin{tabular}[c]{@{}c@{}}{ \hspace{-0.20cm}{59.4}\scalebox{0.70}{$\pm$0.7}\hspace{-0.20cm} }\end{tabular}
& \begin{tabular}[c]{@{}c@{}}{ \hspace{-0.20cm}{62.0}\scalebox{0.70}{$\pm$1.2}\hspace{-0.10cm} }\end{tabular}
\\
& \begin{tabular}[c]{@{}c@{}}{\hspace{-0.20cm}Margin\hspace{-0.20cm}}\end{tabular}
& \begin{tabular}[c]{@{}c@{}}{ \hspace{-0.20cm}{68.5}\scalebox{0.70}{$\pm$2.9}\hspace{-0.20cm} }\end{tabular}
& \begin{tabular}[c]{@{}c@{}}{ \hspace{-0.20cm}{88.5}\scalebox{0.70}{$\pm$0.3}\hspace{-0.20cm} }\end{tabular}
& \begin{tabular}[c]{@{}c@{}}{ \hspace{-0.20cm}{93.2}\scalebox{0.70}{$\pm$0.2}\hspace{-0.20cm} }\end{tabular}
& \begin{tabular}[c]{@{}c@{}}{ \hspace{-0.20cm}{94.7}\scalebox{0.70}{$\pm$0.1}\hspace{-0.10cm} }\end{tabular}
& \begin{tabular}[c]{@{}c@{}}{ \hspace{-0.20cm}{61.3}\scalebox{0.70}{$\pm$0.8}\hspace{-0.20cm} }\end{tabular}
& \begin{tabular}[c]{@{}c@{}}{ \hspace{-0.20cm}{75.1}\scalebox{0.70}{$\pm$0.7}\hspace{-0.20cm} }\end{tabular}
& \begin{tabular}[c]{@{}c@{}}{ \hspace{-0.20cm}{85.3}\scalebox{0.70}{$\pm$0.2}\hspace{-0.20cm} }\end{tabular}
& \begin{tabular}[c]{@{}c@{}}{ \hspace{-0.20cm}{90.2}\scalebox{0.70}{$\pm$0.1}\hspace{-0.10cm} }\end{tabular}
& \begin{tabular}[c]{@{}c@{}}{ \hspace{-0.20cm}{17.1}\scalebox{0.70}{$\pm$0.8}\hspace{-0.20cm} }\end{tabular}
& \begin{tabular}[c]{@{}c@{}}{ \hspace{-0.20cm}{30.8}\scalebox{0.70}{$\pm$1.0}\hspace{-0.20cm} }\end{tabular}
& \begin{tabular}[c]{@{}c@{}}{ \hspace{-0.20cm}{46.3}\scalebox{0.70}{$\pm$2.4}\hspace{-0.20cm} }\end{tabular}
& \begin{tabular}[c]{@{}c@{}}{ \hspace{-0.20cm}{61.2}\scalebox{0.70}{$\pm$1.3}\hspace{-0.10cm} }\end{tabular}
\\
& \begin{tabular}[c]{@{}c@{}}{\hspace{-0.20cm}kCenter\hspace{-0.20cm}}\end{tabular}
& \begin{tabular}[c]{@{}c@{}}{ \hspace{-0.20cm}{87.4}\scalebox{0.70}{$\pm$0.5}\hspace{-0.20cm} }\end{tabular}
& \begin{tabular}[c]{@{}c@{}}{ \hspace{-0.20cm}{\textbf{93.0}}\scalebox{0.70}{$\pm$0.1}\hspace{-0.20cm} }\end{tabular}
& \begin{tabular}[c]{@{}c@{}}{ \hspace{-0.20cm}{94.4}\scalebox{0.70}{$\pm$0.1}\hspace{-0.20cm} }\end{tabular}
& \begin{tabular}[c]{@{}c@{}}{ \hspace{-0.20cm}{{95.0}}\scalebox{0.70}{$\pm$0.0}\hspace{-0.10cm} }\end{tabular}
& \begin{tabular}[c]{@{}c@{}}{ \hspace{-0.20cm}{82.7}\scalebox{0.70}{$\pm$0.8}\hspace{-0.20cm} }\end{tabular}
& \begin{tabular}[c]{@{}c@{}}{ \hspace{-0.20cm}{88.4}\scalebox{0.70}{$\pm$0.1}\hspace{-0.20cm} }\end{tabular}
& \begin{tabular}[c]{@{}c@{}}{ \hspace{-0.20cm}{90.6}\scalebox{0.70}{$\pm$0.1}\hspace{-0.20cm} }\end{tabular}
& \begin{tabular}[c]{@{}c@{}}{ \hspace{-0.20cm}{92.2}\scalebox{0.70}{$\pm$0.0}\hspace{-0.10cm} }\end{tabular}
& \begin{tabular}[c]{@{}c@{}}{ \hspace{-0.20cm}{38.0}\scalebox{0.70}{$\pm$1.0}\hspace{-0.20cm} }\end{tabular}
& \begin{tabular}[c]{@{}c@{}}{ \hspace{-0.20cm}{50.0}\scalebox{0.70}{$\pm$1.8}\hspace{-0.20cm} }\end{tabular}
& \begin{tabular}[c]{@{}c@{}}{ \hspace{-0.20cm}{59.7}\scalebox{0.70}{$\pm$1.3}\hspace{-0.20cm} }\end{tabular}
& \begin{tabular}[c]{@{}c@{}}{ \hspace{-0.20cm}{63.0}\scalebox{0.70}{$\pm$0.9}\hspace{-0.10cm} }\end{tabular}
\\
& \begin{tabular}[c]{@{}c@{}}{\hspace{-0.20cm}Forget\hspace{-0.20cm}}\end{tabular}
& \begin{tabular}[c]{@{}c@{}}{ \hspace{-0.20cm}{85.2}\scalebox{0.70}{$\pm$0.5}\hspace{-0.20cm} }\end{tabular}
& \begin{tabular}[c]{@{}c@{}}{ \hspace{-0.20cm}{\textbf{93.0}}\scalebox{0.70}{$\pm$0.2}\hspace{-0.20cm} }\end{tabular}
& \begin{tabular}[c]{@{}c@{}}{ \hspace{-0.20cm}{\textbf{94.5}}\scalebox{0.70}{$\pm$0.1}\hspace{-0.20cm} }\end{tabular}
& \begin{tabular}[c]{@{}c@{}}{ \hspace{-0.20cm}{\textbf{95.1}}\scalebox{0.70}{$\pm$0.1}\hspace{-0.10cm} }\end{tabular}
& \begin{tabular}[c]{@{}c@{}}{ \hspace{-0.20cm}{78.3}\scalebox{0.70}{$\pm$0.6}\hspace{-0.20cm} }\end{tabular}
& \begin{tabular}[c]{@{}c@{}}{ \hspace{-0.20cm}{88.3}\scalebox{0.70}{$\pm$0.2}\hspace{-0.20cm} }\end{tabular}
& \begin{tabular}[c]{@{}c@{}}{ \hspace{-0.20cm}{90.4}\scalebox{0.70}{$\pm$0.1}\hspace{-0.20cm} }\end{tabular}
& \begin{tabular}[c]{@{}c@{}}{ \hspace{-0.20cm}{92.0}\scalebox{0.70}{$\pm$0.2}\hspace{-0.10cm} }\end{tabular}
& \begin{tabular}[c]{@{}c@{}}{ \hspace{-0.20cm}{26.4}\scalebox{0.70}{$\pm$1.3}\hspace{-0.20cm} }\end{tabular}
& \begin{tabular}[c]{@{}c@{}}{ \hspace{-0.20cm}{54.3}\scalebox{0.70}{$\pm$0.8}\hspace{-0.20cm} }\end{tabular}
& \begin{tabular}[c]{@{}c@{}}{ \hspace{-0.20cm}{63.1}\scalebox{0.70}{$\pm$1.3}\hspace{-0.20cm} }\end{tabular}
& \begin{tabular}[c]{@{}c@{}}{ \hspace{-0.20cm}{66.6}\scalebox{0.70}{$\pm$1.1}\hspace{-0.10cm} }\end{tabular}
\\
& \begin{tabular}[c]{@{}c@{}}{\hspace{-0.20cm}GraNd\hspace{-0.20cm}}\end{tabular}
& \begin{tabular}[c]{@{}c@{}}{ \hspace{-0.20cm}{21.8}\scalebox{0.70}{$\pm$0.6}\hspace{-0.20cm} }\end{tabular}
& \begin{tabular}[c]{@{}c@{}}{ \hspace{-0.20cm}{60.9}\scalebox{0.70}{$\pm$4.5}\hspace{-0.20cm} }\end{tabular}
& \begin{tabular}[c]{@{}c@{}}{ \hspace{-0.20cm}{92.5}\scalebox{0.70}{$\pm$1.8}\hspace{-0.20cm} }\end{tabular}
& \begin{tabular}[c]{@{}c@{}}{ \hspace{-0.20cm}{94.8}\scalebox{0.70}{$\pm$0.1}\hspace{-0.10cm} }\end{tabular}
& \begin{tabular}[c]{@{}c@{}}{ \hspace{-0.20cm}{18.5}\scalebox{0.70}{$\pm$1.7}\hspace{-0.20cm} }\end{tabular}
& \begin{tabular}[c]{@{}c@{}}{ \hspace{-0.20cm}{25.5}\scalebox{0.70}{$\pm$0.9}\hspace{-0.20cm} }\end{tabular}
& \begin{tabular}[c]{@{}c@{}}{ \hspace{-0.20cm}{49.3}\scalebox{0.70}{$\pm$0.9}\hspace{-0.20cm} }\end{tabular}
& \begin{tabular}[c]{@{}c@{}}{ \hspace{-0.20cm}{88.0}\scalebox{0.70}{$\pm$0.5}\hspace{-0.10cm} }\end{tabular}
& \begin{tabular}[c]{@{}c@{}}{ \hspace{-0.20cm}{15.5}\scalebox{0.70}{$\pm$1.2}\hspace{-0.20cm} }\end{tabular}
& \begin{tabular}[c]{@{}c@{}}{ \hspace{-0.20cm}{26.0}\scalebox{0.70}{$\pm$1.9}\hspace{-0.20cm} }\end{tabular}
& \begin{tabular}[c]{@{}c@{}}{ \hspace{-0.20cm}{44.7}\scalebox{0.70}{$\pm$1.5}\hspace{-0.20cm} }\end{tabular}
& \begin{tabular}[c]{@{}c@{}}{ \hspace{-0.20cm}{60.4}\scalebox{0.70}{$\pm$1.6}\hspace{-0.10cm} }\end{tabular}
\\
& \begin{tabular}[c]{@{}c@{}}{\hspace{-0.10cm}SSP\hspace{-0.20cm}}\end{tabular}
& \begin{tabular}[c]{@{}c@{}}{ \hspace{-0.20cm}{85.8}\scalebox{0.70}{$\pm$1.8}\hspace{-0.20cm} }\end{tabular}
& \begin{tabular}[c]{@{}c@{}}{ \hspace{-0.20cm}{92.2}\scalebox{0.70}{$\pm$1.5}\hspace{-0.20cm} }\end{tabular}
& \begin{tabular}[c]{@{}c@{}}{ \hspace{-0.20cm}{93.0}\scalebox{0.70}{$\pm$1.1}\hspace{-0.20cm} }\end{tabular}
& \begin{tabular}[c]{@{}c@{}}{ \hspace{-0.20cm}{94.5}\scalebox{0.70}{$\pm$0.2}\hspace{-0.10cm} }\end{tabular}
& \begin{tabular}[c]{@{}c@{}}{ \hspace{-0.20cm}{81.4}\scalebox{0.70}{$\pm$2.5}\hspace{-0.20cm} }\end{tabular}
& \begin{tabular}[c]{@{}c@{}}{ \hspace{-0.20cm}{86.5}\scalebox{0.70}{$\pm$1.9}\hspace{-0.20cm} }\end{tabular}
& \begin{tabular}[c]{@{}c@{}}{ \hspace{-0.20cm}{89.6}\scalebox{0.70}{$\pm$1.2}\hspace{-0.20cm} }\end{tabular}
& \begin{tabular}[c]{@{}c@{}}{ \hspace{-0.20cm}{91.9}\scalebox{0.70}{$\pm$0.4}\hspace{-0.10cm} }\end{tabular}
& \begin{tabular}[c]{@{}c@{}}{ \hspace{-0.20cm}{30.1}\scalebox{0.70}{$\pm$2.2}\hspace{-0.20cm} }\end{tabular}
& \begin{tabular}[c]{@{}c@{}}{ \hspace{-0.20cm}{52.4}\scalebox{0.70}{$\pm$1.3}\hspace{-0.20cm} }\end{tabular}
& \begin{tabular}[c]{@{}c@{}}{ \hspace{-0.20cm}{58.7}\scalebox{0.70}{$\pm$0.5}\hspace{-0.20cm} }\end{tabular}
& \begin{tabular}[c]{@{}c@{}}{ \hspace{-0.20cm}{63.4}\scalebox{0.70}{$\pm$0.5}\hspace{-0.10cm} }\end{tabular}
\\
& \begin{tabular}[c]{@{}c@{}}{\hspace{-0.20cm}Moderate\hspace{-0.20cm}}\end{tabular}
& \begin{tabular}[c]{@{}c@{}}{ \hspace{-0.20cm}{86.4}\scalebox{0.70}{$\pm$0.8}\hspace{-0.20cm} }\end{tabular}
& \begin{tabular}[c]{@{}c@{}}{ \hspace{-0.20cm}{91.4}\scalebox{0.70}{$\pm$0.3}\hspace{-0.20cm} }\end{tabular}
& \begin{tabular}[c]{@{}c@{}}{ \hspace{-0.20cm}{93.8}\scalebox{0.70}{$\pm$0.5}\hspace{-0.20cm} }\end{tabular}
& \begin{tabular}[c]{@{}c@{}}{ \hspace{-0.20cm}{94.8}\scalebox{0.70}{$\pm$0.2}\hspace{-0.10cm} }\end{tabular}
& \begin{tabular}[c]{@{}c@{}}{ \hspace{-0.20cm}{81.4}\scalebox{0.70}{$\pm$1.2}\hspace{-0.20cm} }\end{tabular}
& \begin{tabular}[c]{@{}c@{}}{ \hspace{-0.20cm}{86.5}\scalebox{0.70}{$\pm$0.6}\hspace{-0.20cm} }\end{tabular}
& \begin{tabular}[c]{@{}c@{}}{ \hspace{-0.20cm}{90.0}\scalebox{0.70}{$\pm$0.6}\hspace{-0.20cm} }\end{tabular}
& \begin{tabular}[c]{@{}c@{}}{ \hspace{-0.20cm}{91.6}\scalebox{0.70}{$\pm$0.2}\hspace{-0.10cm} }\end{tabular}
& \begin{tabular}[c]{@{}c@{}}{ \hspace{-0.20cm}{34.2}\scalebox{0.70}{$\pm$1.4}\hspace{-0.20cm} }\end{tabular}
& \begin{tabular}[c]{@{}c@{}}{ \hspace{-0.20cm}{54.5}\scalebox{0.70}{$\pm$1.3}\hspace{-0.20cm} }\end{tabular}
& \begin{tabular}[c]{@{}c@{}}{ \hspace{-0.20cm}{56.1}\scalebox{0.70}{$\pm$0.5}\hspace{-0.20cm} }\end{tabular}
& \begin{tabular}[c]{@{}c@{}}{ \hspace{-0.20cm}{59.9}\scalebox{0.70}{$\pm$0.6}\hspace{-0.10cm} }\end{tabular}
\\ \cline{2-14} \addlinespace[0.15ex]
& \begin{tabular}[c]{@{}c@{}}{\hspace{-0.20cm}{\tblalgnamec{}}\hspace{-0.20cm}}\end{tabular}
& \begin{tabular}[c]{@{}c@{}}{ \hspace{-0.20cm}{{87.6}}\scalebox{0.70}{$\pm$0.3}\hspace{-0.20cm} }\end{tabular}
& \begin{tabular}[c]{@{}c@{}}{ \hspace{-0.20cm}{92.4}\scalebox{0.70}{$\pm$0.4}\hspace{-0.20cm} }\end{tabular}
& \begin{tabular}[c]{@{}c@{}}{ \hspace{-0.20cm}{\textbf{94.5}}\scalebox{0.70}{$\pm$0.2}\hspace{-0.20cm} }\end{tabular}
& \begin{tabular}[c]{@{}c@{}}{ \hspace{-0.20cm}{\textbf{95.1}}\scalebox{0.70}{$\pm$0.1}\hspace{-0.10cm} }\end{tabular}
& \begin{tabular}[c]{@{}c@{}}{ \hspace{-0.20cm}{\textbf{83.7}}\scalebox{0.70}{$\pm$0.5}\hspace{-0.20cm} }\end{tabular}
& \begin{tabular}[c]{@{}c@{}}{ \hspace{-0.20cm}{\textbf{88.8}}\scalebox{0.70}{$\pm$0.3}\hspace{-0.20cm} }\end{tabular}
& \begin{tabular}[c]{@{}c@{}}{ \hspace{-0.20cm}{{90.2}}\scalebox{0.70}{$\pm$0.3}\hspace{-0.20cm} }\end{tabular}
& \begin{tabular}[c]{@{}c@{}}{ \hspace{-0.20cm}{92.0}\scalebox{0.70}{$\pm$0.3}\hspace{-0.10cm} }\end{tabular}
& \begin{tabular}[c]{@{}c@{}}{ \hspace{-0.20cm}{37.2}\scalebox{0.70}{$\pm$1.0}\hspace{-0.20cm} }\end{tabular}
& \begin{tabular}[c]{@{}c@{}}{ \hspace{-0.20cm}{{55.3}}\scalebox{0.70}{$\pm$0.7}\hspace{-0.20cm} }\end{tabular}
& \begin{tabular}[c]{@{}c@{}}{ \hspace{-0.20cm}{61.2}\scalebox{0.70}{$\pm$0.5}\hspace{-0.20cm} }\end{tabular}
& \begin{tabular}[c]{@{}c@{}}{ \hspace{-0.20cm}{65.5}\scalebox{0.70}{$\pm$0.8}\hspace{-0.10cm} }\end{tabular}
\\
& \begin{tabular}[c]{@{}c@{}}{\hspace{-0.20cm}{$\text{\tblalgnamec{}}_B$}\hspace{-0.20cm}}\end{tabular}
& \begin{tabular}[c]{@{}c@{}}{ \hspace{-0.20cm}{\textbf{88.1}}\scalebox{0.70}{$\pm$0.3}\hspace{-0.20cm} }\end{tabular}
& \begin{tabular}[c]{@{}c@{}}{ \hspace{-0.20cm}{\textbf{93.0}}\scalebox{0.70}{$\pm$0.2}\hspace{-0.20cm} }\end{tabular}
& \begin{tabular}[c]{@{}c@{}}{ \hspace{-0.20cm}{\textbf{94.5}}\scalebox{0.70}{$\pm$0.2}\hspace{-0.20cm} }\end{tabular}
& \begin{tabular}[c]{@{}c@{}}{ \hspace{-0.20cm}\textbf{95.1}\scalebox{0.70}{$\pm$0.1}\hspace{-0.10cm} }\end{tabular}
& \begin{tabular}[c]{@{}c@{}}{ \hspace{-0.20cm}{\textbf{83.7}}\scalebox{0.70}{$\pm$0.4}\hspace{-0.20cm} }\end{tabular}
& \begin{tabular}[c]{@{}c@{}}{ \hspace{-0.20cm}{{88.6}}\scalebox{0.70}{$\pm$0.4}\hspace{-0.20cm} }\end{tabular}
& \begin{tabular}[c]{@{}c@{}}{ \hspace{-0.20cm}{\textbf{90.8}}\scalebox{0.70}{$\pm$0.2}\hspace{-0.20cm} }\end{tabular}
& \begin{tabular}[c]{@{}c@{}}{ \hspace{-0.20cm}{\textbf{92.4}}\scalebox{0.70}{$\pm$0.2}\hspace{-0.10cm} }\end{tabular}
& \begin{tabular}[c]{@{}c@{}}{ \hspace{-0.20cm}{\textbf{39.4}}\scalebox{0.70}{$\pm$0.8}\hspace{-0.20cm} }\end{tabular}
& \begin{tabular}[c]{@{}c@{}}{ \hspace{-0.20cm}{\textbf{56.3}}\scalebox{0.70}{$\pm$0.5}\hspace{-0.20cm} }\end{tabular}
& \begin{tabular}[c]{@{}c@{}}{ \hspace{-0.20cm}{\textbf{63.5}}\scalebox{0.70}{$\pm$0.3}\hspace{-0.20cm} }\end{tabular}
& \begin{tabular}[c]{@{}c@{}}{ \hspace{-0.20cm}{\textbf{67.4}}\scalebox{0.70}{$\pm$0.7}\hspace{-0.10cm} }\end{tabular}
\\
\midrule
\multirow{10}{*}{\hspace{-0.0cm}{SOP+}\hspace{-0.30cm}}                             
& \begin{tabular}[c]{@{}c@{}}{\hspace{-0.20cm}Uniform\hspace{-0.20cm}}\end{tabular}
& \begin{tabular}[c]{@{}c@{}}{ \hspace{-0.20cm}{ {87.5}}\scalebox{0.70}{$\pm$0.3}\hspace{-0.20cm} }\end{tabular}
& \begin{tabular}[c]{@{}c@{}}{ \hspace{-0.20cm}{91.5}\scalebox{0.70}{$\pm$0.1}\hspace{-0.20cm} }\end{tabular}
& \begin{tabular}[c]{@{}c@{}}{ \hspace{-0.20cm}{93.4}\scalebox{0.70}{$\pm$0.0}\hspace{-0.20cm} }\end{tabular}
& \begin{tabular}[c]{@{}c@{}}{ \hspace{-0.20cm}{94.8}\scalebox{0.70}{$\pm$0.2}\hspace{-0.10cm} }\end{tabular}
& \begin{tabular}[c]{@{}c@{}}{ \hspace{-0.20cm}{ {81.9}}\scalebox{0.70}{$\pm$0.1}\hspace{-0.20cm} }\end{tabular}
& \begin{tabular}[c]{@{}c@{}}{ \hspace{-0.20cm}{87.5}\scalebox{0.70}{$\pm$0.1}\hspace{-0.20cm} }\end{tabular}
& \begin{tabular}[c]{@{}c@{}}{ \hspace{-0.20cm}{90.8}\scalebox{0.70}{$\pm$0.1}\hspace{-0.20cm} }\end{tabular}
& \begin{tabular}[c]{@{}c@{}}{ \hspace{-0.20cm}{91.8}\scalebox{0.70}{$\pm$0.1}\hspace{-0.10cm} }\end{tabular}
& \begin{tabular}[c]{@{}c@{}}{ \hspace{-0.20cm}{{46.5}}\scalebox{0.70}{$\pm$0.0}\hspace{-0.20cm} }\end{tabular}
& \begin{tabular}[c]{@{}c@{}}{ \hspace{-0.20cm}{55.7}\scalebox{0.70}{$\pm$0.2}\hspace{-0.20cm} }\end{tabular}
& \begin{tabular}[c]{@{}c@{}}{ \hspace{-0.20cm}{60.8}\scalebox{0.70}{$\pm$0.3}\hspace{-0.20cm} }\end{tabular}
& \begin{tabular}[c]{@{}c@{}}{ \hspace{-0.20cm}{64.4}\scalebox{0.70}{$\pm$0.2}\hspace{-0.10cm} }\end{tabular}
\\
& \begin{tabular}[c]{@{}c@{}}{\hspace{-0.20cm}SmallL\hspace{-0.20cm}}\end{tabular}
& \begin{tabular}[c]{@{}c@{}}{ \hspace{-0.20cm}{77.6}\scalebox{0.70}{$\pm$2.5}\hspace{-0.20cm} }\end{tabular}
& \begin{tabular}[c]{@{}c@{}}{ \hspace{-0.20cm}{86.2}\scalebox{0.70}{$\pm$0.1}\hspace{-0.20cm} }\end{tabular}
& \begin{tabular}[c]{@{}c@{}}{ \hspace{-0.20cm}{90.7}\scalebox{0.70}{$\pm$0.6}\hspace{-0.20cm} }\end{tabular}
& \begin{tabular}[c]{@{}c@{}}{ \hspace{-0.20cm}{94.3}\scalebox{0.70}{$\pm$0.2}\hspace{-0.10cm} }\end{tabular}
& \begin{tabular}[c]{@{}c@{}}{ \hspace{-0.20cm}{78.8}\scalebox{0.70}{$\pm$0.2}\hspace{-0.20cm} }\end{tabular}
& \begin{tabular}[c]{@{}c@{}}{ \hspace{-0.20cm}{84.1}\scalebox{0.70}{$\pm$0.1}\hspace{-0.20cm} }\end{tabular}
& \begin{tabular}[c]{@{}c@{}}{ \hspace{-0.20cm}{89.3}\scalebox{0.70}{$\pm$0.1}\hspace{-0.20cm} }\end{tabular}
& \begin{tabular}[c]{@{}c@{}}{ \hspace{-0.20cm}{92.3}\scalebox{0.70}{$\pm$0.2}\hspace{-0.10cm} }\end{tabular}
& \begin{tabular}[c]{@{}c@{}}{ \hspace{-0.20cm}{{48.5}}\scalebox{0.70}{$\pm$0.8}\hspace{-0.20cm} }\end{tabular}
& \begin{tabular}[c]{@{}c@{}}{ \hspace{-0.20cm}{{59.8}}\scalebox{0.70}{$\pm$0.4}\hspace{-0.20cm} }\end{tabular}
& \begin{tabular}[c]{@{}c@{}}{ \hspace{-0.20cm}{{63.9}}\scalebox{0.70}{$\pm$0.2}\hspace{-0.20cm} }\end{tabular}
& \begin{tabular}[c]{@{}c@{}}{ \hspace{-0.20cm}{{66.1}}\scalebox{0.70}{$\pm$0.6}\hspace{-0.10cm} }\end{tabular}
\\
& \begin{tabular}[c]{@{}c@{}}{\hspace{-0.20cm}Margin\hspace{-0.20cm}}\end{tabular}
& \begin{tabular}[c]{@{}c@{}}{ \hspace{-0.20cm}{52.1}\scalebox{0.70}{$\pm$5.0}\hspace{-0.20cm} }\end{tabular}
& \begin{tabular}[c]{@{}c@{}}{ \hspace{-0.20cm}{79.6}\scalebox{0.70}{$\pm$8.6}\hspace{-0.20cm} }\end{tabular}
& \begin{tabular}[c]{@{}c@{}}{ \hspace{-0.20cm}{92.6}\scalebox{0.70}{$\pm$3.9}\hspace{-0.20cm} }\end{tabular}
& \begin{tabular}[c]{@{}c@{}}{ \hspace{-0.20cm}{95.1}\scalebox{0.70}{$\pm$1.3}\hspace{-0.10cm} }\end{tabular}
& \begin{tabular}[c]{@{}c@{}}{ \hspace{-0.20cm}{45.7}\scalebox{0.70}{$\pm$1.1}\hspace{-0.20cm} }\end{tabular}
& \begin{tabular}[c]{@{}c@{}}{ \hspace{-0.20cm}{61.8}\scalebox{0.70}{$\pm$0.7}\hspace{-0.20cm} }\end{tabular}
& \begin{tabular}[c]{@{}c@{}}{ \hspace{-0.20cm}{84.6}\scalebox{0.70}{$\pm$0.3}\hspace{-0.20cm} }\end{tabular}
& \begin{tabular}[c]{@{}c@{}}{ \hspace{-0.20cm}{{92.5}}\scalebox{0.70}{$\pm$0.0}\hspace{-0.10cm} }\end{tabular}
& \begin{tabular}[c]{@{}c@{}}{ \hspace{-0.20cm}{20.0}\scalebox{0.70}{$\pm$1.2}\hspace{-0.20cm} }\end{tabular}
& \begin{tabular}[c]{@{}c@{}}{ \hspace{-0.20cm}{34.4}\scalebox{0.70}{$\pm$0.3}\hspace{-0.20cm} }\end{tabular}
& \begin{tabular}[c]{@{}c@{}}{ \hspace{-0.20cm}{50.4}\scalebox{0.70}{$\pm$0.6}\hspace{-0.20cm} }\end{tabular}
& \begin{tabular}[c]{@{}c@{}}{ \hspace{-0.20cm}{63.3}\scalebox{0.70}{$\pm$0.1}\hspace{-0.10cm} }\end{tabular}
\\
& \begin{tabular}[c]{@{}c@{}}{\hspace{-0.20cm}kCenter\hspace{-0.20cm}}\end{tabular}
& \begin{tabular}[c]{@{}c@{}}{ \hspace{-0.20cm}{86.3}\scalebox{0.70}{$\pm$0.4}\hspace{-0.20cm} }\end{tabular}
& \begin{tabular}[c]{@{}c@{}}{ \hspace{-0.20cm}{92.2}\scalebox{0.70}{$\pm$0.3}\hspace{-0.20cm} }\end{tabular}
& \begin{tabular}[c]{@{}c@{}}{ \hspace{-0.20cm}{94.1}\scalebox{0.70}{$\pm$0.2}\hspace{-0.20cm} }\end{tabular}
& \begin{tabular}[c]{@{}c@{}}{ \hspace{-0.20cm}{\textbf{95.3}}\scalebox{0.70}{$\pm$0.1}\hspace{-0.10cm} }\end{tabular}
& \begin{tabular}[c]{@{}c@{}}{ \hspace{-0.20cm}{{81.9}}\scalebox{0.70}{$\pm$0.0}\hspace{-0.20cm} }\end{tabular}
& \begin{tabular}[c]{@{}c@{}}{ \hspace{-0.20cm}{{88.0}}\scalebox{0.70}{$\pm$0.0}\hspace{-0.20cm} }\end{tabular}
& \begin{tabular}[c]{@{}c@{}}{ \hspace{-0.20cm}{\textbf{91.3}}\scalebox{0.70}{$\pm$0.1}\hspace{-0.20cm} }\end{tabular}
& \begin{tabular}[c]{@{}c@{}}{ \hspace{-0.20cm}{92.3}\scalebox{0.70}{$\pm$0.0}\hspace{-0.10cm} }\end{tabular}
& \begin{tabular}[c]{@{}c@{}}{ \hspace{-0.20cm}{44.8}\scalebox{0.70}{$\pm$0.6}\hspace{-0.20cm} }\end{tabular}
& \begin{tabular}[c]{@{}c@{}}{ \hspace{-0.20cm}{55.9}\scalebox{0.70}{$\pm$0.4}\hspace{-0.20cm} }\end{tabular}
& \begin{tabular}[c]{@{}c@{}}{ \hspace{-0.20cm}{61.6}\scalebox{0.70}{$\pm$0.3}\hspace{-0.20cm} }\end{tabular}
& \begin{tabular}[c]{@{}c@{}}{ \hspace{-0.20cm}{65.2}\scalebox{0.70}{$\pm$0.6}\hspace{-0.10cm} }\end{tabular}
\\
& \begin{tabular}[c]{@{}c@{}}{\hspace{-0.20cm}Forget\hspace{-0.20cm}}\end{tabular}
& \begin{tabular}[c]{@{}c@{}}{ \hspace{-0.20cm}{82.4}\scalebox{0.70}{$\pm$1.0}\hspace{-0.20cm} }\end{tabular}
& \begin{tabular}[c]{@{}c@{}}{ \hspace{-0.20cm}{{93.0}}\scalebox{0.70}{$\pm$0.2}\hspace{-0.20cm} }\end{tabular}
& \begin{tabular}[c]{@{}c@{}}{ \hspace{-0.20cm}{{94.2}}\scalebox{0.70}{$\pm$0.3}\hspace{-0.20cm} }\end{tabular}
& \begin{tabular}[c]{@{}c@{}}{ \hspace{-0.20cm}{95.0}\scalebox{0.70}{$\pm$0.1}\hspace{-0.10cm} }\end{tabular}
& \begin{tabular}[c]{@{}c@{}}{ \hspace{-0.20cm}{71.1}\scalebox{0.70}{$\pm$0.4}\hspace{-0.20cm} }\end{tabular}
& \begin{tabular}[c]{@{}c@{}}{ \hspace{-0.20cm}{87.7}\scalebox{0.70}{$\pm$0.1}\hspace{-0.20cm} }\end{tabular}
& \begin{tabular}[c]{@{}c@{}}{ \hspace{-0.20cm}{90.6}\scalebox{0.70}{$\pm$0.3}\hspace{-0.20cm} }\end{tabular}
& \begin{tabular}[c]{@{}c@{}}{ \hspace{-0.20cm}{92.2}\scalebox{0.70}{$\pm$0.0}\hspace{-0.10cm} }\end{tabular}
& \begin{tabular}[c]{@{}c@{}}{ \hspace{-0.20cm}{38.0}\scalebox{0.70}{$\pm$0.5}\hspace{-0.20cm} }\end{tabular}
& \begin{tabular}[c]{@{}c@{}}{ \hspace{-0.20cm}{55.3}\scalebox{0.70}{$\pm$0.2}\hspace{-0.20cm} }\end{tabular}
& \begin{tabular}[c]{@{}c@{}}{ \hspace{-0.20cm}{63.2}\scalebox{0.70}{$\pm$0.1}\hspace{-0.20cm} }\end{tabular}
& \begin{tabular}[c]{@{}c@{}}{ \hspace{-0.20cm}{65.8}\scalebox{0.70}{$\pm$0.4}\hspace{-0.10cm} }\end{tabular}
\\
& \begin{tabular}[c]{@{}c@{}}{\hspace{-0.20cm}GraNd\hspace{-0.20cm}}\end{tabular}
& \begin{tabular}[c]{@{}c@{}}{ \hspace{-0.20cm}{24.2}\scalebox{0.70}{$\pm$5.5}\hspace{-0.20cm} }\end{tabular}
& \begin{tabular}[c]{@{}c@{}}{ \hspace{-0.20cm}{51.6}\scalebox{0.70}{$\pm$3.2}\hspace{-0.20cm} }\end{tabular}
& \begin{tabular}[c]{@{}c@{}}{ \hspace{-0.20cm}{85.9}\scalebox{0.70}{$\pm$1.2}\hspace{-0.20cm} }\end{tabular}
& \begin{tabular}[c]{@{}c@{}}{ \hspace{-0.20cm}{94.9}\scalebox{0.70}{$\pm$0.2}\hspace{-0.10cm} }\end{tabular}
& \begin{tabular}[c]{@{}c@{}}{ \hspace{-0.20cm}{15.4}\scalebox{0.70}{$\pm$1.6}\hspace{-0.20cm} }\end{tabular}
& \begin{tabular}[c]{@{}c@{}}{ \hspace{-0.20cm}{25.7}\scalebox{0.70}{$\pm$0.8}\hspace{-0.20cm} }\end{tabular}
& \begin{tabular}[c]{@{}c@{}}{ \hspace{-0.20cm}{51.0}\scalebox{0.70}{$\pm$0.5}\hspace{-0.20cm} }\end{tabular}
& \begin{tabular}[c]{@{}c@{}}{ \hspace{-0.20cm}{86.8}\scalebox{0.70}{$\pm$0.5}\hspace{-0.10cm} }\end{tabular}
& \begin{tabular}[c]{@{}c@{}}{ \hspace{-0.20cm}{11.0}\scalebox{0.70}{$\pm$0.1}\hspace{-0.20cm} }\end{tabular}
& \begin{tabular}[c]{@{}c@{}}{ \hspace{-0.20cm}{19.0}\scalebox{0.70}{$\pm$0.6}\hspace{-0.20cm} }\end{tabular}
& \begin{tabular}[c]{@{}c@{}}{ \hspace{-0.20cm}{38.7}\scalebox{0.70}{$\pm$0.5}\hspace{-0.20cm} }\end{tabular}
& \begin{tabular}[c]{@{}c@{}}{ \hspace{-0.20cm}{62.1}\scalebox{0.70}{$\pm$0.5}\hspace{-0.10cm} }\end{tabular}
\\
& \begin{tabular}[c]{@{}c@{}}{\hspace{-0.20cm}SSP\hspace{-0.20cm}}\end{tabular}
& \begin{tabular}[c]{@{}c@{}}{ \hspace{-0.20cm}{80.5}\scalebox{0.70}{$\pm$2.6}\hspace{-0.20cm} }\end{tabular}
& \begin{tabular}[c]{@{}c@{}}{ \hspace{-0.20cm}{91.7}\scalebox{0.70}{$\pm$1.5}\hspace{-0.20cm} }\end{tabular}
& \begin{tabular}[c]{@{}c@{}}{ \hspace{-0.20cm}{93.8}\scalebox{0.70}{$\pm$1.0}\hspace{-0.20cm} }\end{tabular}
& \begin{tabular}[c]{@{}c@{}}{ \hspace{-0.20cm}{95.0}\scalebox{0.70}{$\pm$0.2}\hspace{-0.10cm} }\end{tabular}
& \begin{tabular}[c]{@{}c@{}}{ \hspace{-0.20cm}{70.8}\scalebox{0.70}{$\pm$2.7}\hspace{-0.20cm} }\end{tabular}
& \begin{tabular}[c]{@{}c@{}}{ \hspace{-0.20cm}{86.6}\scalebox{0.70}{$\pm$1.9}\hspace{-0.20cm} }\end{tabular}
& \begin{tabular}[c]{@{}c@{}}{ \hspace{-0.20cm}{89.2}\scalebox{0.70}{$\pm$0.9}\hspace{-0.20cm} }\end{tabular}
& \begin{tabular}[c]{@{}c@{}}{ \hspace{-0.20cm}{92.3}\scalebox{0.70}{$\pm$0.4}\hspace{-0.10cm} }\end{tabular}
& \begin{tabular}[c]{@{}c@{}}{ \hspace{-0.20cm}{39.2}\scalebox{0.70}{$\pm$2.2}\hspace{-0.20cm} }\end{tabular}
& \begin{tabular}[c]{@{}c@{}}{ \hspace{-0.20cm}{54.9}\scalebox{0.70}{$\pm$1.5}\hspace{-0.20cm} }\end{tabular}
& \begin{tabular}[c]{@{}c@{}}{ \hspace{-0.20cm}{62.7}\scalebox{0.70}{$\pm$0.7}\hspace{-0.20cm} }\end{tabular}
& \begin{tabular}[c]{@{}c@{}}{ \hspace{-0.20cm}{65.0}\scalebox{0.70}{$\pm$0.3}\hspace{-0.10cm} }\end{tabular}
\\
& \begin{tabular}[c]{@{}c@{}}{\hspace{-0.20cm}Moderate\hspace{-0.20cm}}\end{tabular}
& \begin{tabular}[c]{@{}c@{}}{ \hspace{-0.20cm}{87.8}\scalebox{0.70}{$\pm$1.0}\hspace{-0.20cm} }\end{tabular}
& \begin{tabular}[c]{@{}c@{}}{ \hspace{-0.20cm}{92.8}\scalebox{0.70}{$\pm$0.5}\hspace{-0.20cm} }\end{tabular}
& \begin{tabular}[c]{@{}c@{}}{ \hspace{-0.20cm}{94.0}\scalebox{0.70}{$\pm$0.3}\hspace{-0.20cm} }\end{tabular}
& \begin{tabular}[c]{@{}c@{}}{ \hspace{-0.20cm}{94.9}\scalebox{0.70}{$\pm$0.2}\hspace{-0.10cm} }\end{tabular}
& \begin{tabular}[c]{@{}c@{}}{ \hspace{-0.20cm}{75.2}\scalebox{0.70}{$\pm$1.5}\hspace{-0.20cm} }\end{tabular}
& \begin{tabular}[c]{@{}c@{}}{ \hspace{-0.20cm}{81.9}\scalebox{0.70}{$\pm$1.2}\hspace{-0.20cm} }\end{tabular}
& \begin{tabular}[c]{@{}c@{}}{ \hspace{-0.20cm}{87.7}\scalebox{0.70}{$\pm$0.7}\hspace{-0.20cm} }\end{tabular}
& \begin{tabular}[c]{@{}c@{}}{ \hspace{-0.20cm}{91.8}\scalebox{0.70}{$\pm$0.3}\hspace{-0.10cm} }\end{tabular}
& \begin{tabular}[c]{@{}c@{}}{ \hspace{-0.20cm}{46.4}\scalebox{0.70}{$\pm$1.8}\hspace{-0.20cm} }\end{tabular}
& \begin{tabular}[c]{@{}c@{}}{ \hspace{-0.20cm}{54.6}\scalebox{0.70}{$\pm$1.7}\hspace{-0.20cm} }\end{tabular}
& \begin{tabular}[c]{@{}c@{}}{ \hspace{-0.20cm}{60.2}\scalebox{0.70}{$\pm$0.4}\hspace{-0.20cm} }\end{tabular}
& \begin{tabular}[c]{@{}c@{}}{ \hspace{-0.20cm}{64.6}\scalebox{0.70}{$\pm$0.4}\hspace{-0.10cm} }\end{tabular}
\\ \cline{2-14} \addlinespace[0.15ex]
& \begin{tabular}[c]{@{}c@{}}{\hspace{-0.20cm}{\tblalgnamec{}}\hspace{-0.20cm}}\end{tabular}
& \begin{tabular}[c]{@{}c@{}}{ \hspace{-0.20cm}{{87.8}}\scalebox{0.70}{$\pm$1.2}\hspace{-0.20cm} }\end{tabular}
& \begin{tabular}[c]{@{}c@{}}{ \hspace{-0.20cm}{{92.7}}\scalebox{0.70}{$\pm$0.3}\hspace{-0.20cm} }\end{tabular}
& \begin{tabular}[c]{@{}c@{}}{ \hspace{-0.20cm}{\textbf{94.4}}\scalebox{0.70}{$\pm$0.2}\hspace{-0.20cm} }\end{tabular}
& \begin{tabular}[c]{@{}c@{}}{ \hspace{-0.20cm}{95.1}\scalebox{0.70}{$\pm$0.1}\hspace{-0.10cm} }\end{tabular}
& \begin{tabular}[c]{@{}c@{}}{ \hspace{-0.20cm}{{82.7}}\scalebox{0.70}{$\pm$0.5}\hspace{-0.20cm} }\end{tabular}
& \begin{tabular}[c]{@{}c@{}}{ \hspace{-0.20cm}{{88.1}}\scalebox{0.70}{$\pm$0.4}\hspace{-0.20cm} }\end{tabular}
& \begin{tabular}[c]{@{}c@{}}{ \hspace{-0.20cm}{\textbf{91.3}}\scalebox{0.70}{$\pm$0.3}\hspace{-0.20cm} }\end{tabular}
& \begin{tabular}[c]{@{}c@{}}{ \hspace{-0.20cm}{{92.5}}\scalebox{0.70}{$\pm$0.2}\hspace{-0.10cm} }\end{tabular}
& \begin{tabular}[c]{@{}c@{}}{ \hspace{-0.20cm}{{50.2}}\scalebox{0.70}{$\pm$0.2}\hspace{-0.20cm} }\end{tabular}
& \begin{tabular}[c]{@{}c@{}}{ \hspace{-0.20cm}{59.1}\scalebox{0.70}{$\pm$0.5}\hspace{-0.20cm} }\end{tabular}
& \begin{tabular}[c]{@{}c@{}}{ \hspace{-0.20cm}{{63.9}}\scalebox{0.70}{$\pm$0.3}\hspace{-0.20cm} }\end{tabular}
& \begin{tabular}[c]{@{}c@{}}{ \hspace{-0.20cm}{65.7}\scalebox{0.70}{$\pm$0.5}\hspace{-0.10cm} }\end{tabular}
\\
& \begin{tabular}[c]{@{}c@{}}{\hspace{-0.20cm}{$\text{\tblalgnamec{}}_B$}\hspace{-0.60cm}}\end{tabular}
& \begin{tabular}[c]{@{}c@{}}{ \hspace{-0.20cm}{\textbf{88.5}}\scalebox{0.70}{$\pm$0.3}\hspace{-0.20cm} }\end{tabular}
& \begin{tabular}[c]{@{}c@{}}{ \hspace{-0.20cm}{\textbf{93.1}}\scalebox{0.70}{$\pm$0.2}\hspace{-0.20cm} }\end{tabular}
& \begin{tabular}[c]{@{}c@{}}{ \hspace{-0.20cm}{\textbf{94.4}}\scalebox{0.70}{$\pm$0.1}\hspace{-0.20cm} }\end{tabular}
& \begin{tabular}[c]{@{}c@{}}{ \hspace{-0.20cm}{\textbf{95.3}}\scalebox{0.70}{$\pm$0.1}\hspace{-0.10cm} }\end{tabular}
& \begin{tabular}[c]{@{}c@{}}{ \hspace{-0.20cm}{\textbf{84.9}}\scalebox{0.70}{$\pm$0.6}\hspace{-0.20cm} }\end{tabular}
& \begin{tabular}[c]{@{}c@{}}{ \hspace{-0.20cm}{\textbf{89.2}}\scalebox{0.70}{$\pm$0.6}\hspace{-0.20cm} }\end{tabular}
& \begin{tabular}[c]{@{}c@{}}{ \hspace{-0.20cm}{\textbf{91.3}}\scalebox{0.70}{$\pm$0.3}\hspace{-0.20cm} }\end{tabular}
& \begin{tabular}[c]{@{}c@{}}{ \hspace{-0.20cm}{\textbf{92.9}}\scalebox{0.70}{$\pm$0.1}\hspace{-0.10cm} }\end{tabular}
& \begin{tabular}[c]{@{}c@{}}{ \hspace{-0.20cm}{\textbf{52.9}}\scalebox{0.70}{$\pm$0.8}\hspace{-0.20cm} }\end{tabular}
& \begin{tabular}[c]{@{}c@{}}{ \hspace{-0.20cm}{\textbf{60.1}}\scalebox{0.70}{$\pm$0.6}\hspace{-0.20cm} }\end{tabular}
& \begin{tabular}[c]{@{}c@{}}{ \hspace{-0.20cm}{\textbf{64.1}}\scalebox{0.70}{$\pm$0.4}\hspace{-0.20cm} }\end{tabular}
& \begin{tabular}[c]{@{}c@{}}{ \hspace{-0.20cm}{\textbf{66.2}}\scalebox{0.70}{$\pm$0.3}\hspace{-0.10cm} }\end{tabular}
\\
\bottomrule
\end{tabular}
\label{tbl:overall_perf}
\end{table*}

\subsection{Main Results on Real Noisy Datasets}
\label{sec:results}

\begin{figure*}[t!]
\begin{center}
\includegraphics[width=0.95\linewidth]{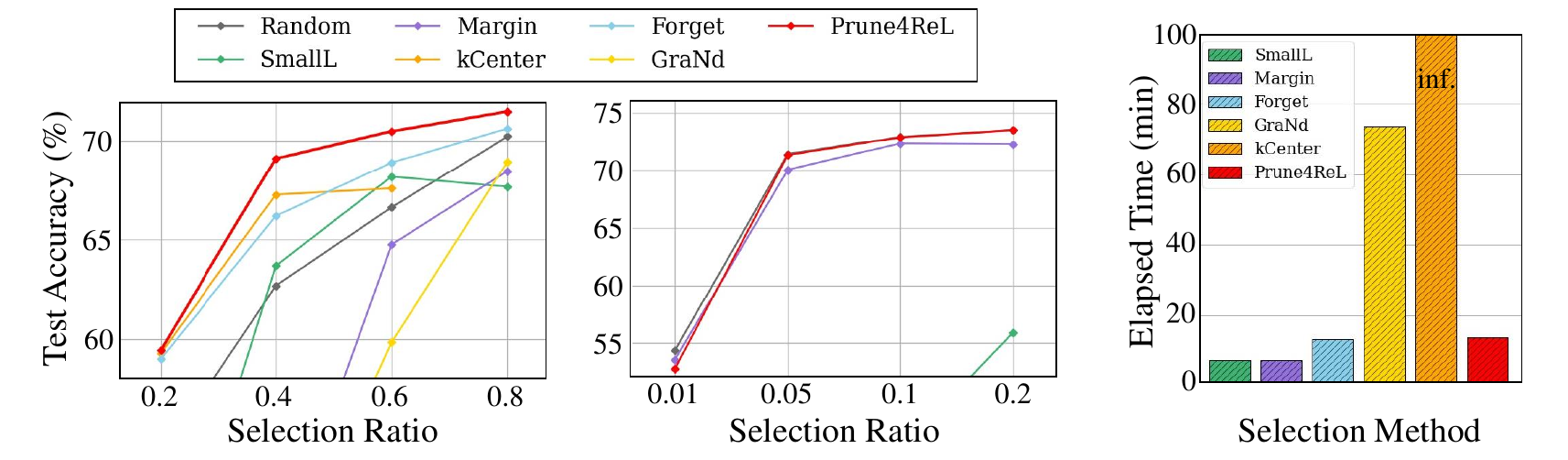}
\end{center}
\hspace*{2cm} {\small (a) WebVision.} \hspace*{2.2cm} {\small (b) Clothing-1M.} \hspace*{1.4cm} {\small (c) GPU Time for Selection.}
\caption{Data pruning performance comparison: (a) test accuracy of SOP+ trained on each selected subset of WebVision; (b) test accuracy of DivideMix trained on each selected subset of Clothing-1M; (c) elapsed GPU time for selecting a subset on WebVision with a selection ratio of 0.8.}
\label{fig:overall_performance_chart}
\end{figure*}

\noindent\textbf{Test Accuracy.}
Table \ref{tbl:overall_perf} summarizes the test accuracy of \emph{eight} baselines and \algnamec{} on CIFAR-10N and CIFAR-100N trained with two Re-labeling models.
Overall, \algnamec{} consistently achieves the best performance for all datasets across varying selection ratios. Numerically, \algnamec{} improves DivideMix and SOP+ by up to {3.7\%} and {9.1\%}, respectively.
Compared with six data pruning baselines, they show rapid performance degradation as the size of the subset decreases; most of them, which are designed to favor hard examples, tend to select a large number of noisy examples, resulting in unreliable re-labeling.
While Moderate selects moderately hard examples, it is still worse than \algnamec{} since it is not designed for noise-robust learning scenarios with Re-labeling models.
On the other hand, SmallLoss, a clean sample selection baseline, shows poor performance in CIFAR-10N (Random), because this dataset contains a relatively low noise ratio, and selecting clean examples is less critical.
Although SmallLoss shows robust performance in CIFAR-100N, it is worse than \algnamec{} because it loses many informative noisy examples that help generalization if re-labeled correctly.
Meanwhile, Uniform is a fairly robust baseline as it selects easy (clean) and hard (noisy) examples in a balanced way; many selected noisy examples may be relabeled correctly by other selected clean neighbors, resulting in satisfactory test accuracy.

Similarly, Figures \ref{fig:overall_performance_chart}(a) and \ref{fig:overall_performance_chart}(b) visualize the efficacy of the baselines and \algnamec{} on the WebVision and Clothing-1M datasets.
We train SOP+ on WebVision and DivideMix on Clothing-1M.
Similar to CIFAR-N datasets, \algnamec{} achieves better performance than existing baselines on two datasets.
Quantitatively, \algnamec{} outperforms the existing sample selection methods by up to 2.7\% on WebVision with a selection ratio of 0.4.
This result confirms that the subset selected by \algnamec{}, which maximizes the total neighborhood confidence of the training set, successfully maintains the performance of Re-labeling models and is effective for model generalization.

\noindent\textbf{Efficiency.}
In Figure \ref{fig:overall_performance_chart}(c), we further show the GPU time taken for selecting subsets within the warm-up training.
We train SOP+ on WebVision with a selection ratio of 0.8.
Powered by our efficient greedy approximation, \algnamec{} fastly prunes the dataset in a reasonable time.
GraNd takes almost 10 times longer than SmallLoss or Margin, since it trains the warm-up classifier multiple times for the ensemble.
kCenterGreedy is infeasible to run in our GPU configuration due to its huge computation and memory costs.


\def\arraystretch{1.15}
\begin{table*}[t!]
\scriptsize
\centering
\caption{Performance comparison of the standard cross-entropy model and Re-labeling models when combined with data pruning methods on CIFAR-10N and CIFAR-100N.} %
\begin{tabular}[c]
{@{}c|c|cccc|cccc|ccccc@{}}
\toprule
\multirow{3}{*}{\hspace{-0.0cm}\makecell[c]{Learning\\Models}\hspace{-0.10cm}}&\multirow{3}{*}{\hspace{-0.30cm}\makecell[c]{Selection\\Methods}\hspace{-0.30cm}}& \multicolumn{8}{c|}{{\underline{CIFAR-10N}}} 
& \multicolumn{4}{c}{{\underline{CIFAR-100N}}}\\
& & \multicolumn{4}{c|}{{Random}} & \multicolumn{4}{c|}{{Worst}} & \multicolumn{4}{c}{{Noisy}} \\
& & \multicolumn{1}{c}{\hspace{-0.20cm}{0.2}}
& \multicolumn{1}{c}{\hspace{-0.20cm}{0.4}}
& \multicolumn{1}{c}{\hspace{-0.20cm}{0.6}}
& \multicolumn{1}{c|}{\hspace{-0.20cm}{0.8}}
& \multicolumn{1}{c}{\hspace{-0.20cm}{0.2}}
& \multicolumn{1}{c}{\hspace{-0.20cm}{0.4}}
& \multicolumn{1}{c}{\hspace{-0.20cm}{0.6}}
& \multicolumn{1}{c|}{\hspace{-0.20cm}{0.8}}
& \multicolumn{1}{c}{\hspace{-0.20cm}{0.2}}
& \multicolumn{1}{c}{\hspace{-0.20cm}{0.4}}
& \multicolumn{1}{c}{\hspace{-0.20cm}{0.6}}
& \multicolumn{1}{c}{\hspace{-0.20cm}{0.8}}
\\ \midrule
\multirow{3}{*}{\hspace{-0.0cm}{CE}\hspace{-0.10cm}}                           
& \begin{tabular}[c]{@{}c@{}}{\hspace{-0.10cm}Uniform\hspace{-0.30cm}}\end{tabular}
& \begin{tabular}[c]{@{}c@{}}{ \hspace{-0.20cm}{75.6}\scalebox{0.68}{$\pm$0.1}\hspace{-0.20cm} }\end{tabular}
& \begin{tabular}[c]{@{}c@{}}{ \hspace{-0.20cm}{81.0}\scalebox{0.68}{$\pm$0.1}\hspace{-0.20cm} }\end{tabular}
& \begin{tabular}[c]{@{}c@{}}{ \hspace{-0.20cm}{83.0}\scalebox{0.68}{$\pm$0.1}\hspace{-0.20cm} }\end{tabular}
& \begin{tabular}[c]{@{}c@{}}{ \hspace{-0.20cm}{84.3}\scalebox{0.68}{$\pm$0.5}\hspace{-0.10cm} }\end{tabular}
& \begin{tabular}[c]{@{}c@{}}{ \hspace{-0.20cm}{58.6}\scalebox{0.68}{$\pm$0.3}\hspace{-0.20cm} }\end{tabular}
& \begin{tabular}[c]{@{}c@{}}{ \hspace{-0.20cm}{63.9}\scalebox{0.68}{$\pm$0.2}\hspace{-0.20cm} }\end{tabular}
& \begin{tabular}[c]{@{}c@{}}{ \hspace{-0.20cm}{66.4}\scalebox{0.68}{$\pm$0.1}\hspace{-0.20cm} }\end{tabular}
& \begin{tabular}[c]{@{}c@{}}{ \hspace{-0.20cm}{67.5}\scalebox{0.68}{$\pm$0.5}\hspace{-0.10cm} }\end{tabular}
& \begin{tabular}[c]{@{}c@{}}{ \hspace{-0.20cm}{37.4}\scalebox{0.68}{$\pm$0.2}\hspace{-0.20cm} }\end{tabular}
& \begin{tabular}[c]{@{}c@{}}{ \hspace{-0.20cm}{46.1}\scalebox{0.68}{$\pm$0.2}\hspace{-0.20cm} }\end{tabular}
& \begin{tabular}[c]{@{}c@{}}{ \hspace{-0.20cm}{50.1}\scalebox{0.68}{$\pm$0.0}\hspace{-0.20cm} }\end{tabular}
& \begin{tabular}[c]{@{}c@{}}{ \hspace{-0.20cm}{52.3}\scalebox{0.68}{$\pm$0.1}\hspace{-0.10cm} }\end{tabular}
\\
& \begin{tabular}[c]{@{}c@{}}{\hspace{-0.10cm}SmallL\hspace{-0.30cm}}\end{tabular}
& \begin{tabular}[c]{@{}c@{}}{ \hspace{-0.20cm}{75.0}\scalebox{0.68}{$\pm$1.5}\hspace{-0.20cm} }\end{tabular}
& \begin{tabular}[c]{@{}c@{}}{ \hspace{-0.20cm}{83.4}\scalebox{0.68}{$\pm$0.6}\hspace{-0.20cm} }\end{tabular}
& \begin{tabular}[c]{@{}c@{}}{ \hspace{-0.20cm}{87.5}\scalebox{0.68}{$\pm$0.3}\hspace{-0.20cm} }\end{tabular}
& \begin{tabular}[c]{@{}c@{}}{ \hspace{-0.20cm}{90.1}\scalebox{0.68}{$\pm$0.2}\hspace{-0.10cm} }\end{tabular}
& \begin{tabular}[c]{@{}c@{}}{ \hspace{-0.20cm}{70.1}\scalebox{0.68}{$\pm$0.3}\hspace{-0.20cm} }\end{tabular}
& \begin{tabular}[c]{@{}c@{}}{ \hspace{-0.20cm}{77.5}\scalebox{0.68}{$\pm$1.6}\hspace{-0.20cm} }\end{tabular}
& \begin{tabular}[c]{@{}c@{}}{ \hspace{-0.20cm}{80.6}\scalebox{0.68}{$\pm$0.5}\hspace{-0.20cm} }\end{tabular}
& \begin{tabular}[c]{@{}c@{}}{ \hspace{-0.20cm}{76.4}\scalebox{0.68}{$\pm$0.1}\hspace{-0.10cm} }\end{tabular}
& \begin{tabular}[c]{@{}c@{}}{ \hspace{-0.20cm}{42.2}\scalebox{0.68}{$\pm$0.4}\hspace{-0.20cm} }\end{tabular}
& \begin{tabular}[c]{@{}c@{}}{ \hspace{-0.20cm}{54.7}\scalebox{0.68}{$\pm$0.5}\hspace{-0.20cm} }\end{tabular}
& \begin{tabular}[c]{@{}c@{}}{ \hspace{-0.20cm}{57.7}\scalebox{0.68}{$\pm$0.1}\hspace{-0.20cm} }\end{tabular}
& \begin{tabular}[c]{@{}c@{}}{ \hspace{-0.20cm}{57.4}\scalebox{0.68}{$\pm$0.2}\hspace{-0.10cm} }\end{tabular}
\\
& \begin{tabular}[c]{@{}c@{}}{\hspace{-0.10cm}Forget\hspace{-0.30cm}}\end{tabular}
& \begin{tabular}[c]{@{}c@{}}{ \hspace{-0.20cm}{82.2}\scalebox{0.68}{$\pm$0.5}\hspace{-0.20cm} }\end{tabular}
& \begin{tabular}[c]{@{}c@{}}{ \hspace{-0.20cm}{86.2}\scalebox{0.68}{$\pm$0.1}\hspace{-0.20cm} }\end{tabular}
& \begin{tabular}[c]{@{}c@{}}{ \hspace{-0.20cm}{86.1}\scalebox{0.68}{$\pm$0.0}\hspace{-0.20cm} }\end{tabular}
& \begin{tabular}[c]{@{}c@{}}{ \hspace{-0.20cm}{85.4}\scalebox{0.68}{$\pm$0.2}\hspace{-0.10cm} }\end{tabular}
& \begin{tabular}[c]{@{}c@{}}{ \hspace{-0.20cm}{71.2}\scalebox{0.68}{$\pm$0.3}\hspace{-0.20cm} }\end{tabular}
& \begin{tabular}[c]{@{}c@{}}{ \hspace{-0.20cm}{73.3}\scalebox{0.68}{$\pm$0.2}\hspace{-0.20cm} }\end{tabular}
& \begin{tabular}[c]{@{}c@{}}{ \hspace{-0.20cm}{71.4}\scalebox{0.68}{$\pm$0.2}\hspace{-0.20cm} }\end{tabular}
& \begin{tabular}[c]{@{}c@{}}{ \hspace{-0.20cm}{69.6}\scalebox{0.68}{$\pm$0.1}\hspace{-0.10cm} }\end{tabular}
& \begin{tabular}[c]{@{}c@{}}{ \hspace{-0.20cm}{43.5}\scalebox{0.68}{$\pm$0.4}\hspace{-0.20cm} }\end{tabular}
& \begin{tabular}[c]{@{}c@{}}{ \hspace{-0.20cm}{54.5}\scalebox{0.68}{$\pm$0.1}\hspace{-0.20cm} }\end{tabular}
& \begin{tabular}[c]{@{}c@{}}{ \hspace{-0.20cm}{57.5}\scalebox{0.68}{$\pm$0.4}\hspace{-0.20cm} }\end{tabular}
& \begin{tabular}[c]{@{}c@{}}{ \hspace{-0.20cm}{56.6}\scalebox{0.68}{$\pm$0.3}\hspace{-0.10cm} }\end{tabular}
\\
\midrule
\multirow{1}{*}{\hspace{-0.0cm}{DivMix}\hspace{-0.10cm}} 
& \begin{tabular}[c]{@{}c@{}}{\hspace{-0.10cm}{$\text{\tblalgnamec{}}_B$}\hspace{-0.50cm}}\end{tabular}
& \begin{tabular}[c]{@{}c@{}}{ \hspace{-0.20cm}{\textbf{88.1}}\scalebox{0.68}{$\pm$0.3}\hspace{-0.20cm} }\end{tabular}
& \begin{tabular}[c]{@{}c@{}}{ \hspace{-0.20cm}{\textbf{93.0}}\scalebox{0.68}{$\pm$0.2}\hspace{-0.20cm} }\end{tabular}
& \begin{tabular}[c]{@{}c@{}}{ \hspace{-0.20cm}{\textbf{94.5}}\scalebox{0.68}{$\pm$0.2}\hspace{-0.20cm} }\end{tabular}
& \begin{tabular}[c]{@{}c@{}}{ \hspace{-0.20cm}{\textbf{95.1}}\scalebox{0.68}{$\pm$0.1}\hspace{-0.10cm} }\end{tabular}
& \begin{tabular}[c]{@{}c@{}}{ \hspace{-0.20cm}{\textbf{83.7}}\scalebox{0.68}{$\pm$0.4}\hspace{-0.20cm} }\end{tabular}
& \begin{tabular}[c]{@{}c@{}}{ \hspace{-0.20cm}{\textbf{88.6}}\scalebox{0.68}{$\pm$0.4}\hspace{-0.20cm} }\end{tabular}
& \begin{tabular}[c]{@{}c@{}}{ \hspace{-0.20cm}{\textbf{90.8}}\scalebox{0.68}{$\pm$0.2}\hspace{-0.20cm} }\end{tabular}
& \begin{tabular}[c]{@{}c@{}}{ \hspace{-0.20cm}{\textbf{92.4}}\scalebox{0.68}{$\pm$0.2}\hspace{-0.10cm} }\end{tabular}
& \begin{tabular}[c]{@{}c@{}}{ \hspace{-0.20cm}{39.4}\scalebox{0.68}{$\pm$0.8}\hspace{-0.20cm} }\end{tabular}
& \begin{tabular}[c]{@{}c@{}}{ \hspace{-0.20cm}{\textbf{56.3}}\scalebox{0.68}{$\pm$0.5}\hspace{-0.20cm} }\end{tabular}
& \begin{tabular}[c]{@{}c@{}}{ \hspace{-0.20cm}{\textbf{63.5}}\scalebox{0.68}{$\pm$0.3}\hspace{-0.20cm} }\end{tabular}
& \begin{tabular}[c]{@{}c@{}}{ \hspace{-0.20cm}{\textbf{67.4}}\scalebox{0.68}{$\pm$0.7}\hspace{-0.10cm} }\end{tabular}
\\ 
\multirow{1}{*}{\hspace{-0.0cm}{SOP+}\hspace{-0.30cm}}                             
& \begin{tabular}[c]{@{}c@{}}{\hspace{-0.10cm}{$\text{\tblalgnamec{}}_B$}\hspace{-0.50cm}}\end{tabular}
& \begin{tabular}[c]{@{}c@{}}{ \hspace{-0.20cm}{\textbf{88.5}}\scalebox{0.68}{$\pm$0.3}\hspace{-0.20cm} }\end{tabular}
& \begin{tabular}[c]{@{}c@{}}{ \hspace{-0.20cm}{\textbf{93.1}}\scalebox{0.68}{$\pm$0.2}\hspace{-0.20cm} }\end{tabular}
& \begin{tabular}[c]{@{}c@{}}{ \hspace{-0.20cm}{\textbf{94.4}}\scalebox{0.68}{$\pm$0.1}\hspace{-0.20cm} }\end{tabular}
& \begin{tabular}[c]{@{}c@{}}{ \hspace{-0.20cm}{\textbf{95.3}}\scalebox{0.68}{$\pm$0.1}\hspace{-0.10cm} }\end{tabular}
& \begin{tabular}[c]{@{}c@{}}{ \hspace{-0.20cm}{\textbf{84.9}}\scalebox{0.68}{$\pm$0.6}\hspace{-0.20cm} }\end{tabular}
& \begin{tabular}[c]{@{}c@{}}{ \hspace{-0.20cm}{\textbf{89.2}}\scalebox{0.68}{$\pm$0.6}\hspace{-0.20cm} }\end{tabular}
& \begin{tabular}[c]{@{}c@{}}{ \hspace{-0.20cm}{\textbf{91.3}}\scalebox{0.68}{$\pm$0.3}\hspace{-0.20cm} }\end{tabular}
& \begin{tabular}[c]{@{}c@{}}{ \hspace{-0.20cm}{\textbf{92.9}}\scalebox{0.68}{$\pm$0.1}\hspace{-0.10cm} }\end{tabular}
& \begin{tabular}[c]{@{}c@{}}{ \hspace{-0.20cm}{\textbf{52.9}}\scalebox{0.68}{$\pm$0.8}\hspace{-0.20cm} }\end{tabular}
& \begin{tabular}[c]{@{}c@{}}{ \hspace{-0.20cm}{\textbf{60.1}}\scalebox{0.68}{$\pm$0.6}\hspace{-0.20cm} }\end{tabular}
& \begin{tabular}[c]{@{}c@{}}{ \hspace{-0.20cm}{\textbf{64.1}}\scalebox{0.68}{$\pm$0.4}\hspace{-0.20cm} }\end{tabular}
& \begin{tabular}[c]{@{}c@{}}{ \hspace{-0.20cm}{\textbf{66.2}}\scalebox{0.68}{$\pm$0.3}\hspace{-0.10cm} }\end{tabular}
\\
\bottomrule
\end{tabular}
\label{tbl:perf_standard}
\vspace{0.3cm}
\end{table*}

\subsection{Necessity of Data Pruning with Re-labeling under Label Noise}
\label{sec:why_relabeling}

Table \ref{tbl:perf_standard} shows the superiority of the Re-labeling models over the standard learning model, \emph{i.e.}, only with the cross-entropy loss, for data pruning under label noise.
When combined with the data pruning methods, the performance of the Re-labeling models such as DivideMix and SOP+ significantly surpasses those of the standard models on CIFAR-10N and CIFAR-100N by up to 21.6$\%$. That is, the re-labeling capacity can be well preserved by a proper data pruning strategy.
This result demonstrates the necessity of data pruning for the Re-labeling models in the presence of noisy labels.

\begin{table*}[!t]
\centering
\captionof{table}{Effect of the confidence metrics on \algnamec{}.}
\vspace*{+0.05cm}
\scriptsize
\begin{tabular}{c|c|c|c c c c}\toprule
\multirow{2}{*}{\makecell[c]{\!\!\!Re-label\\\!\!\!Model}}&\multirow{2}{*}{\makecell[c]{\!\!\!Dataset\!\!\!}}& \multirow{2}{*}{\makecell[c]{\!\!\!Conf. Metric\!\!\!}} & \multicolumn{4}{c}{Selection Ratio} \\
& & & 0.2&0.4&0.6&0.8\\ \midrule \addlinespace[0.2ex]
\multirow{4}{*}{\makecell[c]{\!\!\!SOP+\!\!\!}}& \multirow{2}{*}{\makecell[c]{\!\!\!CIFAR-10N\\(Worst)\!\!\!}}& MaxProb & 82.7&88.1&91.3&92.5\\
& & DiffProb &82.5&88.5&91.2&92.5\\ \addlinespace[0.6ex] \cline{2-7} \addlinespace[0.5ex]
& \multirow{2}{*}{\makecell[c]{\!\!\!CIFAR-100N\!\!\!}}& MaxProb & 50.2&59.1&63.9&65.7\\
& & DiffProb &49.2&59.3&64.1&66.0\\ \addlinespace[0.2ex] \bottomrule 
\end{tabular}
\label{tab:conf_metrics}
\end{table*}

\begin{figure}[!t]
\centering
\includegraphics[width=0.45\textwidth, ]{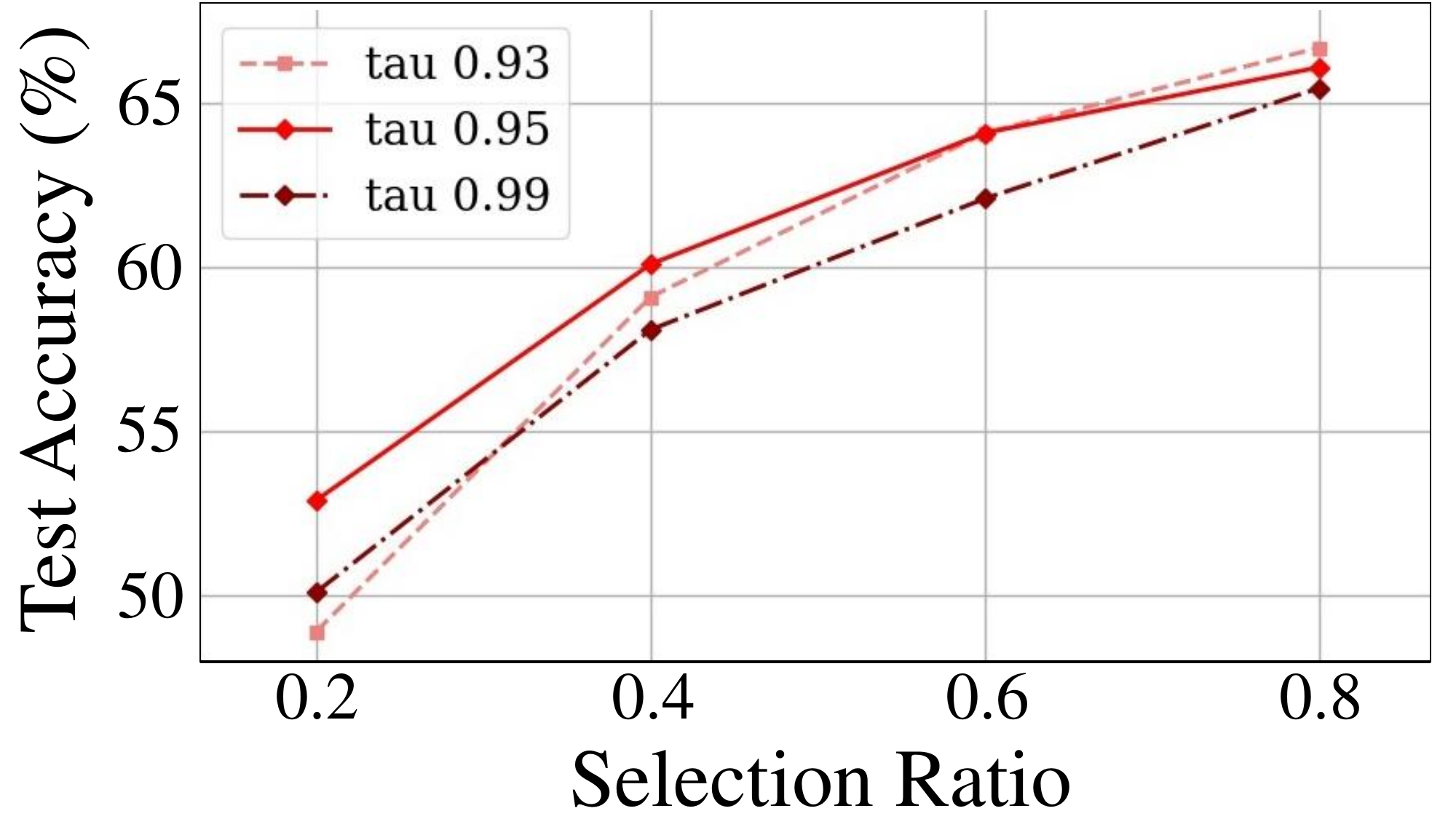}
\captionof{figure}{Effect of the neighborhood threshold $\tau$ on $\text{\algnamec{}}_B$.}
\label{fig:neighbor_sz}
\end{figure}

\subsection{Ablation Studies}
\label{sec:ablation}

\noindent\textbf{Effect of Confidence Metrics.}
\algnamec{} can be integrated with various metrics for the confidence of predictions in Eq.~\eqref{eq:objective}.
In our investigation, we consider two widely-used metrics: (1) MaxProb, which represents the maximum value of softmax probability and (2) DiffProb, which measures the difference between the highest and second highest softmax probabilities.
Table \ref{tab:conf_metrics} shows the effect of the two confidence metrics on the test accuracy of SOP+ on CIFAR-10N\,(Worst) and CIFAR-100N.
The result indicates that both metrics perform similarly and yield higher accuracy compared to existing data pruning baselines, which demonstrates \algnamec{} is open to the choice of the confidence metric.


\noindent\textbf{Effect of Neighborhood Size.}
\algnamec{} involves a hyperparameter $\tau$ in Eq.~\eqref{eq:objective} to determine the neighborhood of each example, where a larger\,(smaller) value introduces a fewer\,(more) number of neighbors.
Figure \ref{fig:neighbor_sz} shows the effect of $\tau\in\{0.93, 0.95, 0.99\}$ on the test accuracy of SOP+ trained on CIFAR-100N with varying selection ratios.
In general, \algnamec{} shows better or comparable performance than other sample selection baselines.
Among them, \algnamec{} with $\tau=0.95$ shows a satisfactory accuracy across varying selection ratios.
With a smaller value of $\tau=0.93$, it shows the best performance in a high selection ratio of 0.8, but it becomes less effective in low selection ratios, due to the increasing influence of noisy examples caused by a relatively large neighborhood size. 
On the contrary, with a large value of $\tau=0.99$, it primarily selects clean yet easy examples due to a small neighborhood size, leading to a relatively less improvement in performance.

\begin{table*}[!t]
\noindent
\small
\caption{Ratio ($\%$) of noisy examples in the selected subset.} 
\vspace*{0.2cm}
\centering
\scriptsize
\begin{tabular}{c|c|c c c c|c c c c|c c c c}\toprule
\multirow{2}{*}{\hspace{-0.0cm}\makecell[c]{Re-label\\Model}\hspace{-0.10cm}}&\multirow{2}{*}{\makecell[c]{Selection\\Methods}}& \multicolumn{4}{c|}{\underline{CIFAR-10N (Random, $\approx\!18\%$)}} & \multicolumn{4}{c|}{\underline{CIFAR-10N (Worst, $\approx\!40\%$)}} & \multicolumn{4}{c}{\underline{CIFAR-100N (Noisy, $\approx\!40\%$)}} \\
& & 0.2 & 0.4 & 0.6 & 0.8 & 0.2 & 0.4 & 0.6 & 0.8 & 0.2 & 0.4 & 0.6 & 0.8 \\ \midrule
\multirow{8}{*}{SOP+}&SmallL & 0.1 & 0.2 & 1.0 & 4.0 & 0.8 & 3.6 & 11.5 & 26.2 & 3.5 & 8.7 & 16.3 & 27.4 \\
&Margin & 29.7 & 25.7 & 22.5 & 19.7 & 54.6 & 52.1 & 48.5 & 44.6 & 61.5 & 56.8 & 51.6 & 46.2 \\
&kCenter& 19.0 & 18.8 & 19.1 & 18.6 & 40.0 & 41.1 & 41.6 & 42.1 & 37.5 & 38.7 & 39.9 & 40.4 \\
&Forget& 17.0 & 17.7 & 17.5 & 17.8 & 37.7 & 38.8 & 39.2 & 40.2 & 37.9 & 34.6 & 33.0 & 36.8 \\
&GraNd& 67.5 & 41.7 & 28.6 & 21.5 & 91.5 & 79.4 & 64.2 & 50.1 & 93.9 & 57.3 & 61.2 & 49.3 \\
&SSP& 25.2 & 23.7 & 21.6 & 19.5 & 48.5 & 46.4 & 43.2 & 42.1 & 52.7 & 52.3 & 46.8 & 43.0 \\
&Moderate& 6.6 & 7.1 & 8.4 & 13.5 & 31.7 & 33.4 & 34.7 & 40.0 & 33.2 & 54.6 & 60.2 & 64.6 \\ 
&\textbf{\tblalgnamec{}}& 17.0 & 18.7 & 19.3 & 22.5 & 38.7 & 42.7 & 43.5 & 46.5 & 28.3 & 29.1 & 33.3 & 37.2 \\ \bottomrule
\end{tabular}
\vspace*{0.2cm}
\label{table:noise_ratio_in_subset}
\end{table*}

\subsection{In-depth Analysis of Noisy Examples in Selected Subset}
\label{sec:subset_analysis}

\noindent\textbf{Noise Ratio of Selected Subset.}
Table \ref{table:noise_ratio_in_subset} shows the ratio of noisy examples in the subset selected by each sample selection method.
SmallLoss shows a very low ratio of noisy examples in the subset because it prefers clean examples.
Many data pruning methods, including Margin, GraNd, SSP, and Moderate (for CIFAR-100N), tend to select a higher ratio of noisy examples compared with that of each original dataset since they prefer to select hard examples.
On the other hand, \algnamec{} selects a low ratio of noisy examples when the subset size is small and gradually increases the noise ratio as the subset size increases.
This result indicates that \algnamec{} expands the confident subset through Algorithm \ref{alg:greedy}---\emph{i.e.}, selecting the most confident (clean) examples first and then trying to select less confident (hard or noisy) neighbors to ensure accurate re-labeling.
While some baselines, such as kCenterGreedy, Forget, and Moderate (for CIFAR-10N), also select a somewhat low ratio of noisy examples, their data pruning performances are worse than \algnamec{} because the quality (or self-correctability) of noisy examples is not considered when selecting the subset, which is further investigated in Table \ref{table:self_correctability}.


\noindent\textbf{Self-correctability of Selected Noisy Examples.}
Table \ref{table:self_correctability} shows the self-correctability of selected subsets, which indicates the ratio of correctly re-labeled noisy examples out of all selected noisy examples.
Here, we compare \algnamec{} with kCenterGreedy and Forgetting on CIFAR-10N (Random) with the selection ratio of 0.2, where the ratio of noisy examples (\emph{i.e.}, $\% Noisy$) in the selected subset of each method is similar (\emph{i.e.}, from 17$\%$ to 19$\%$).
Although these methods select almost equal amounts of noisy examples, there were differences in the self-correctability (\emph{i.e.}, $\% Correct$) of the selected subsets.
Noisy examples selected by \algnamec{} are mostly self-correctable as it maximizes the total neighborhood confidence of the training set. In contrast, those selected by existing data pruning methods such as kCenter and Forget are not guaranteed to be self-correctable.
This result confirms that \algnamec{} not only selects a low ratio of noisy examples but also considers the quality of the selected subset in terms of maximizing re-labeling accuracy.
Therefore, \algnamec{} fully takes advantage of the Re-labeling methods.


\subsection{Results on ImageNet-N with Synthetic Label Noise}
\label{sec:results_imagenet}

We further validate the efficacy of \algnamec{} on ImageNet-N by injecting synthetic label noise of 20$\%$ to the commonly-used benchmark dataset ImageNet-1K. 
Table \ref{table:results_imagenet} shows the test accuracy of \algnamec{} and three representative sample selection baselines with varying selection ratios of $\{0.05, 0.1, 0.2, 0.4\}$.
Similar to the result in Section \ref{sec:results}, \algnamec{} consistently outperforms the baselines by up to 8.6$\%$, thereby adding more evidence of the superiority of \algnamec{}.
In addition, owing to its great computation efficiency, \algnamec{} is able to scale to ImageNet-N, a large-scale dataset with approximately 1.2M training examples.

\begin{table*}[!t]
\captionof{table}{Ratio of correctly re-labeled noisy examples in the selected subset (denoted as \emph{$\%$ Correct}).}
\centering
\scriptsize
\begin{tabular}{c|c|c c c}\toprule
\multirow{2}{*}{\hspace{-0.0cm}\makecell[c]{Re-label\\Model}\hspace{-0.10cm}}&\multirow{2}{*}{\makecell[c]{Selection\\Methods}}& \multicolumn{3}{c}{\underline{CIFAR-10N (Random)}} \\
& &\!\!\emph{Test Acc.}\!\!\!&\!\!\!\emph{$\%$~Noisy}\!\!\!&\!\!\!\emph{$\%$~Correct}\!\!\\ \midrule
\multirow{3}{*}{SOP+}&kCenter & 86.3 & 19.0 & 75.2 \\
&Forget& 82.4 & 17.0 & 61.7\\
&\textbf{\text{\tblalgnamec{}}}& {88.1} & {17.0} & \textbf{90.3}\\ \bottomrule
\end{tabular}
\label{table:self_correctability}
\end{table*}

\begin{table*}[!t]
\captionof{table}{Data pruning performance on ImageNet with a 20$\%$ synthetic label noise.}
\centering
\scriptsize
\begin{tabular}{c|c|c c c c}\toprule
\multirow{2}{*}{\hspace{-0.0cm}\makecell[c]{Re-label\\Model}\hspace{-0.10cm}}&\multirow{2}{*}{\makecell[c]{\!Selection\!\\Methods}}& \multicolumn{4}{c}{\underline{ImageNet-1K (Syn, $\approx\!20\%$)}} \\
& & 0.05 & 0.1 & 0.2 & 0.4 \\ \midrule
\multirow{4}{*}{SOP+}&Uniform & 27.8 & 42.5 & 52.7 & 59.2 \\
&SmallL & 22.8 & 31.4 & 42.7 & 54.4\\
&Forget& 4.1 & 8.3 & 50.6 & 57.2\\ 
&\textbf{$\text{\tblalgnamec{}}_B$}& \textbf{30.2} & \textbf{44.3} & \textbf{53.5} & \textbf{60.0}\\ \bottomrule
\end{tabular}
\label{table:results_imagenet}
\end{table*}
\section{Conclusion and Future Work}
\label{sec:conclusion3}
\vspace*{-0.2cm}

In this chapter, we present a noise-robust data pruning method for Re-labeling called \algnamec{} that finds a subset that maximizes the total neighborhood confidence of the training examples, thereby maximizing the re-labeling accuracy and generalization performance.
To identify a subset that maximizes the re-labeling accuracy, \algnamec{} introduces a novel metric, \textit{reduced neighborhood confidence} which is the prediction confidence of each neighbor example in the selected subset, and the effectiveness of this metric in estimating the Re-labeling capacity of a subset is theoretically and empirically validated. Furthermore, we optimize \algnamec{} with an efficient greedy algorithm that expands the subset by selecting the example that contributes the most to increasing the total neighborhood confidence. Experimental evaluations demonstrate the substantial superiority of \algnamec{} compared to existing pruning methods in the presence of label noise.

Although \algnamec{} has demonstrated consistent effectiveness in the classification task with real and synthetic label noises, we have not validated its applicability on datasets with open-set noise or out-of-distribution examples\,\cite{yu2020multi, park2021task}.
Also, we have not validated its applicability to state-of-the-art deep learning models, such as large language models\,\cite{brown2020language} and vision-language models\,\cite{radford2021learning}.
This verification would be valuable because the need for data pruning in the face of annotation noise is consistently high across a wide range of real-world tasks. 
In addition, \algnamec{} has not been validated in other realistic applications of data pruning, such as continual learning\,\cite{de2021continual} and neural architecture search\,\cite{elsken2019neural}. In these scenarios, selecting informative examples is very important, and we leave them for future research.

\chapter{Prioritizing Informative Examples for Instruction Selection from Labeled Text Noisy Data}
\label{chap:part_4}
\section{Overview}
\label{sec:overview4}

Aligning large language models (LLMs) with human preferences is essential to enhance LLMs' ability to understand human instructions and generate proper responses.
\emph{Instruction tuning}, which aligns LLMs on instruction datasets composed of question-answer pairs by fine-tuning, has been shown to significantly enhance the zero-shot performance of LLMs\,\cite{zhang2023instruction, wei2021finetuned}.
Accordingly, many instruction datasets covering wide domains of knowledge are being actively released and their size is growing exponentially, e.g., the {\sc Super-NaturalInstructions} dataset contains 5M instructions collected from more than 1K tasks\,\cite{wang2022self, wang2022super}.

Despite the great success of aligned LLMs, their performances are highly contingent on the quality of the instruction dataset as it contains many \emph{uninformative} or \emph{redundant} instructions\,\cite{wang2023far}.
Recently, a few studies have attempted to combat this data quality problem in LLMs by \emph{instruction selection}.
They show that fine-tuned LLMs on small but high-quality instructions, selected manually by humans\,\cite{zhou2023lima} or automatically with a strong LLM such as ChatGPT\,\cite{chen2023alpagasus, cao2023instruction}, generate responses more preferable to humans for open-ended questions\,\cite{alpaca_eval}.

While prior selection approaches can guide LLMs to generate human-preferable responses, we found that they often lose the \emph{factuality} of the responses.
That is, when applying the prior selection approaches to the Alpaca instruction dataset\,\cite{alpaca}, the performances of aligned LLMs are degraded on the factuality benchmarks such as MMLU\,\cite{hendrycks2020measuring, openllm}.
Since generating factual and clear responses is crucial in practical usage, this calls for a new instruction selection approach that can \emph{both} enhance the factuality and preferability of LLMs.

To this end, we first provide a comprehensive study of which factors in instruction selection matter for the factuality of LLM's responses.
We explored that both factuality are affected by three factors such as cleanness, diversity, and quality of selected instructions.
That is, by fine-tuning LLMs on instructions with higher cleanness, diversity, or quality, the aligned LLMs become more factual.

Based on this observation, we propose a new instruction selection framework called \algnamed{}.
To diversify the selected instructions, \algnamed{} first performs clustering on embedding representations of instruction examples obtained from a teacher LLM.
Then, to ensure the cleanness and quality of selected instructions, \algnamed{} uses cluster-wise prompting from the teacher LLM that asks the LLM to rank the cleanest and most helpful instruction in each cluster.
By selecting the best instruction in each cluster, \algnamed{} can select a clean, diverse, high-quality instruction subset that leads to factual and preferable LLMs.

Experiments on both factuality (e.g., MMLU\,\cite{hendrycks2020measuring}) and preference (e.g., MMLU\,\cite{dubois2023alpacafarm}) benchmarks, \algnamed{} outperforms the existing instruction selection baselines including LIMA\,\cite{zhou2023lima} and Alpagasus\,\cite{chen2023alpagasus}, and it is the first method that enhances both factuality and preference performances.
\section{Case Study: Which Factors Affect LLM's Factuality?}
\label{sec:case_study}

\subsection{Problem Statement}
\label{sec:problem_statement_d}

We formalize a problem of instruction selection for LLMs such that it finds the most informative subset $\mathcal{S}\subset\mathcal{D}$ from the entire instruction set $\mathcal{D}$ that maximizes the alignment performance of an LLM $\theta_{\text{LLM}}(\mathcal{S})$ fine-tuned on the subset $\mathcal{S}$. Formally, we aim to find an optimal subset $\mathcal{S}^{*}$ that satisfies
\begin{equation}
\begin{gathered}
\mathcal{S}^{*} = \argmaxB_{\mathcal{S}:\ |\mathcal{S}|\leq s} ~ Q(\theta_{\text{LLM}}(\mathcal{S})) ~~ : ~~\theta_{\text{LLM}}(\mathcal{S})=\argminB_{{\theta}} ~~ \mathcal{L}_{cross-entropy}(\mathcal{S}; \theta_{\text{LLM}}),
\end{gathered}
\label{eq:goal_d}
\end{equation}
where $Q(\theta_{\text{LLM}}(\mathcal{S}))$ is the alignment performance (e.g., factuality or preferability) of the fine-tuned LLM $\theta_{\text{LLM}}(\mathcal{S})$ and $s$ is the target subset size. 
In the following case study, we explore which instruction examples' characteristics affect LLMs' alignment performance.
As the data quality issues mostly come from noise, redundant, and poor-quality instructions, we analyze three factors of instruction examples, cleanness, diversity, and quality.

\subsection{Case Study I: Cleanness}
\label{sec:case_study1}

Instruction datasets often contain noisy (i.e., hallucinated), toxic, and factually wrong contents, which may wrongly guide LLMs to generate hallucinated contents.
Here, we measure the effect of \emph{cleanness} of the instruction subset on the performance of LLMs.

\noindent\textbf{Setup.} 
To measure the effect of cleanness on instruction tuning, we use Alpaca dataset, an instruction dataset with 52k examples generated from InstructGPT, and Alpaca-halu dataset\,\cite{li2023halueval}, a hallucinated instruction dataset of Alpaca with 5k examples.
Each example in Alpaca-halu dataset can be matched with an example in Alpaca dataset, as those examples share the same question but the example from the Alpaca-halu contains a hallucinated answer.
Specifically, we first select 5k examples from Alpaca dataset containing the same questions with Alpaca-halu, and substitute a portion of 5k examples with the corresponding hallucinated examples in Alpaca-halu.
That is, we fine-tune LLaMA-2-7B on three subsets with varying cleanness, Alpaca-5k without examples from Alpaca-halu, Alpaca-5k (halu400), and Alpaca-5k (halu800) with 400 and 800 examples substituted from Alpaca-halu, respectively.
We use LLaMA-2-7B as the base LLM for tuning, and use the same training configuration as in Section\,\ref{sec:experiment4}.

\begin{table*}[h]
\noindent
\small
\vspace*{-0.0cm}
\caption{Effect of instruction cleanness for alignment on MMLU factuality benchmark.} 
\centering
\begin{tabular}{c|c c c}\toprule
Selection Methods& Alpaca-5k & Alpaca-5k (halu400) & Alpaca-5k (halu800) \\ \midrule
MMLU perf.& 27.7 & 26.8 & 26.7\\ \bottomrule
\end{tabular}
\vspace*{-0.2cm}
\label{table:cleanness}
\end{table*}

\noindent\textbf{Result.}
Table\,\ref{table:cleanness} shows the effect of cleanness in the instruction set for alignment on MMLU benchmark for factuality test\,\cite{hendrycks2020measuring}.
Overall, more hallucinated instructions result in worse performance, which demonstrates that the cleanness of the instruction set is a crucial factor for the factuality of LLMs.

\subsection{Case Study II: Diversity}
\label{sec:case_study2}

Instruction datasets often contain many redundant examples, which may induce training bias for alignment due to the data imbalance.
Thus, we measure the effect of instruction \emph{diversity} of the instruction subset on the performance of LLMs.


\noindent\textbf{Setup.} 
To measure the effect of diversity on instruction tuning, we select three subsets consisting of 9k examples each from Alpaca dataset with varying diversity, MaxCover-9k, Random-9k, and MinCover-9k.
For MaxCover-9k, we first extract all the latent embeddings of instructions in Alpaca dataset using GPT, and then apply kCenterGreedy\,\cite{sener2018active} algorithm that approximately maximizes the distance coverage of selected examples in embedding space for selection.
For Random-9k, we randomly extract 9k examples from Alpaca dataset to construct the subset.
For MinCover-9k, from a random example in Alpaca dataset, we select 9k neighbor examples to construct the subset.
We fine-tune LLaMA-2-7B on the three subsets with varying diversity.

\begin{table}[h]
\noindent
\small
\vspace*{-0.0cm}
\caption{Effect of instruction diversity for alignment on MMLU factuality benchmark.} 
\centering
\begin{tabular}{c|c c c}\toprule
Selection Methods& MaxCover-9k & Random-9k & MinCover-9k \\ \midrule
MMLU perf.& 27.7 & 27.5 & 27.1\\ \bottomrule
\end{tabular}
\vspace*{-0.2cm}
\label{table:diversity}
\end{table}

\noindent\textbf{Result.}
Table\,\ref{table:diversity} shows the effect of diversity of the instruction set for alignment on MMLU benchmark\,\cite{hendrycks2020measuring}.
Overall, as the diversity increases, the fine-tuned LLMs become more factual.
This indicates the diversity of the instruction set is also an important factor in the factuality of LLMs.

\subsection{Case Study III: Quality}
\label{sec:case_study3}

Instruction datasets often contain many unclear sentences, which may result in sub-optimal performance of LLMs.
We measure the effect of \emph{quality} of the instruction subset on the performance of LLMs.

\noindent\textbf{Setup.} 
To measure the effect of instruction quality on instruction tuning, we select three subsets containing 9k examples each with varying quality from alpaca dataset.
We utilize the GPT-measured quality scores of each instruction in the Alpaca dataset released by Alpagasus\,\cite{chen2023alpagasus}.
We construct a subset with 9k highest score examples as HighQual-9k, a random subset with 9k examples as Random-9k, and a subset with 9k lowest score as LowQual-9k.
We fine-tune LLaMA-2-7B on the constructed three subsets with varying quality.

\begin{table*}[h]
\noindent
\small
\vspace*{-0.0cm}
\caption{Effect of instruction quality for alignment on MMLU factuality benchmark.} 
\centering
\begin{tabular}{c|c c c}\toprule
Selection Methods& HighQual-9k & Random-9k & LowQual-9k \\ \midrule
MMLU perf.& 32.3 & 27.5 & 26.9\\ \bottomrule
\end{tabular}
\vspace*{-0.2cm}
\label{table:quality}
\end{table*}

\noindent\textbf{Result.}
Table\,\ref{table:quality} shows the effect of the quality of instruction set for alignment on MMLU benchmark\,\cite{hendrycks2020measuring}.
Overall, as the quality increases the fine-tuned LLMs become more factual, which indicates the quality of the instruction set is also an important factor for factual LLMs.

\section{Methodology}
\label{sec:method4}

Based on our exploration in Section \ref{sec:case_study}, we propose an automatic instruction selection approach that considers all three factors including cleanness, diversity, and quality for alignment.
To systematically incorporate the three factors into selection, we further observe that the diversity can be efficiently captured in embedding space, while the cleanness and quality should be carefully captured by leveraging a teacher LLM with ranking prompts.
Based on this observation, we propose \algnamed{} that ensures diversity by clustering and cleanness and quality by cluster-wise score prompting with a teacher LLM.
In detail, \algnamed{} performs clustering in embeddings obtained from a teacher LLM, and selects the best-scored instruction in each cluster obtained from cluster-wise score prompting to a teacher LLM.

\subsection{Chellenges for Selecting Clean, Diverse, and High-quality Instructions}
\label{sec:challenges4}

While there are various ways to select clean, diverse, and high-quality instructions, we aim to carefully incorporate three factors into our selection framework with the support of a teacher LLM such as GPT.
In general, when utilizing a teacher LLM for instruction selection, two types of LLM outputs are widely used; 1) \emph{embedding representations} and 2) \emph{answer responses} for each input instruction to validate its value for alignment.
When incorporating two types of LLM outputs for instruction selection, we observe there are two challenges that should be considered.
\begin{itemize}[leftmargin=10pt,noitemsep]
\item \textbf{Embedding Challenge:} Cleanness and quality of instruction can not be measured in embedding space but should be carefully considered using answers from a teacher LLM.
\item \textbf{Prompting Challenge:} Sample-wise prompting to the teacher LLM for instruction scoring is often coarse and redundant, so can not guarantee the fine-grained instruction selection.
\end{itemize}

\noindent\textbf{Embedding Challenge.}
We observe that noisy (i.e., hallucinated) instructions are almost impossible to distinguish from the clean instructions in the embedding space.
That is, the noisy and lower-quality instructions are very close to the clean instructions with the same context.
We reveal this by a controlled study comparing the distance between clean instructions in Alpaca dataset and the corresponding noisy instructions in Alpaca-halu dataset\,\cite{li2023halueval}.
The noisy instruction has the same question with the corresponding clean example in Alpaca dataset, but contains a different yet wrong answer.
Specifically, to calculate how close a noisy instruction is from the corresponding clean one, we obtain the distance from the clean instruction to the noisy one and to the other clean instructions in Alpaca dataset, and obtain the rank of the noisy instruction out of the other clean instructions.

\begin{figure}[h]
\begin{center}
\includegraphics[width=8cm]{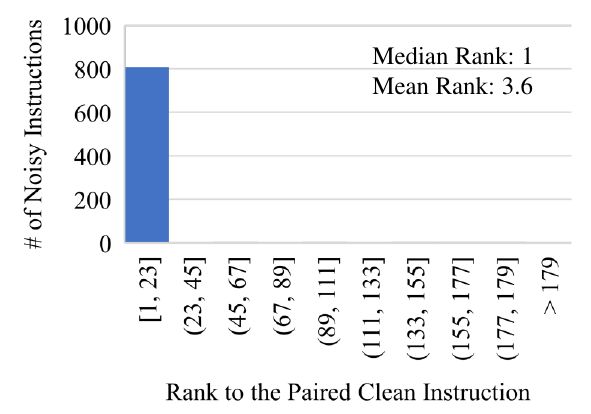}
\end{center}
\caption{Ranking histogram of noisy instructions in Alpaca-halu dataset out of all the clean instructions in Alpaca dataset.}
\label{fig:histogram}
\end{figure}

Figure\,\ref{fig:histogram} shows the ranking histogram of all noisy instructions in Alpaca-halu dataset.
It turns out that almost every noisy instruction is very close to the corresponding clean instruction with the same question, showing a median ranking of 1 and a mean ranking of 3.6.
This indicates that distinguishing the noisy instructions based solely on the embedding representations is almost impossible, so the cleanness and quality of instructions should be carefully considered using a more sophisticated way such as direct prompting to LLM.

\vspace{0.3cm}
\noindent\textbf{Prompting Challenge.}
When prompting instructions to a teacher LLM for valuation, a recent work uses \emph{sample-wise} prompting that prompts each instruction individually to the LLM and obtain a quality score in a certain range\,\cite{chen2023alpagasus}.
For example, one may prompt like "Please output a score of given instruction in a range of 0 to 5 according to its quality or helpfulness. Higher score, higher quality".
While this approach achieved a certain amount of progress in validating the quality of instructions, we observe that it can not ensure fine-grained scoring because the teacher LLM does not refer to the scores of other instructions when scoring resulting in many redundant scores for instructions with similar context.

To further quantify the coarseness of the sample-wise scores, we investigate the quality scores from GPT on Alpaca dataset obtained by Alpagasus\,\cite{chen2023alpagasus}.
Specifically, we first perform k-means clustering with $k=10000$ for all instructions on Alpaca-52k, and extract clusters with more than 5 instruction examples.
Then, within each instruction cluster of a similar context, we calculate how many numbers of instruction pairs in the clusters are non-rankable by the scores from Alpagasus.
It turns out 48.32$\%$ of pairs are non-rankable on average, meaning that the sample-wise prompting generate many redundant and coarse scores 
This result calls for a new approach to design a proper way to select clean, diverse, and high-quality instructions for alignment.

\begin{figure}[t!]
\begin{center}
\includegraphics[width=0.9\columnwidth]{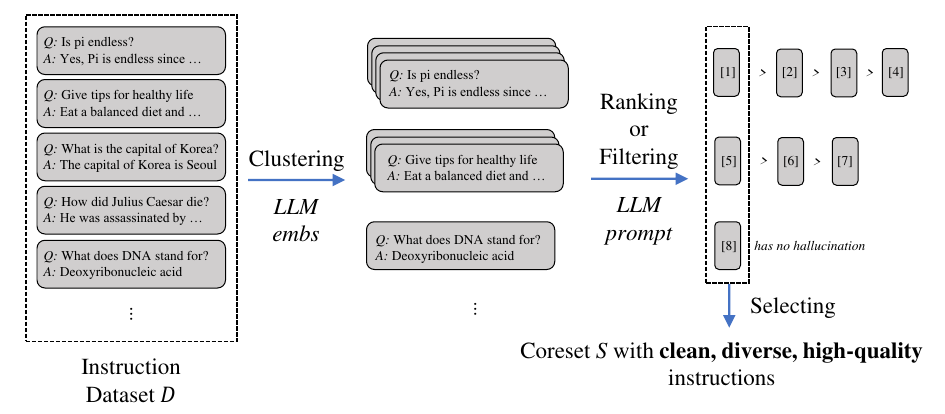}
\end{center}
\caption{Overview of \algnamed{}.}
\label{fig:overview}
\end{figure}

\subsection{\algnamed{}}
\label{sec:algnamed}

\noindent\textbf{Overview.}
Figure\,\ref{fig:overview} illustrates the key idea of \algnamed{}.
\algnamed{} consists of two steps, 1) \emph{pre-clustering} and 2) \emph{cluster-wise prompting}, to ensure clean, diverse, and high-quality instruction selection for aligning factual and preferable LLMs.
To overcome the embedding and prompting challenges, \algnamed{} first performs constrained k-means clustering in embedding space, and then performs cluster-wise prompting for fine-grained instruction selection.

\noindent\textbf{Diversifying Instructions with Pre-clustering.}
To ensure the diversity of the selected instruction subset, we first perform clustering for the entire instruction examples in $\mathcal{D}$ to find the most representative clusters of instructions that maximally cover the entire domain knowledge of $\mathcal{D}$.
Specifically, for the embedding representations of instructions obtained from GPT, we use constrained $k$-means clustering\,\cite{bradley2000constrained} that constrains the number of instructions included in each cluster to a certain value $r$.
The constrained clusters can prevent the over-grouping of many instructions to a single cluster, thereby avoiding exceeding the maximum prompt length available to the teacher LLM when applying cluster-wise prompting.

\begin{figure}[t!]
\begin{center}
\includegraphics[width=\columnwidth]{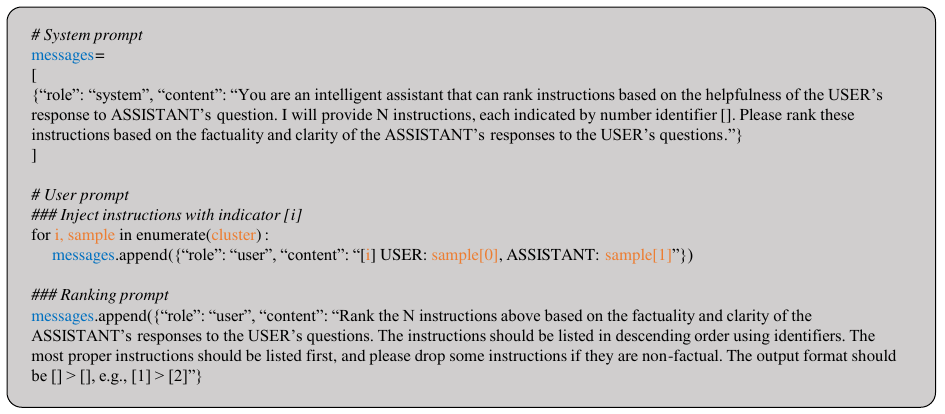}
\end{center}
\caption{Cluster-wise prompt of \algnamed{}.}
\label{fig:prompt}
\end{figure}

\noindent\textbf{Cluster-wise Prompting.}
After getting instruction clusters, we perform cluster-wise prompting to the teacher LLM to get more fine-grained quality rankings.
That is, with our cluster-wise prompting, all the instructions in each cluster are fed together into a single prompt in order to induce the LLM to compare the fine-grained quality of instructions with a similar context.

As illustrated in Figure \ref{fig:prompt}, our approach inputs a cluster of instructions into the LLM, each identified by a unique indicator (e.g., [1], [2], [3]). Then, we prompt the LLM to generate the rank of instructions based on the cleanness and quality of instructions in descending order. In answer, the instructions are ranked with the indicators following a format as [3]>[2]>[1]. For clusters with a single instruction, we prompt the LLM if the instruction contains hallucinated or wrong contents. If the LLM answers yes, we exclude the cluster for instruction selection.

\begin{algorithm}[b!]
\caption{Instruction Selection by \algnamed{}}
\label{alg:instruct_selection}
    \begin{algorithmic}[1]
    {\small 
        \REQUIRE ${\mathcal{D}}$: instruction dataset, ${\mathcal{C}}$: instruction clusters, $k$: clustering size, $r$: maximum number of instructions in a cluster
        \STATE Initialize
        $\mathcal{S}\leftarrow\emptyset;$
        \STATE $\mathcal{C} = \text{Constrained-k-means}(\mathcal{D}, k, r)$
        \STATE \textbf{for all} $c\in {\mathcal{C}}$ ~\textbf{do}
        \STATE \quad \quad $S\leftarrow LLM^{rank}_{prompt}(c)$
        \ENSURE Final selected subset $\mathcal{S}$
    }
    \end{algorithmic}
\end{algorithm}

\noindent\textbf{Instruction Selection.}
To construct a clean, diverse, and high-quality instruction subset, we select the most helpful instruction for each cluster.
In detail, for clusters with more than two instructions, we select the highest-ranked instruction into the subset. 
For clusters with a single instruction, we add the instruction if it does not contain hallucination.
Therefore, the selected subset can preserve diverse domain knowledge in the original instruction dataset, while enhancing the cleanness and quality of the instructions.
The detailed selection procedure is elaborated in Algorithm \ref{alg:instruct_selection}

\section{Experiments}
\label{sec:experiment4}

\subsection{Experiment Setting}
\label{sec:experiment_setting4}

\noindent\textbf{Datasets.} 
We perform the instruction selection task for Alpaca-52k dataset\,\cite{alpaca}, a widely used instruction dataset generated from InstructGPT consisting of 52k instructions with a wide range of domain knowledge. 
For testing factuality, we use the MMLU benchmark\,\cite{hendrycks2020measuring}, consisting of a set of questions with 57 subjects in multiple-choice format.
For testing preferability, we use the AlpacaFarm benchmark\,\cite{dubois2023alpacafarm}, consisting of 805 open-ended test questions, and GPT-4 as a judge to identify the win rate between aligned LLMs.

\noindent\textbf{Algorithms.}
We compare \algnamed{} with a random selection and two recent instruction selection baselines, LIMA\,\cite{zhou2023lima} and Alpagasus\,\cite{chen2023alpagasus}.
LIMA constructs an informative instruction set by humans and Alpagasus automatically selects helpful instructions by using a teacher LLM with sample-wise prompting.

\noindent\textbf{Implementation Details.}
We fine-tune LLaMA-2-7B for all the selection methods. Following the instruction-tuning configurations\,\cite{zhou2023lima}, we fine-tune the LLMs for 3 epochs using AdamW with $\beta_1 = 0.9, \beta_2 = 0.95$, weight decay of 0.1, and a beach size of 64. The initial learning rate is 1e-5, and it is decayed with a cosine annealing scheduler.
Texts longer than 2048 tokens are trimmed when fine-tuning.
For \algnamed{}, we set its hyperparameters $k$ as 10000 and $r$ as 10.

\begin{table*}[t!]
\noindent
\small
\vspace*{-0.0cm}
\caption{Performance of \algnamed{} over selection baselines on MMLU factuality benchmark.} 
\centering
\begin{tabular}{c|c c c c c}\toprule
Selection Methods& Vanilla LLaMA & Alpaca-52k & LIMA-1k & Alpagasus-9k & OURS-9k \\ \midrule
MMLU perf.& 26.6 & 32.5 & 32.3 & 29.6 & \textbf{38.7}\\ \bottomrule
\end{tabular}
\vspace*{-0.2cm}
\label{table:main_result4}
\end{table*}

\begin{figure}[t!]
\begin{center}
\includegraphics[width=0.65\columnwidth]{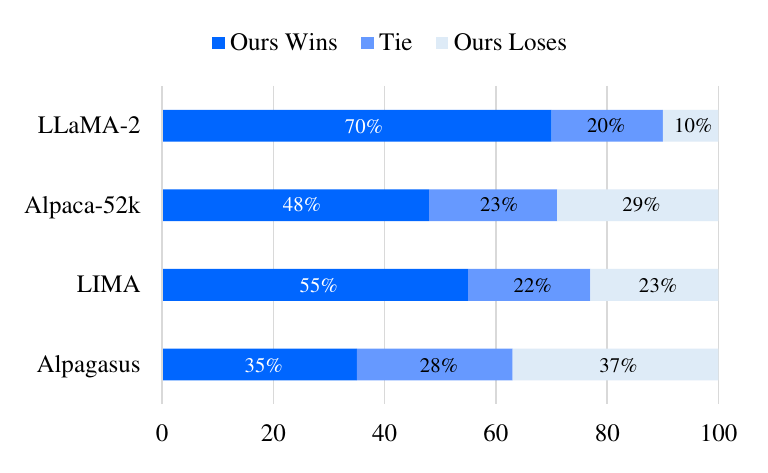}
\end{center}
\caption{Preference evaluation results using GPT-4 as a judge.}
\label{fig:preference_test}
\end{figure}

\subsection{Main Results}
\label{sec:main_results4}

\noindent\textbf{Factuality Test.}
Table \ref{table:main_result4} summarizes the factuality performance of instruction selection algorithms on MMLU benchmark. 
While LIMA and Alpagasus perform instruction selection, they lose the aligned LLM's factuality compared to the original Alpaca-52k dataset.
On the other hand, \algnamed{} is the only method that even enhances the factuality of LLaMA as it considers cleanness, diversity, and quality altogether for instruction selection.

\noindent\textbf{Preference Test.}
Figure \ref{fig:preference_test} shows the preference performance of \algnamed{} over existing instruction selection algorithms on AlpacaFarm benchmark using GPT-4 as the annotator.
LLaMA fine-tuned with \algnamed{} wins the model fine-tuned on the original Alpaca-52k, with the winning ratio of $48\%$.
Also, \algnamed{} shows better preferability than LIMA with the winning ratio of $55\%$, and shows similar preference performance with Alpagasus.
Overall, \algnamed{} is the only method that enhances factuality while maintaining the preferability of LLMs.

\section{Conclusion and Future Work}
\label{sec:conclusion4}
\vspace*{-0.2cm}

In this chapter, we present a novel instruction selection method that can align LLMs to be more factual and preferable.
To do so, we first provide a comprehensive study showing the cleanness, diversity, and quality of the selected instruction set are crucial to the factuality of LLMs.
Based on this, we propose a systemic selection framework called \algnamed{} that ensures diversity with clustering on embedding representations and ensures cleanness and quality with cluster-wise prompting for ranking instruction.
Experiments on both factuality and preference benchmark, \algnamed{} outperforms existing instruction selection algorithms and the aligned LLM with the original Alpaca-52k dataset.

As a future work, we plan to apply \algnamed{} on more instruction datasets such as {\sc Super-NaturalInstructions}\,\cite{wang2022super} to further validate its efficacy for inducing factual and preferable alignment of LLMs. Also, we plan to extend our study to more LLM evaluation criteria beyond factuality; LLM's output diversity, friendliness, and so on. We believe our study can facilitate future research on instruction selection to be more comprehensive to reach artificial general intelligence.

\chapter{Conclusion and Future Works}
\label{chap:conclusion}

In this dissertation, we propose a systemic framework that \emph{prioritize informative features and examples} to enhance each stage of the development process including feature learning, data labeling, and data selection.
Specifically, we first propose an approach to extract only informative features that are inherent to solving a target task by using auxiliary out-of-distribution data.
Next, we introduce an approach that prioritizes informative examples from unlabeled noisy data in order to reduce the labeling cost of active learning.
Lastly, we suggest an approach that prioritizes informative examples from labeled noisy data to preserve the performance of data subset selection.

For the first approach to extracting informative features, we propose \algnamea{}, a novel \textit{task-agnostic} framework to reduce the bias toward uninformative features when training DNNs. We overcome the limited applicability of the softmax-level calibration by introducing the \emph{feature-level} calibration that directly manipulates the feature output of a general feature extractor\,(e.g., a convolutional neural network). To remove the effect of undesirable features on the final task-specific module, \algnamea{} simply deactivates all undesirable features extracted from the OOD data by regularizing them as zero vectors. 
Moreover, we provide insight into how differently feature-level and softmax-level calibrations affect feature extraction by theoretic and empirical analysis of the penultimate layer activation.
We show consistent performance improvements on three types of tasks clearly demonstrating the task-agnostic nature of \algnamea{}.

For the second approach to prioritizing informative examples from unlabeled noisy data, we propose \algname{}, a novel meta-model for open-set active learning that deals with the purity-informativeness dilemma.
In detail, \algname{} finds the best balancing between the two factors, being adaptive to the noise ratio and target model status.
A clean validation set for the meta-model is obtained for free by exploiting the procedure of active learning.
A ranking loss with the skyline constraint optimizes \algname{} to make the output a {legitimate} meta-score that keeps the obvious order of two examples.
\algname{} is shown to yield the best test accuracy throughout the entire active learning rounds, thereby empirically proving the correctness of our solution to the purity-informativeness dilemma. 
Overall, we expect that our work will raise the practical usability of active learning with open-set noise.

For the third approach to prioritizing informative examples from labeled noisy data, we propose two approaches: \algnamec{} for noisy image data and \algnamed{} for noisy text data.
\algnamec{} is a noise-robust data pruning method for Re-labeling that finds a subset that maximizes the total neighborhood confidence of the training examples, thereby maximizing the re-labeling accuracy and generalization performance.
To identify a subset that maximizes the re-labeling accuracy, \algnamec{} introduces a novel metric, \textit{neighborhood confidence} which is the prediction confidence of each neighbor example in the selected subset, and the effectiveness of this metric in estimating the Re-labeling capacity of a subset is theoretically and empirically validated. Furthermore, we optimize \algnamec{} with an efficient greedy algorithm that expands the subset by selecting the example that contributes the most to increasing the total neighborhood confidence. Experimental evaluations demonstrate the substantial superiority of \algnamec{} compared to existing pruning methods in the presence of label noise.

\algnamed{} is a novel instruction selection method that can align LLMs to be more factual and preferable.
We first provide a comprehensive study showing the cleanness, diversity, and quality of the selected instruction set are crucial to the factuality of LLMs.
Based on this, we propose a systemic selection framework called \algnamed{} that ensures diversity with clustering on embedding representations and ensures cleanness and quality with cluster-wise prompting for ranking instruction.
Experiments on both factuality and preference benchmark, \algnamed{} outperforms existing instruction selection algorithms and the aligned LLM with the original Alpaca-52k dataset.

Overall, we believe our systemic framework to prioritize informative features and examples can enhance the development cycle of deep learning in a data-centric AI view. 



\bibliographystyle{unsrt}
\bibliography{reference}




\curriculumvitae[3]

    \begin{personaldata}
        \name       {Dongmin Park}
        \email      {dongminpark@kaist.ac.kr}    
     \end{personaldata}

    \begin{education}
        \item[2013. 3.\ --\ 2018. 2.] Pohang University of Science and Technology\,(POSTECH), Pohang, Korea B.S. in Industrial Management Engineering (major) and in Computer Science Engineering (minor)
        \item[2018. 3.\ --\ 2020. 2.] Korea Advanced Institute of Science and Technology(KAIST), Daejeon, Korea\\ M.S. in Graduate School of Data Science
        \item[2020. 3.\ -- 2024. 2.] Korea Advanced Institute of Science and Technology(KAIST), Daejeon, Korea\\ Ph.D. in Graduate School of Data Science
    \end{education}

    \nobibliography*
    \begin{publication}
        \item \bibentry{park2023robust}.
        \item \bibentry{park2022multi}.
        \item \bibentry{park2022meta}.
        \item \bibentry{song2022learning_bio}.
        \item \bibentry{park2022active_bio}.
        \item \bibentry{kim2022meta}.
        \item \bibentry{park2021task_bio}.
        \item \bibentry{song2021morph}.
        \item \bibentry{kim2020hi}.
        \item \bibentry{park2020trap_bio}.
        \item \bibentry{song2019does}.
        \item \bibentry{park2019mlat}.
    \end{publication}
  \label{paperlastpagelabel}     
\end{document}